\documentclass[runningheads]{llncs}

\usepackage{eccv}

\usepackage{eccvabbrv}

\usepackage{graphicx}
\usepackage{booktabs}
\usepackage{colortbl}
\usepackage{multirow}
\usepackage{rotating}
\usepackage{tikz}
\usepackage{float}
\newcommand{\win}[1]{\cellcolor{orange!25}\textbf{#1}}

\usepackage[accsupp]{axessibility}  

\usepackage[hidelinks]{hyperref}

\usepackage{orcidlink}

\begin{document}

\title{Deformable Triangle Splatting:\\ Flexible Primitives for Real-Time\\ Radiance Field Rendering}

\titlerunning{DETRIS}

\author{Oriol Jiménez-Ayguadé\,\orcidlink{0009-0006-1280-6820} \and
	Antonio Agudo\,\orcidlink{0000-0001-6845-4998}}

\authorrunning{O.~Jiménez-Ayguadé and A.~Agudo}

\institute{Institut de Robòtica i Informàtica Industrial, CSIC-UPC, Barcelona, Spain\\
	\email{\{ojimenez,aagudo\}@iri.upc.edu}}

\maketitle

\vspace{-0.45cm}
{\begin{center}
	\centering
	\includegraphics[width=0.95\linewidth]{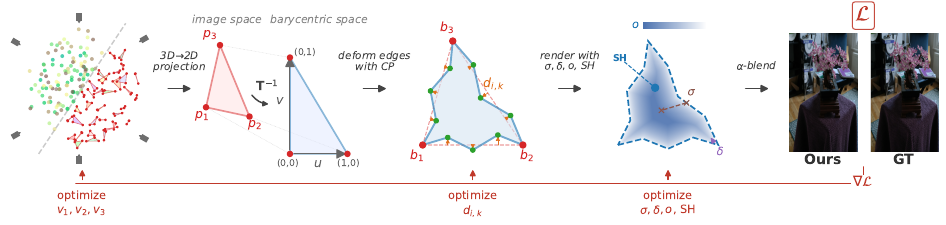}
	\vspace{-0.4cm}
	\captionof{figure}{\textbf{Method overview.} From left to right: A sparse SfM point cloud initializes the primitives; Each triangle is first projected from 3D to image space ($\mathbf{p}_i$), then mapped via~$\mathbf{T}^{-1}$ to a canonical barycentric coordinate frame, ensuring view-independent deformations; $K$ learnable Control Points (CP) per edge (green dots) displace the boundary by~$d_{i,k}$ along edge normals (orange arrows), producing a deformable piecewise-linear contour; The deformed polygon is rendered with a smooth distance-based falloff controlled by~$\sigma$, corner smoothness~$\delta$, per-primitive opacity~$o$, and Spherical-Harmonic (SH) view-dependent color; Finally, primitives are composed via front-to-back $\alpha$-blending; gradients of the loss~$\mathcal{L}$ optimize vertex positions~$\mathbf{v}_i$, displacements~$d_{i,k}$, and rendering parameters~($\sigma,\delta,o$,~SH).}
	\label{fig:pipeline}
    \vspace{-0.2cm}
\end{center}}

\begin{abstract}
	Recent radiance field methods represent scenes with 2D primitives that offer surface alignment and efficient rasterization, from Gaussian disks to triangles, yet all rely on convex boundaries: curved and concave structures demand excessive primitives. We introduce Deformable Triangle Splatting, which augments each triangle with $K$ control points per edge, each parameterized by a single learnable scalar displacement that shifts the boundary inward or outward, enabling non-convex shape representation while preserving the three base vertices that define the 3D plane. To render these non-convex primitives differentiably, we design a rasterization pipeline in the triangle's barycentric coordinate space, ensuring view-consistent rendering. A winding number test determines whether each pixel lies inside the deformed primitive, and a window function controlled by two learnable parameters, sharpness and corner smoothness, together with a per-primitive scalar opacity, produces the smooth opacity transition from interior to boundary. Validation is done in a variety of real-world scenes, outperforming recent works based on non-volumetric primitives in terms of visual quality and versatility while still achieving competitive rendering efficiency.\\
	Project page: \href{https://orioljim1.github.io/detris}{\textcolor{blue}{orioljim1.github.io/detris}}
\end{abstract}

\section{Introduction}
\label{sec:intro}

The field of novel view synthesis has been transformed by two major paradigms: Neural Radiance Fields (NeRF)~\cite{mildenhall2020nerf} and 3D Gaussian Splatting (3DGS)~\cite{kerbl20233dgs}. While NeRF~\cite{mildenhall2020nerf} established the potential of continuous volumetric fields for high-fidelity reconstruction, its reliance on expensive ray marching motivated acceleration techniques such as multi-resolution hash encodings~\cite{muller2022instant}. 3DGS~\cite{kerbl20233dgs} subsequently advanced the field by employing explicit anisotropic Gaussians, enabling real-time rendering without sacrificing visual quality. In spite of these advances, a fundamental challenge remains: accurately representing complex scene geometry with a unified primitive. Volumetric primitives like Gaussians inherently struggle with sharp boundaries and flat surfaces due to their smooth fall-off. Other works have explored alternative primitives such as 2D Gaussian Splatting (2DGS)~\cite{huang20242dgs} which flattens primitives for better surface alignment. Meanwhile 3D Convex Splatting (3DCS)~\cite{held2025convex} and Universal Beta Splatting (UBS)~\cite{liu2025universal} employ smooth convex hulls and Beta kernels, respectively, to capture hard edges and dense volumes. However, these methods rely on fixed or convex shapes, limiting their ability to model non-convex local geometry without excessive primitive proliferation. More recently, the most standard primitive of computer graphics and simulation was rescued in Triangle Splatting (TS)~\cite{held2025trianglesplatting}, where a soup of triangles was directly optimized, bridging differentiable rendering and traditional mesh pipelines. The use of triangle primitives is a key factor in the topic as GPUs feature dedicated hardware pipelines for ultra-efficient triangle rendering and processing. However, this representation still imposes a rigid and fixed topology based on three vertices, leading to inefficient over-tessellation when representing curved or non-convex structures.

Unlike previous work, our approach augments each triangle with $K$ learnable control points per edge, producing a soup of non-uniform primitives capable of representing a wide variety of convex and non-convex shapes. In this sense, the proposed primitive can be interpreted as a versatile representation for 3D content, as multiple specialized primitives can be learned directly from data using the same underlying structure.
To achieve that, each control point is associated with a scalar displacement that shifts the boundary inward or outward along the edge normal in the primitive's intrinsic barycentric coordinate space. This allows a single primitive to represent non-convex shapes such as arrows, crescents, or any shape with concavities; without modifying the three base vertices that define its 3D plane. Moreover, our formulation based on barycentric coordinates ensures that all per-pixel deformation computations remain strictly view-independent, unlike several existing methods that rely on a dependent, as shown later.

To render these deformed primitives differentiably, we introduce three complementary mechanisms. First, a ray-casting winding number test classifies pixels as inside or outside the deformed polygon, handling non-convex and self-intersecting boundaries robustly. Second, a polynomial smooth minimum over the point-to-segment distances to all boundary edges produces a $C^1$-continuous distance field whose corner rounding is governed by a learnable parameter. Third, a power-law window function controls an opacity falloff from the primitive's center to its boundary. Together, these ensure that gradients flow from the photometric loss through the opacity to the boundary control points. We provide an efficient CUDA implementation that is extensively evaluated on challenging scenes, outperforming recent works in terms of visual fidelity and versatility while still obtaining competitive rendering speed.

\section{Related Work}
\label{sec:related}

We next review the most related work dealing with neural radiance fields and primitive-based differentiable rendering.

\noindent\textbf{Neural Radiance Fields.}
In recent years, NeRF~\cite{mildenhall2020nerf} has emerged as a leading approach for image-based 3D reconstruction and novel view synthesis. This work introduced a continuous implicit scene representation parameterized by multilayer perceptrons, demonstrating high-fidelity view synthesis from sparse input images where the camera locations are obtained by Structure-from-Motion (SfM)~\cite{schonberger2016sfmrevisited}. Since then, NeRF-based methods have progressed rapidly~\cite{muller2022instant, xu2022point, verbin2021ref, zhang2022nerfusion, niemeyer2022regnerf}, improving robustness through anti-aliasing techniques~\cite{barron2022mipnerf360, barron2023zipnerf}, modeling dynamic scenes~\cite{nerfies,Fridovich_cvpr23,4DPV,salortISBI26}, applying piecewise linear models for volume density~\cite{Mikaela_nips23}, quantifying the uncertainty~\cite{nerf_wild_cvpr21,shen_eccv22} and considering few-shot generalization~\cite{chan_cvpr22, du_cvpr23}. A major limitation is their high computational cost, resulting in slow training and rendering. To mitigate that, subsequent work has explored multi-resolution grid encodings~\cite{muller2022instant, chen_eccv22, kulhanek_iccv23}, and even baking strategies for real-time rendering~\cite{chen_cvpr23, reiser_tog23}. Despite these developments, existing methods remain largely reliant on implicit field queries rather than explicit rasterization pipelines.

\noindent\textbf{Differentiable Rendering based on primitives.}
To bridge the gap between visual fidelity and rendering speed, recent approaches have shifted towards explicit primitive-based representations. While points~\cite{Kato_cvpr18}, meshes~\cite{Liu_iccv19}, and voxels~\cite{Fridovich_cvpr22}, had been used, the proposal of 3DGS~\cite{kerbl20233dgs} revolutionized this direction by representing scenes with millions of anisotropic 3D Gaussians projected via a tile-based rasterizer, enabling real-time rendering at 1080p and beyond. Based on that work, other approaches focus on enhancing shape representation~\cite{guedon_eccv24}, improving memory efficiency~\cite{chan_cvpr24, lu_cvpr24}, proposing better densification and pruning strategies~\cite{kheradmand2024probabilistic} or handling dynamic models~\cite{wu_arxiv24,DeFan_cvpr25,Hu_cvpr24,Bae_eccv24}. Moreover, another line of approaches extended this with more flexible kernel functions~\cite{Hamdi_cvpr24,chen_arxiv24,Huang_arxiv24,liu2025deformable}, linear~\cite{Lutzow_arxiv25} and quadratic~\cite{zhang_iccv25} primitives as well as texture billboards~\cite{Rong_arxiv24,svitov2024bbsplat}. 3DCS~\cite{held2025convex} utilized smooth convex hulls to capture hard edges and dense volumes. However, the volumetric nature of 3D Gaussians poses inherent limitations for surface modeling: their smooth falloff produces {\em fuzzy} geometry lacking well-defined surfaces, problematic for downstream tasks requiring accurate normals such as relighting or material editing. 2DGS~\cite{huang20242dgs} addressed this by flattening 3D Gaussians into 2D oriented disks, enforcing stronger surface alignment. TS~\cite{held2025trianglesplatting} exploited surface triangular primitives, leveraging its ubiquity in graphics to ensure mesh compatibility and demonstrating state-of-the-art results among non-volumetric methods. However, this method still produces tessellation artifacts in some challenging geometries as well as low sharpness.

We overcome most of the limitations of previous methods with an approach that introduces deformable triangular primitives. To this end, our approach retains the triangle as the base primitive for its geometric interpretability, but augments it with learnable boundary deformations to increase the versatility and adaptability of our primitives to capture a wide spectrum of real-world scenarios. Unlike 2DGS~\cite{huang20242dgs}, 3DCS~\cite{held2025convex}, or TS~\cite{held2025trianglesplatting}, which rely on fixed shapes (disks, convex hulls, or rigid triangles, respectively), our primitive is adaptable according to the data, resulting in a soup of non-uniform primitives that correctly represents both convex and non-convex shapes by means of a unified formulation while reducing the number of primitives required to cover complex geometries. Moreover, we propose a view-independent rendering pipeline based on projecting primitives onto barycentric coordinates. We are not aware of any other work jointly offering all these characteristics.

\section{Our Method}
\label{sec:method}

We present \textbf{DEformable TRIangle Splatting} (DETRIS), a differentiable rendering method that augments triangle-based primitives with learnable boundary deformations. Conceptually, our approach follows the deformable template paradigm introduced by~\cite{terzopoulos1987elastically}, where each primitive begins as a reference triangle whose boundary deforms under photometric supervision, rather than relying on internal energy terms based on classical deformable models. Our approach enables the representation of complex, non-convex shapes while maintaining the efficiency of explicit surface-based rendering~\cite{kerbl20233dgs,liu2025deformable,huang20242dgs,held2025trianglesplatting}. Figure~\ref{fig:pipeline} provides an overview of our approach. In this section, we first describe our deformable primitive model and its differentiable rasterization pipeline, followed by our adaptive densification strategy and the optimization procedure.

\subsection{Differentiable Rendering of Deformable Primitives}
\label{sec:rendering}

\subsubsection{Deformable K-Point Primitives.}
Each primitive is defined by three freely optimizable 3D vertices $\{\mathbf{v}_1, \mathbf{v}_2, \mathbf{v}_3\} \subset \mathbb{R}^3$ that determine a triangular base, as in TS~\cite{held2025trianglesplatting}. After perspective projection, these vertices map to 2D image-space positions $\mathbf{p}_i \in \mathbb{R}^2$, which in turn define a \emph{barycentric coordinate system} $\mathbf{b} = (u, v)$ with $\mathbf{p}_1 \mapsto (0,0)$, $\mathbf{p}_2 \mapsto (1,0)$, $\mathbf{p}_3 \mapsto (0,1)$. The forward affine map from barycentric to image coordinates is $\mathbf{p} = \mathbf{p}_1 + \mathbf{T}\,\mathbf{b}$, where $\mathbf{T} = [\mathbf{p}_2 {-} \mathbf{p}_1,\; \mathbf{p}_3 {-} \mathbf{p}_1] \in \mathbb{R}^{2 \times 2}$. During rasterization, each pixel location $\mathbf{p}$ is mapped to barycentric coordinates via the per-primitive inverse $\mathbf{b} = \mathbf{T}^{-1}(\mathbf{p} - \mathbf{p}_1)$, which we precompute once per triangle before the per-pixel render pass.

Every boundary deformation is performed in this intrinsic barycentric space, making it {\em view-independent}. Each of the three edges is subdivided by $K$ control points. For edge $i \in \{1,2,3\}$ connecting barycentric vertices $\mathbf{b}_i^{\text{start}}$ and $\mathbf{b}_i^{\text{end}}$, the $k$-th control point with $k \in \{1, \dots, K\}$ is placed at:
\begin{equation}
	\mathbf{b}_{i,k} = \mathbf{b}_i^{\text{start}} + \frac{k}{K+1}\,\mathbf{e}_i + d_{i,k}\,\hat{\mathbf{n}}_i ,
	\label{eq:control_point}
\end{equation}
where $\mathbf{e}_i = \mathbf{b}_i^{\text{end}} - \mathbf{b}_i^{\text{start}}$ is the barycentric edge vector, $d_{i,k} \in \mathbb{R}$ is a learnable scalar displacement, and $\hat{\mathbf{n}}_i$ is the unit outward normal to edge $i$ in barycentric space, computed as the normalized perpendicular to $\mathbf{e}_i$ oriented away from the opposite vertex. The original triangle vertices are not displaced in this step,  they are instead freely optimized on the 3D space. This produces a piecewise-linear boundary with a total of $M = 3(K{+}1)$ vertices. A positive $d_{i,k}$ expands the boundary outward (beyond the original triangle edge), while a negative value contracts it inward, allowing concavities. Since the displacements are in barycentric units, they are scale-invariant with respect to the triangle's image-space size.

\subsubsection{Inside/Outside Classification via Winding Number.}
\label{sec:winding}
For convex shapes, a simple half-plane intersection test suffices for rasterization~\cite{MarschnerCG2021}. However, once CPs deform the boundary into non-convex shapes, this test is no longer valid. To handle this, we employ a non-zero winding rule~\cite{jacobson2013bounded}, implemented via ray casting. For a query point $\mathbf{x}$ in barycentric coordinates, we cast a horizontal ray and count signed intersections with the deformed polygon edges as:
\begin{equation}
	{\small
		\omega(\mathbf{x}) = \sum_{m=1}^{M} \psi_m(\mathbf{x}), \;\, \psi_m = \begin{cases}
			{+}1  & \text{if } y_m \le y_{\mathbf{x}} < y_{m+1} \;\land\; (\mathbf{p}_m - \mathbf{x}) \times (\mathbf{p}_{m+1} - \mathbf{x}) > 0 \\
			{-}1  & \text{if } y_{m+1} \le y_{\mathbf{x}} < y_m \;\land\; (\mathbf{p}_m - \mathbf{x}) \times (\mathbf{p}_{m+1} - \mathbf{x}) < 0 \\
			\;\;0 & \text{otherwise}
		\end{cases}}\nonumber
	\label{eq:winding}
\end{equation}
where $\mathbf{p}_m$ are the vertices of the deformed polygon in barycentric space, $y_m$ their vertical coordinates, and $\times$ denotes a 2D cross product. A pixel is classified as \emph{inside} if $\omega(\mathbf{x}) \neq 0$. For simple (non-self-intersecting) polygons, this is equivalent to the even-odd ray casting rule; for self-intersecting edges, it correctly handles overlapping regions. Pixels with $\omega(\mathbf{x}) = 0$ are skipped, contributing no opacity.

\subsubsection{Distance Field with Polynomial Smooth Minimum.}
\label{sec:distance_field}
For each interior pixel, we compute an Euclidean distance from $\mathbf{x}$ to every boundary segment $(\mathbf{p}_m, \mathbf{p}_{m+1})$ as:
\begin{equation}
	D_m(\mathbf{x}) = \left\| \mathbf{x} - \left[\mathbf{p}_m + \text{clamp}\!\left(\frac{(\mathbf{x} - \mathbf{p}_m) \cdot \mathbf{e}_m}{\|\mathbf{e}_m\|^2},\; 0,\; 1\right)\mathbf{e}_m\right] \right\| ,
	\label{eq:segment_dist}
\end{equation}
where $\mathbf{e}_m = \mathbf{p}_{m+1} - \mathbf{p}_m$. This is the standard point-to-line-segment distance, computed in barycentric coordinates.

3DCS~\cite{held2025convex} introduced a smoothness parameter $\delta$ that controls corner rounding via a LogSumExp approximation to the max function over half-plane distances. We adopt the same conceptual role for $\delta$---controlling how sharp or rounded the corners of the primitive appear---but replace the LogSumExp formulation with a \textbf{polynomial smooth minimum}~\cite{iquilezles2013smin} over segment distances. This choice has three advantages: 1) the polynomial smooth minimum is $C^1$-continuous and produces bounded, localized corner rounding (the smoothing only affects regions within distance $g$ of a corner), whereas LogSumExp produces global smoothing~\cite{iquilezles2013smin}; 2) it requires only additions, multiplications, and a \texttt{max}---no transcendental functions---making it substantially cheaper to evaluate per pixel on GPU than the \texttt{exp}/\texttt{log} calls required by LogSumExp; and 3) it operates directly on the minimum of unsigned distances rather than the maximum of signed half-plane distances, which is more natural for our piecewise-linear boundary with $M$ parts. Specifically, we apply a sequential pairwise fold as:
\begin{equation}
	\text{smin}(a, b;\, g) = \min(a, b) - \frac{g}{4}\,h^2, \quad \textrm{where} \quad h = \max\!\left(1 - \frac{|b - a|}{g},\; 0\right) ,
	\label{eq:smin}
\end{equation}
with smoothing radius $g = \delta \,\, r_{\text{bary}}$, where $\delta$ is a learnable per-primitive parameter (activated via softplus to ensure positivity) and $r_{\text{bary}} = \frac{2 - \sqrt{2}}{2} \approx 0.293$ is the inradius of the unit barycentric triangle. The final smoothed distance is obtained by sequentially folding all segment distances:
\begin{equation}
	S_1 = D_1, \qquad S_m = \mathrm{smin}(S_{m},\, D_m;\, g), \;\; m = \{2,\dots,M\}, \qquad S(\mathbf{x}) = S_{M} .
	\label{eq:smooth_dist}
\end{equation}

\begin{figure}[t]
	\centering
	\newcommand{\sdimg}[1]{\includegraphics[width=\linewidth]{#1}}
	\newcommand{\sdlbl}[1]{\rotatebox[origin=c]{90}{\tiny #1}}
	\setlength{\tabcolsep}{0.5pt}
	\renewcommand{\arraystretch}{0}
	\begin{tikzpicture}
		\node[inner sep=0pt] (grid) {%
			\begin{tabular}{@{}
				>{\centering\arraybackslash}m{3.5mm} @{\,}
				>{\centering\arraybackslash}m{0.105\linewidth} @{\,}
				>{\centering\arraybackslash}m{0.105\linewidth} @{\,}
				>{\centering\arraybackslash}m{0.105\linewidth} @{\,}
				>{\centering\arraybackslash}m{0.105\linewidth} @{\,}
				>{\centering\arraybackslash}m{0.105\linewidth} @{}}
				                                                      & \tiny $\sigma{=}0.1$
				                                                      & \tiny $\sigma{=}0.3$
				                                                      & \tiny $\sigma{=}0.7$
				                                                      & \tiny $\sigma{=}1.5$
				                                                      & \tiny $\sigma{=}3.0$ \\[1pt]
				\sdlbl{$\delta{=}0.01$}                               &
				\sdimg{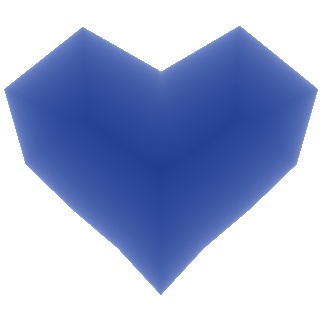} &
				\sdimg{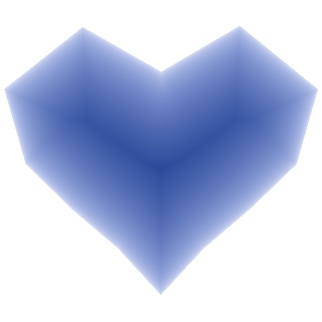} &
				\sdimg{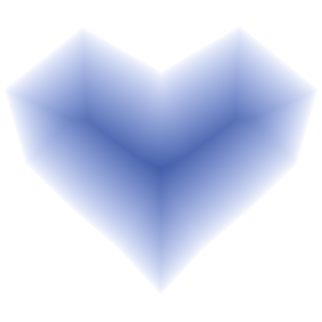} &
				\sdimg{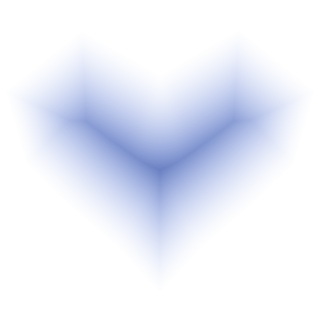} &
				\sdimg{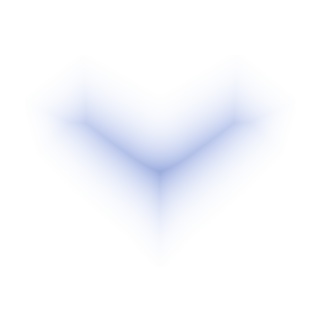}                        \\
				\sdlbl{$\delta{=}0.50$}                               &
				\sdimg{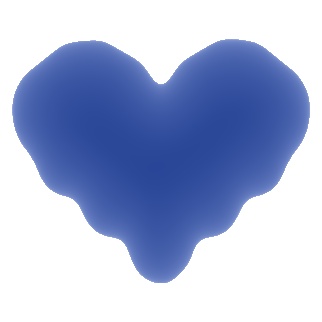}  &
				\sdimg{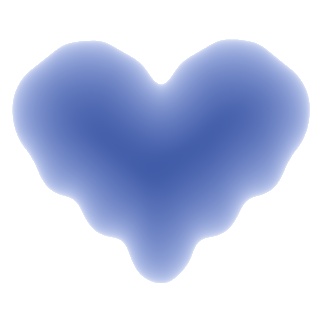}  &
				\sdimg{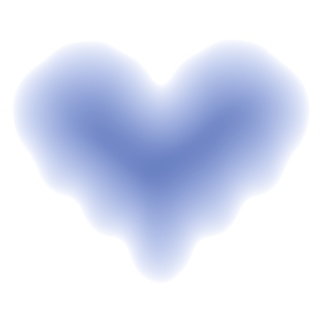}  &
				\sdimg{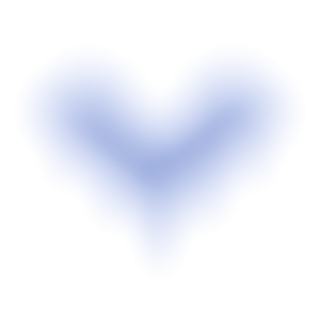}  &
				\sdimg{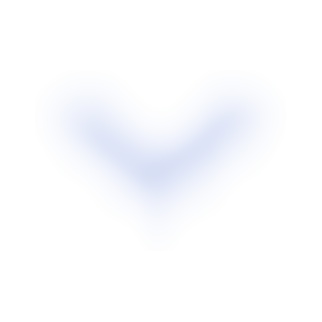}                         \\
				\sdlbl{$\delta{=}1.0$}                                &
				\sdimg{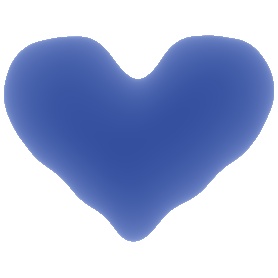}  &
				\sdimg{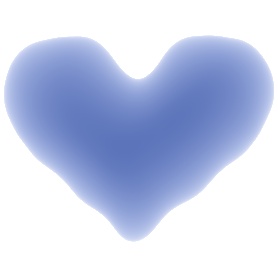}  &
				\sdimg{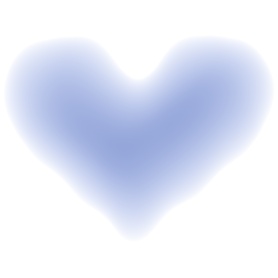}  &
				\sdimg{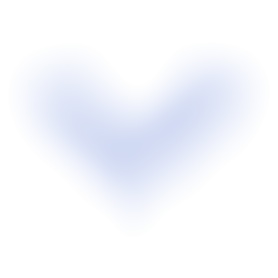}  &
				\sdimg{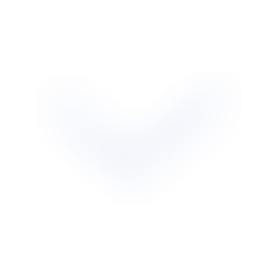}                         \\
			\end{tabular}%
		};
	\end{tikzpicture}
	\vspace{-0.3cm}
	\caption{\textbf{$\sigma$ and $\delta$ interpretation.} A deformable heart primitive with $K$=3 is rendered with increasing falloff exponent $\sigma$ and corner smoothness $\delta$. Higher $\sigma$ concentrates opacity near the medial axis; higher $\delta$ rounds the sharp corners into smooth curves.}
	\label{fig:sigma_delta}
\end{figure}

When $\delta$ is small, $g \to 0$ and the smooth minimum reduces to the hard minimum, yielding sharp corners. In contrast, for large values, corners become visibly rounded. Figure~\ref{fig:sigma_delta} visualizes the effect of $\delta$ and $\sigma$ on a rendered primitive.

\subsubsection{Power-Law Window Function.}
\label{sec:window}
The smoothed distance $S(\mathbf{x})$ is converted to opacity via a power-law window, following TS~\cite{held2025trianglesplatting}. Unlike the sigmoid-based indicator of 3DCS~\cite{held2025convex}, whose tail spreads indefinitely as $\sigma$ decreases, causing every primitive to contribute opacity to every pixel, the power-law window is exactly zero outside the primitive boundary. The key adaptation for our deformable boundary is the normalization radius $R$: TS~\cite{held2025trianglesplatting} uses the incenter distance of the rigid triangle, which is no longer valid when the boundary deforms. We instead compute $R$ as the \emph{polygon inradius} of the deformed boundary via the shoelace formula~\cite{orourke1998computational}:
\begin{equation}
	\phi(\mathbf{x}) = \min\!\left(1,\;\frac{S(\mathbf{x})}{R}\right), \quad R = \frac{2\,A}{P},
	\label{eq:phi}
\end{equation}
where $A = \frac{1}{2}\left|\sum_{m} (\mathbf{p}_m \times \mathbf{p}_{m+1})\right|$ is the polygon area and $P = \sum_{m} \|\mathbf{p}_{m+1} - \mathbf{p}_m\|$ its perimeter, both in barycentric coordinates. For a convex polygon, $R$ equals the true inradius; for a non-convex one it is an upper bound, so $\phi \leq 1$ everywhere in the interior. For an undeformed triangle, $R$ recovers $r_{\text{bary}}$, and the computation requires only $\mathcal{O}(M)$ operations during preprocessing. The per-pixel opacity is then:
\begin{equation}
	\alpha(\mathbf{x}) = \min\!\left(0.99,\;\; o \cdot \phi(\mathbf{x})^{\sigma}\right) ,
	\label{eq:alpha}
\end{equation}
where $o \in [0,1]$ is a per-primitive learnable base opacity and $\sigma > 0$ a learnable falloff exponent (activated via an exponential function): large $\sigma$ concentrates opacity near the center, while small $\sigma$ yields a nearly uniform profile.

\subsubsection{Why Barycentric Space?}
\label{sec:why_bary}
\begin{figure}[t!]
	\centering
	\resizebox{0.86\textwidth}{!}{
		\setlength{\tabcolsep}{2pt}
		\renewcommand{\arraystretch}{0}
		\begin{tabular}{@{}c@{\hspace{8pt}}cccc@{\hspace{180pt}}c@{\hspace{8pt}}cccc@{}}
			{ }                                                                                                                              & {\fontsize{240}{130}\selectfont 0\textdegree} & {\fontsize{240}{130}\selectfont 30\textdegree} & {\fontsize{240}{130}\selectfont 45\textdegree} & {\fontsize{240}{130}\selectfont 60\textdegree} & { } & {\fontsize{240}{130}\selectfont 0\textdegree} & {\fontsize{240}{130}\selectfont 30\textdegree} & {\fontsize{240}{130}\selectfont 45\textdegree} & {\fontsize{240}{130}\selectfont 60\textdegree} \\[4pt]
			\smash{\raisebox{0.323\linewidth}{\rotatebox[origin=c]{90}{\fontsize{340}{130}\selectfont TS~\cite{held2025trianglesplatting}}}} &
			\includegraphics[clip,width=0.56\linewidth]{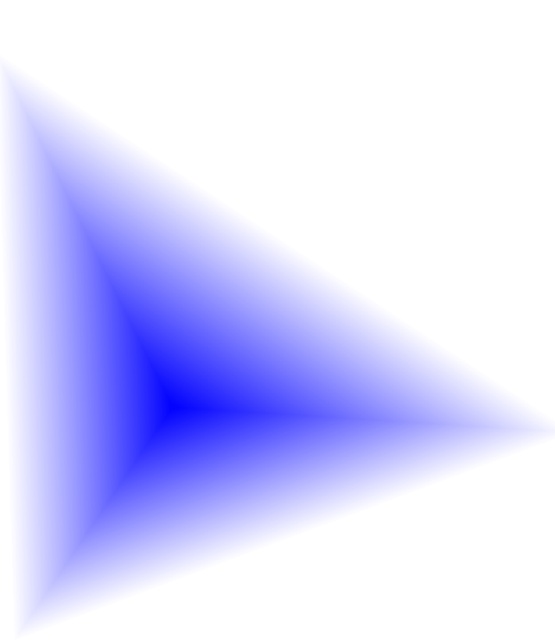}                                              &
			\includegraphics[clip,width=0.56\linewidth]{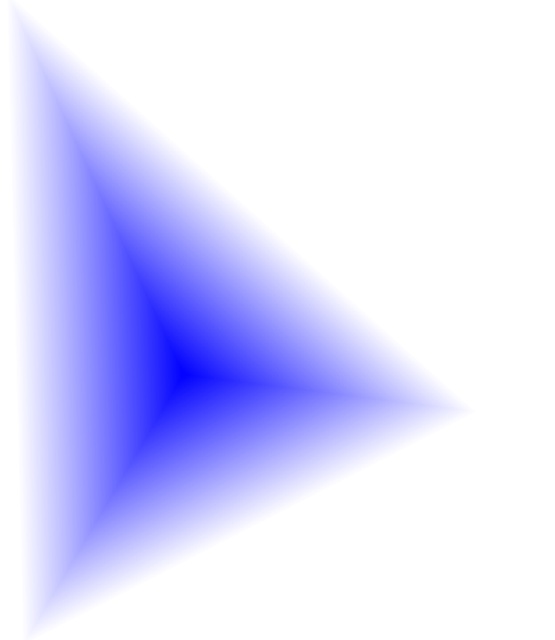}                                              &
			\includegraphics[clip,width=0.56\linewidth]{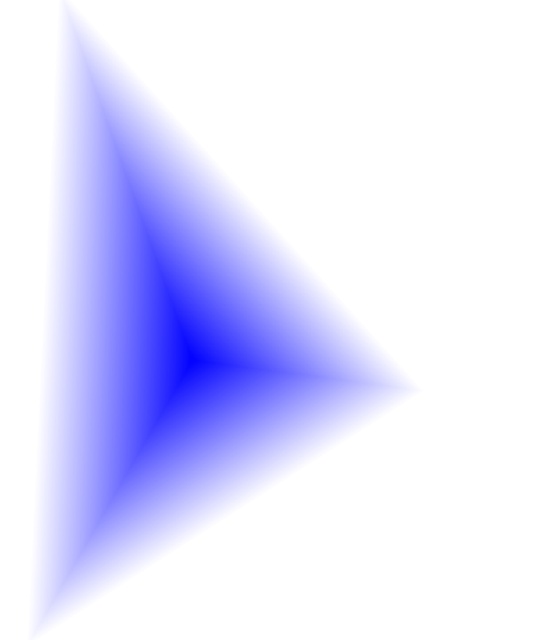}                                              &
			\includegraphics[clip,width=0.56\linewidth]{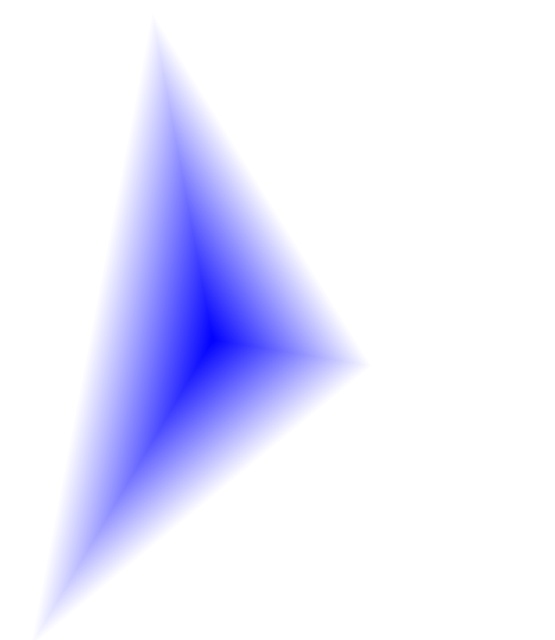}                                              &
			\smash{\raisebox{0.323\linewidth}{\rotatebox[origin=c]{90}{\fontsize{340}{130}\selectfont Image space}}}                         &
			\includegraphics[clip,width=0.7\linewidth]{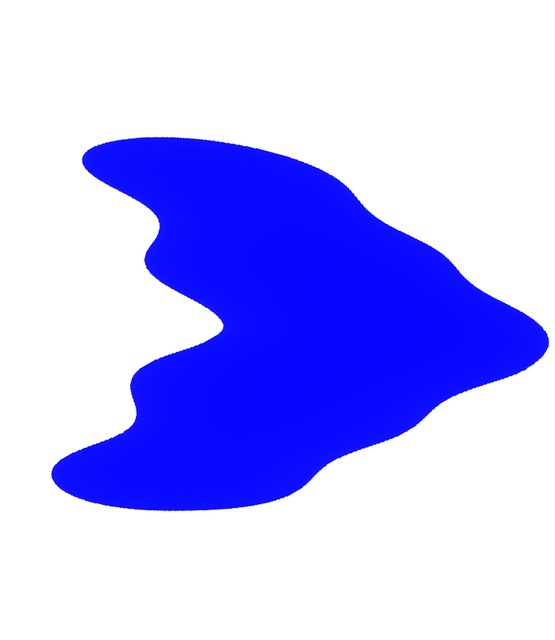}                                   &
			\includegraphics[clip,width=0.7\linewidth]{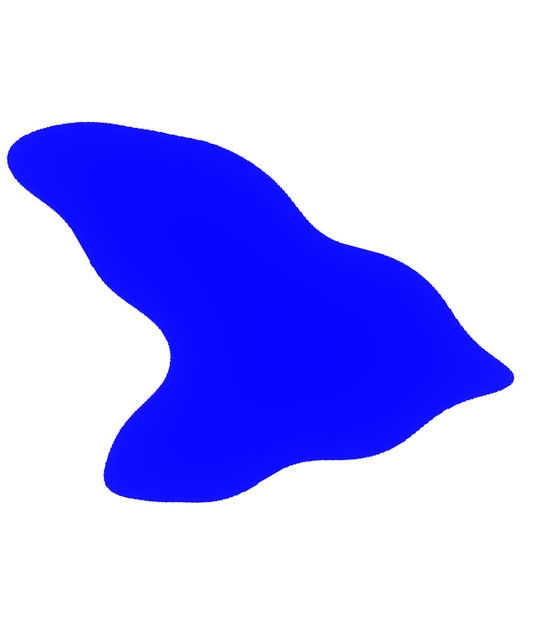}                                   &
			\includegraphics[clip,width=0.7\linewidth]{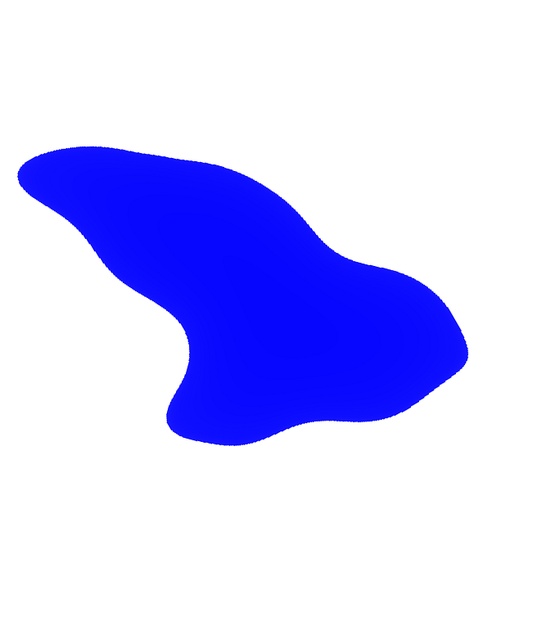}                                   &
			\includegraphics[clip,width=0.7\linewidth]{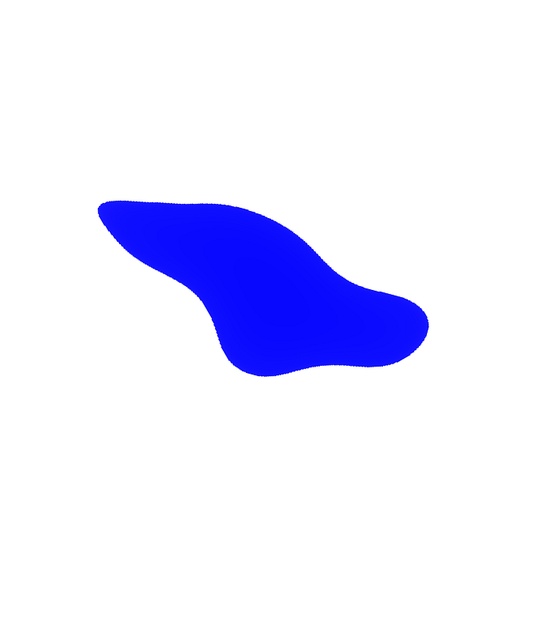}                                                                                                                                                                                                                                                                                                                                                                                                                                               \\[4pt]
			\smash{\raisebox{0.323\linewidth}{\rotatebox[origin=c]{90}{\fontsize{340}{130}\selectfont Ours}}}                                &
			\includegraphics[clip,width=0.56\linewidth]{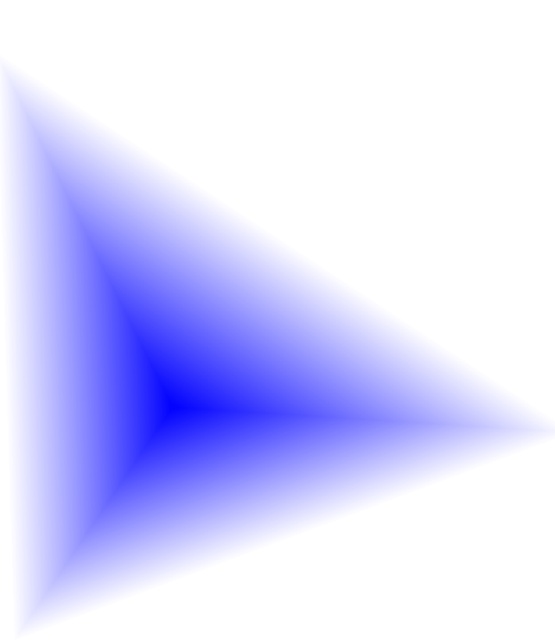}                                             &
			\includegraphics[clip,width=0.56\linewidth]{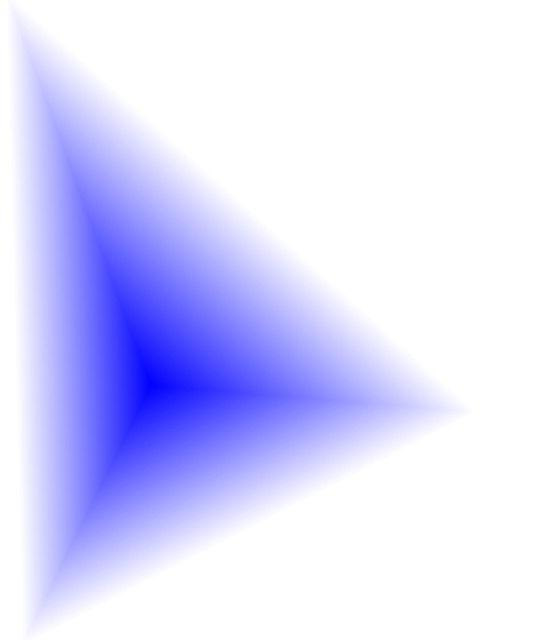}                                             &
			\includegraphics[clip,width=0.56\linewidth]{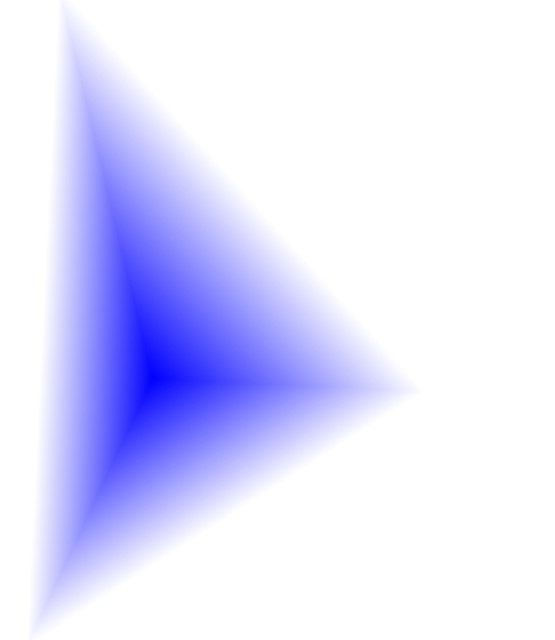}                                             &
			\includegraphics[clip,width=0.56\linewidth]{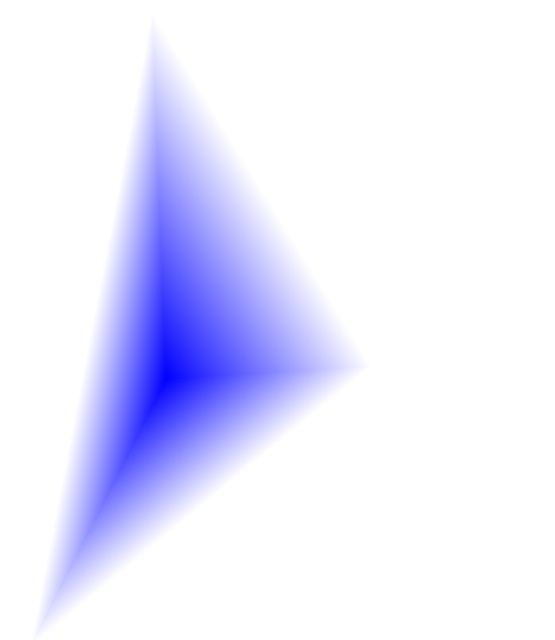}                                             &
			\smash{\raisebox{0.323\linewidth}{\rotatebox[origin=c]{90}{\fontsize{340}{130}\selectfont Ours}}}                                &
			\includegraphics[clip,width=0.7\linewidth]{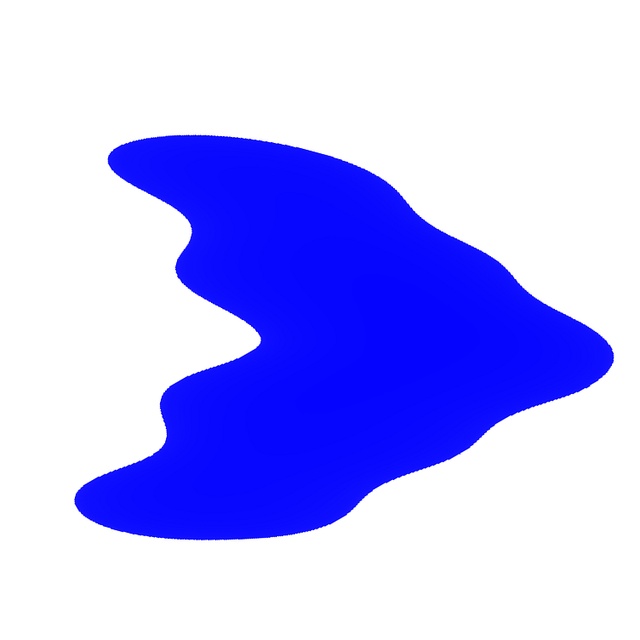}                                     &
			\includegraphics[clip,width=0.7\linewidth]{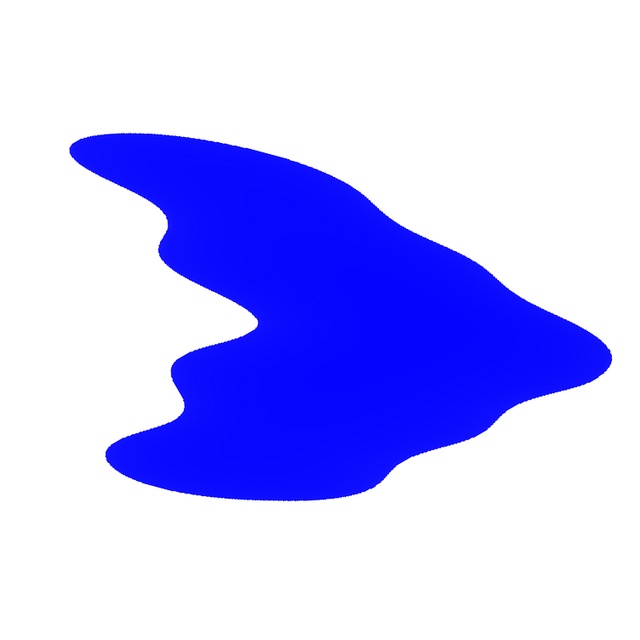}                                     &
			\includegraphics[clip,width=0.7\linewidth]{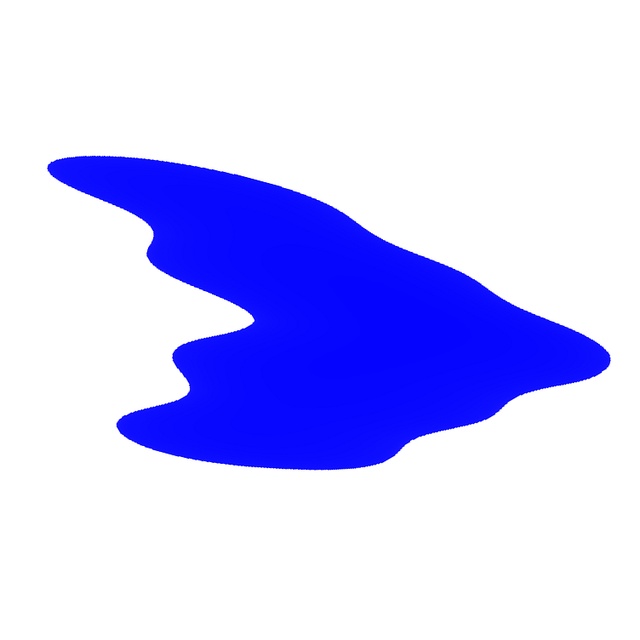}                                     &
			\includegraphics[clip,width=0.7\linewidth]{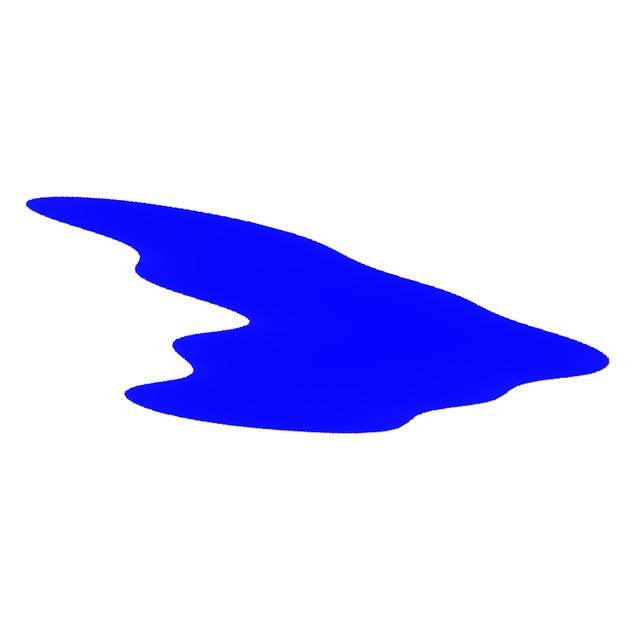}                                                                                                                                                                                                                                                                                                                                                                                                                                                 \\[4pt]
			\smash{\raisebox{0.323\linewidth}{\rotatebox[origin=c]{90}{\fontsize{340}{130}\selectfont $|\Delta|$}}}                          &
			\includegraphics[clip,width=0.56\linewidth]{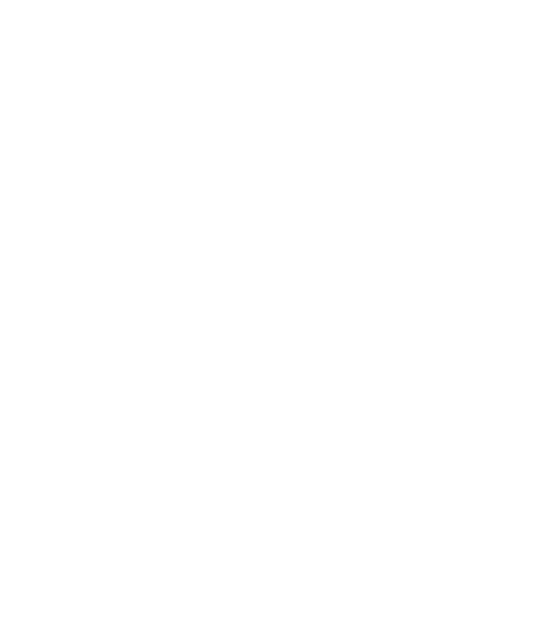}                                            &
			\includegraphics[clip,width=0.56\linewidth]{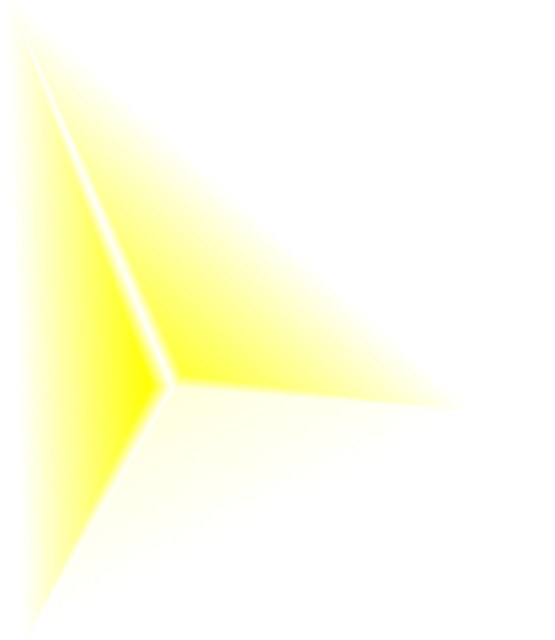}                                            &
			\includegraphics[clip,width=0.56\linewidth]{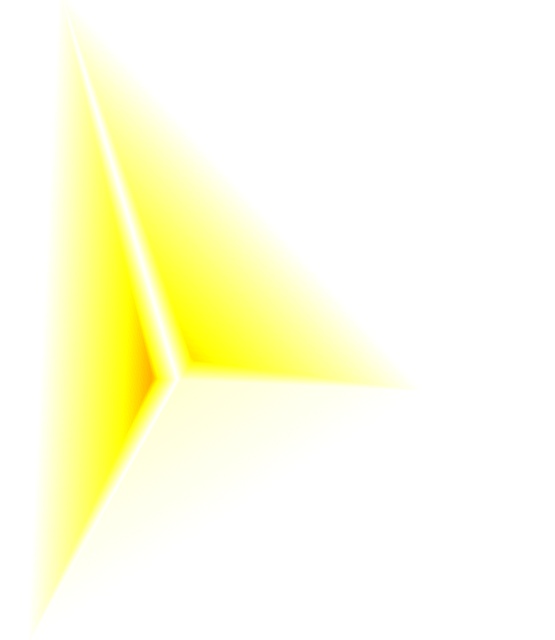}                                            &
			\includegraphics[clip,width=0.56\linewidth]{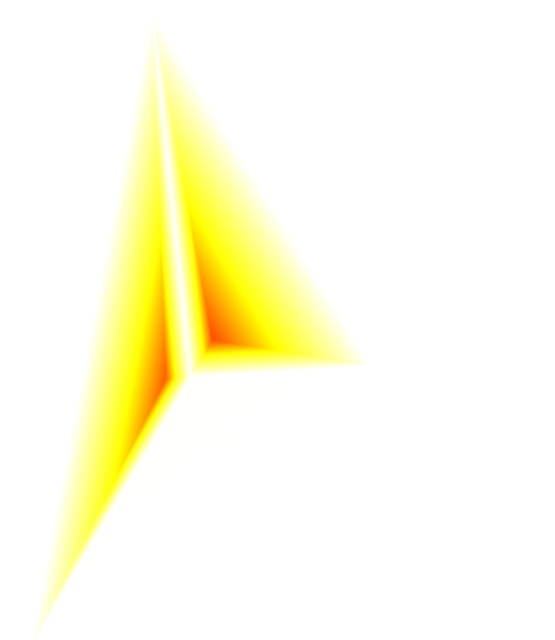}                                            &
			\smash{\raisebox{0.323\linewidth}{\rotatebox[origin=c]{90}{\fontsize{340}{130}\selectfont $|\Delta|$}}}                          &
			\includegraphics[clip,width=0.7\linewidth]{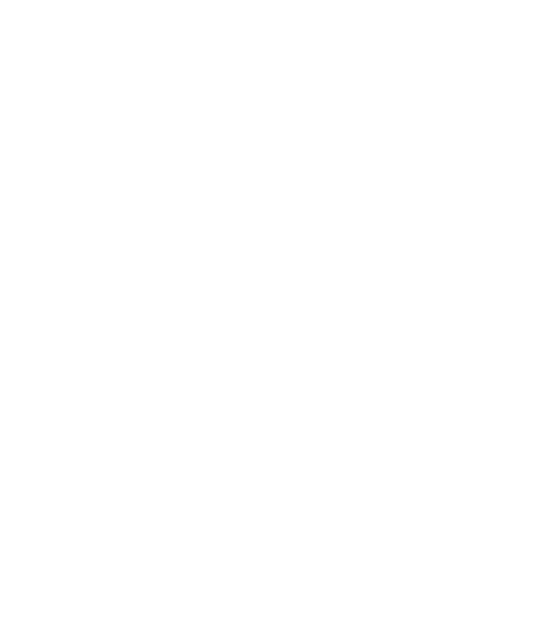}                                     &
			\includegraphics[clip,width=0.7\linewidth]{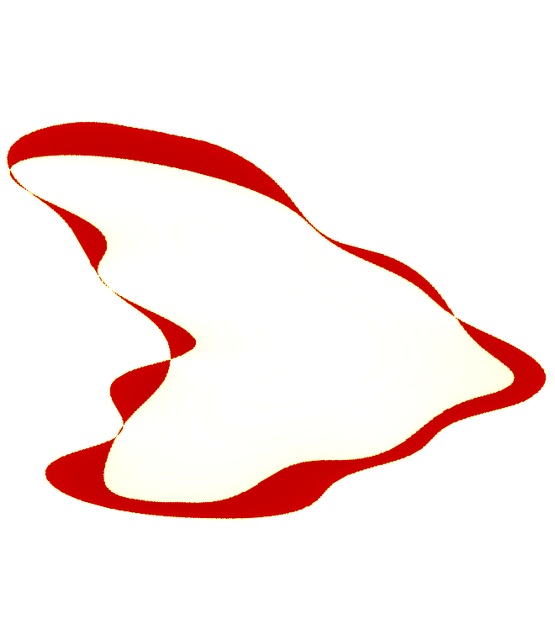}                                     &
			\includegraphics[clip,width=0.7\linewidth]{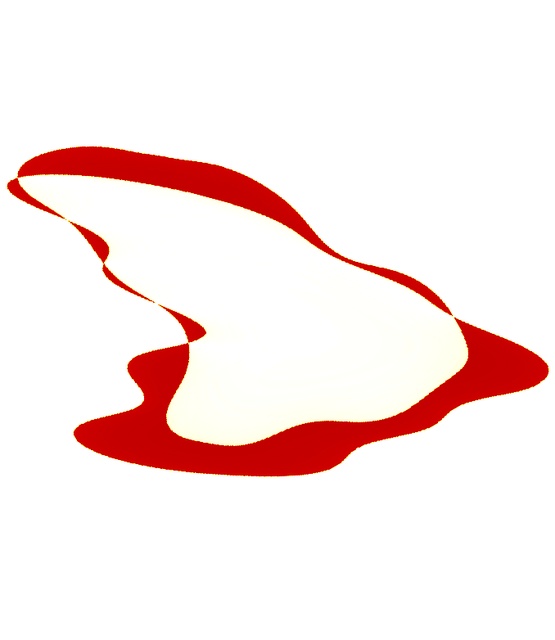}                                     &
			\includegraphics[clip,width=0.7\linewidth]{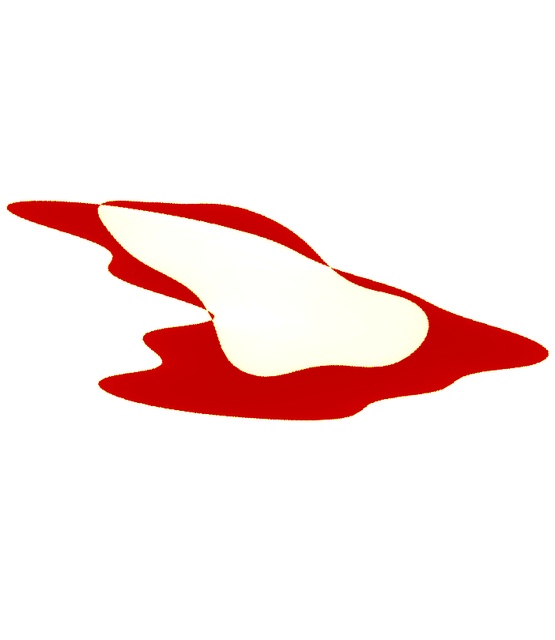}                                                                                                                                                                                                                                                                                                                                                                                                                                                 \\
		\end{tabular}}
	\vspace{-0.3cm}
	\caption{\textbf{Image-space vs.\ barycentric formulation under viewpoint change.} The same primitive is rendered from three X-axis rotations; both formulations start from a pixel-identical frontal view. \textbf{Left:} Window function $\phi^\sigma$ effect. The image-space ratio $\hat{d}/\hat{d}_{\mathrm{inc}}$~\cite{held2025trianglesplatting} diverges from our barycentric window as perspective foreshortening increases (max $|\Delta|$\,=\,181 at 60\textdegree). \textbf{Right:} Boundary deformation $\delta$: a deformed arrow with matched parameters. The image-space formulation distorts under oblique views while our barycentric shape is view-independent. Bottom rows show per-pixel absolute difference with the smallest and largest errors represented in white and red, respectively. }
	\label{fig:bary_vs_screen}
	\vspace{-0.2cm}
\end{figure}

A natural alternative, which we initially implemented and evaluated, is to define displacements and distance computations directly in image space. However, this formulation has a fundamental flaw: \textbf{the learned primitive shape becomes view-dependent}. Because the same 3D triangle projects differently under each camera, all per-pixel rendering computations (segment distances, smooth-minimum field, opacity falloff) are evaluated in pixel units that change with viewpoint, even when the displacements are fixed. As a result, a primitive that appears sharp from one viewpoint may look excessively rounded from another, and multi-view supervision produces conflicting gradients for the same displacement parameters.
By performing all deformations in the canonical barycentric frame, we decouple intrinsic shape from image projection: all boundary tests are evaluated in a fixed reference frame, and both $d_{i,k}$ and $\delta$ carry exactly the same geometric meaning in every training view, eliminating the multi-view gradient conflict. Figure~\ref{fig:bary_vs_screen}-right shows that a deformed arrow primitive maintains a consistent shape across viewpoints under our formulation, while the image-space version visibly distorts.

\textbf{View-dependence of the image-space window.}
Even the base window function of TS~\cite{held2025trianglesplatting}, $\phi(\mathbf{x}){=}(\hat{d}/\hat{d}_{\mathrm{inc}})^{\sigma}$ with distances in pixel coordinates, is affected. Although this ratio cancels uniform scaling, perspective projection is \emph{projective}, not affine: vertices at different depths undergo non-uniform foreshortening, so $\hat{d}/\hat{d}_{\mathrm{inc}}$ is not preserved across viewpoints. Figure~\ref{fig:bary_vs_screen}-left confirms this: rendering an undeformed triangle under progressive rotation, the discrepancy grows to max~$|\Delta|$\,=\,76 at $45$\textdegree{} and~181 at $60$\textdegree, concentrated on the most foreshortened edges, while the edge aligned with the rotation axis shows near-zero error, consistent with the analysis. Our barycentric formulation is exactly view-independent by construction.

\subsection{Adaptive Pruning and Densification}
\label{sec:densification}

We follow the pruning and densification strategy of TS~\cite{held2025trianglesplatting}, adapted for our novel deformable primitives. Next, we introduce the most important details.

\textbf{Pruning.} During rasterization, we track the maximum blending weight $T \cdot \alpha$ (importance score) for each primitive, with $T$ the transmittance. We prune primitives whose importance score falls below a threshold $\tau_{\text{prune}}$, as well as those not rendered in at least two training views with more than one pixel of coverage.

\textbf{Densification.} We sample primitives for densification from a probability distribution based on learned opacity and inverse sharpness ($1/\sigma$), following the spirit of probabilistic densification~\cite{kheradmand2024probabilistic}. This preferentially targets solid primitives for splitting. In particular, for primitives larger than an image-space threshold $\tau_{\text{split}}$, we apply \textbf{midpoint subdivision}, splitting each triangle into four children by connecting edge midpoints. A key challenge is preserving the deformed boundary across this topological change. We \textbf{interpolate the parent's control point displacements} onto the child edges where each parent edge's displacement profile is sampled at the appropriate half to produce the child's control points. Crucially, child edges that traverse in the opposite direction to the parent edge have their displacement sequences \textbf{reversed}, ensuring geometric continuity and preventing visual artifacts. For primitives smaller than $\tau_{\text{split}}$, we \textbf{clone} the primitive and add random in-plane noise.

\subsection{Optimization}
\label{sec:optimization}

The training loss combines photometric $l_1$ and D-SSIM terms~\cite{kerbl20233dgs}, an opacity regularizer $\mathcal{L}_{\text{op}} = \overline{|\text{sigmoid}(\hat{o})|}$ encouraging transparency, and distortion $\mathcal{L}_{d}$ and normal consistency $\mathcal{L}_{n}$ losses~\cite{huang20242dgs}. During the densification phase (iterations $< T_{\text{dense}}$), we additionally apply a size regularizer $\mathcal{L}_{s}$ that penalizes the inverse square root of the triangle area, discouraging degenerate (collapsed) primitives~\cite{held2025trianglesplatting}. A curvature term $\mathcal{L}_{\text{curv}} = \overline{\|\mathbf{d}\|^2}$ penalizes large control point displacements, encouraging smooth boundaries and preventing overfitting. Our loss is:
\begin{equation}
	\mathcal{L} = (1{-}\lambda_1)\,\mathcal{L}_1 + \lambda_1\,\mathcal{L}_{\text{D-SSIM}} + \lambda_2\,\mathcal{L}_{\text{op}} + \lambda_3\,\mathcal{L}_{d} + \lambda_4\,\mathcal{L}_{n} + \lambda_5\,\mathcal{L}_{s} + \lambda_{6}\,\mathcal{L}_{\text{curv}} ,
	\label{eq:loss}
\end{equation}
where $\lambda_{1-6}$ represent the weight coefficients.

\noindent\textbf{Gradient Management and Initialization.}
\label{sec:gradient_management}
Under joint optimization, vertex position gradients dominate the control point gradients because they directly affect projection, tile assignment, and barycentric coordinates. To mitigate this, we delay the control point learning rate to zero for the first $T_{\text{delay}} = 3000$ iterations, allowing vertices to settle into reasonable positions before enabling boundary deformation. After this warm-up, control points train jointly with vertices using the standard optimizer, a simple delay that is sufficient for effective control point utilization, as validated later. For initialization, starting from a sparse SfM point cloud~\cite{schonberger2016sfmrevisited}, we create one triangle per 3D point. Let $\mathbf{q} \in \mathbb{R}^3$ be a point and $r$ the distance to its nearest neighbor. Three vertices are sampled from the unit sphere, scaled by $r$, randomly rotated, and offset by $\mathbf{q}$. All control point displacements $d_{i,k}$ are set to zero, so primitives begin as standard triangles.

\section{Experimental Evaluation}
\label{sec:experiments}

In this section, we present our experimental results on Mip-NeRF 360~\cite{barron2022mipnerf360} and Tanks \& Temples~\cite{knapitsch2017tanks} datasets. We provide both quantitative and qualitative evaluation and compare our approach against state-of-the-art solutions, including the implicit methods Mip-NeRF 360~\cite{barron2022mipnerf360}, Zip-NeRF~\cite{barron2023zipnerf} and Instant-NGP~\cite{muller2022instant}; volumetric primitives such as 3DGS~\cite{kerbl20233dgs}, 3DGS-MCMC~\cite{kheradmand2024probabilistic}, Deformable Beta Splatting~\cite{liu2025deformable} and 3D Convex Splatting~\cite{held2025convex}; as well as non-volumetric primitives such as BBSplat~\cite{svitov2024bbsplat}, 2DGS~\cite{huang20242dgs} and Triangle Splatting~\cite{held2025trianglesplatting}. In our case,  we train for 30k iterations with $K{=}3$ control points per edge until otherwise stated, using an Adam optimizer~\cite{adam_opt}. To make a fair comparison, the maximum total number of primitives and the rest of hyperparameters are the same as in~\cite{held2025trianglesplatting}, the only additions are the learning rate for CPs, $\delta$, and $\lambda_6$. which can all be seen in the supplementary. For quantitative analysis, the metrics LPIPS, PSNR, and SSIM are reported~\cite{mildenhall2020nerf}.

\subsection{Control Points Deformation Analysis}
First of all, we evaluate the effect of the $K$ control points in our approach. When this value is increased, it provides greater boundary flexibility but also enhances per-primitive parameter count and, therefore, the corresponding rasterization cost (more segments means additional distance tests per pixel). To validate that, Table~\ref{tab:ablation_K} reports quality metrics together with training cost, rendering speed, and model size as a function of $K$ on the Bonsai and Garden scenes included in Mip-NeRF 360~\cite{barron2022mipnerf360}. In general, performance degrades as $K$ decreases, but both training time and model size grow approximately linearly with $K$. As the results always remain within reasonable bounds, it is not necessary to adjust this parameter precisely. In practice, we set $K$=3 in our experiments, as this value provides a good trade off between visual quality and rendering speed.

\begin{table}[t!]
	\centering
	\caption{\textbf{Quantitative results} in terms of $K$ per edge on Bonsai (left) and Garden scenes (right)~\cite{barron2022mipnerf360}. $K{=}0$ disables control-point deformation (our rasterizer only), isolating the gain due to the learned deformations.}
	\label{tab:ablation_K}
    \vspace{-0.3cm}
	\resizebox{\textwidth}{!}{%
		\begin{tabular}{l|cccccc}
			\toprule
			\multicolumn{7}{c}{\textbf{Bonsai (Indoor)}} \\
			\midrule
			$K$ & PSNR$\uparrow$ & SSIM$\uparrow$ & LPIPS$\downarrow$ & Time    & FPS$\uparrow$ & Size    \\
			\midrule
			0   & 31.58          & 0.944          & 0.163             & 43\,min & 63            & 824\,MB \\
				1   & 32.30          & 0.948          & 0.165             & 45\,min & \textbf{65}   & 816\,MB \\
			2   & 32.39          & 0.948          & 0.163             & 51\,min & 55            & 832\,MB \\
			3   & 32.44          & 0.948          & 0.162             & 59\,min & 47            & 863\,MB \\
			4   & 32.51          & 0.948          & 0.162             & 68\,min & 40            & 881\,MB \\
			5   & \textbf{32.52} & \textbf{0.948} & \textbf{0.159}    & 79\,min & 33            & 911\,MB \\
			\bottomrule
		\end{tabular}%
		\qquad
		\begin{tabular}{l|cccccc}
			\toprule
			\multicolumn{7}{c}{\textbf{Garden (Outdoor)}} \\
			\midrule
			$K$ & PSNR$\uparrow$ & SSIM$\uparrow$ & LPIPS$\downarrow$ & Time    & FPS$\uparrow$ & Size    \\
			\midrule
			0   & 27.08          & 0.860          & 0.107            & 56\,min & \textbf{49}   & 1384\,MB \\
				1   & 27.35          & 0.867          & 0.101            & 52\,min & 46            & 1460\,MB \\
			2   & 27.40          & 0.868          & 0.098            & 61\,min & 38            & 1508\,MB \\
			3   & 27.41          & 0.868          & 0.098            & 71\,min & 33            & 1542\,MB \\
			4   & 27.42          & 0.868          & 0.098            & 83\,min & 27            & 1577\,MB \\
			5   & \textbf{27.42} & \textbf{0.868} & \textbf{0.098}   & 96\,min & 23            & 1605\,MB \\
			\bottomrule
		\end{tabular}%
	}
    	\vspace{-0.4cm} 
\end{table}

\subsection{Novel View Synthesis Results}
\label{sec:real_world}

We now report our results in a quantitative manner on Table~\ref{tab:main_results}. As can be seen, our approach obtains competitive solutions by achieving among non-volumetric primitives the best LPIPS on all dataset splits, the best indoor PSNR and SSIM on Mip-NeRF 360~\cite{barron2022mipnerf360}, as well as the best T\&T~\cite{knapitsch2017tanks} LPIPS. It is worth noting that our method outperforms TS~\cite{held2025trianglesplatting} on all average metrics across the three dataset splits. On a per-scene basis, we only fall behind on two outdoor scenes (Stump and Treehill), while matching or exceeding TS~\cite{held2025trianglesplatting} on the remaining nine (see the supplementary material for full per-scene results). The consistent LPIPS improvement across \emph{all} outdoor scenes (0.206 vs.\ 0.217) confirms that the perceptual quality of our renders is higher even where PSNR is slightly lower.

\begin{table}[b!]
	\centering
	\caption{\textbf{Quantitative comparison on Mip-NeRF 360~\cite{barron2022mipnerf360} and Tanks \& Temples~\cite{knapitsch2017tanks}} datasets. The table includes implicit methods~\cite{muller2022instant,barron2022mipnerf360,barron2023zipnerf}, those based on volumetric~\cite{kerbl20233dgs,kheradmand2024probabilistic,liu2025deformable,held2025convex} and non-volumetric~\cite{svitov2024bbsplat,huang20242dgs,held2025trianglesplatting} primitives. Average columns report Mip-NeRF 360~\cite{barron2022mipnerf360} for all scenes. Train (in minutes) and Mem (in MB) are measured on the same benchmark. \textbf{Bold} indicates best among non-volumetric methods. All results reported in~\cite{held2025trianglesplatting}; $\dagger$ results reported in~\cite{liu2025deformable}; $^*$ measured on an RTX\,5090.}
	\label{tab:main_results}
	\resizebox{\textwidth}{!}{
		\begin{tabular}{ll|ccc|ccc|cccc|cccc}
			\toprule
			{ }                                       & { }                                                       & \multicolumn{3}{c|}{Outdoor Mip-NeRF 360} & \multicolumn{3}{c|}{Indoor Mip-NeRF 360} & \multicolumn{4}{c|}{Average Mip-NeRF 360} & \multicolumn{4}{c}{Tanks \& Temples}                                                                                                                                                                                               \\
			{ }                                       & Method                                                    & LPIPS$\downarrow$                         & PSNR$\uparrow$                           & SSIM$\uparrow$                            & LPIPS$\downarrow$                    & PSNR$\uparrow$    & SSIM$\uparrow$    & LPIPS$\downarrow$ & FPS$\uparrow$ & Train$\downarrow$ & Mem$\downarrow$ & LPIPS$\downarrow$ & PSNR$\uparrow$    & SSIM$\uparrow$    & FPS$\uparrow$ \\
			\midrule
			\multirow{3}{*}{\rotatebox{90}{Implicit}} & {Instant-NGP~\cite{muller2022instant}}                    & --                                        & --                                       & --                                        & --                                   & --                & --                & 0.331             & 9.4           & --                & --              & 0.305             & 21.92             & 0.745             & 14.4          \\
			{}                                        & Mip-NeRF 360~\cite{barron2022mipnerf360}                  & 0.283                                     & 24.47                                    & 0.691                                     & 0.179                                & 31.72             & 0.917             & 0.237             & 0.06          & --                & --              & 0.257             & 22.22             & 0.759             & 0.14          \\
			{}                                        & Zip-NeRF~\cite{barron2023zipnerf}                         & 0.207                                     & 25.58                                    & 0.750                                     & 0.167                                & 32.25             & 0.926             & 0.189             & 0.18          & 300               & 569             & --                & --                & --                & --            \\
			\midrule
			\multirow{4}{*}{\rotatebox{90}{Vol.}}     & {3DGS~\cite{kerbl20233dgs}}                               & 0.234                                     & 24.64                                    & 0.731                                     & 0.189                                & 30.41             & 0.920             & 0.214             & 134           & 42                & 734             & 0.183             & 23.14             & 0.841             & 154           \\
			{}                                        & 3DGS-MCMC~\cite{kheradmand2024probabilistic}$^{\dagger}$ & 0.210                                     & 25.51                                    & 0.760                                     & 0.208                                & 31.08             & 0.917             & 0.210             & 82            & --                & --              & 0.190             & 24.29             & 0.860             & 129           \\
			{}                                        & DBS~\cite{liu2025deformable}                  & 0.246                                     & 25.10                                    & 0.751                                     & 0.220                                & 32.29             & 0.936             & 0.234             & 123           & --                & --              & 0.140             & 24.85             & 0.870             & 150           \\
			{}                                        & 3DCS~\cite{held2025convex}                                & 0.238                                     & 24.07                                    & 0.700                                     & 0.166                                & 31.33             & 0.927             & 0.207             & 25            & 87                & 666             & 0.156             & 23.94             & 0.851             & 33            \\
			\midrule
			\multirow{4}{*}{\rotatebox{90}{Non-Vol.}} & {BBSplat~\cite{svitov2024bbsplat}}            & 0.281                                     & 23.55                                    & 0.669                                     & 0.178                                & 30.62             & 0.921             & 0.236             & 25            & 96                & 175             & 0.172             & \textbf{25.12}    & \textbf{0.868}    & 66            \\
			{}                                        & 2DGS~\cite{huang20242dgs}                                 & 0.246                                     & \textbf{24.34}                           & 0.717                                     & 0.195                                & 30.40             & 0.916             & 0.252             & 64            & 29                & 484             & 0.212             & 23.13             & 0.831             & 122           \\
			{}                                        & Triangle Splatting~\cite{held2025trianglesplatting}       & \underline{0.217}                         & 24.27                                    & \underline{0.722}                         & \underline{0.160}                    & \underline{30.80} & \underline{0.928} & \underline{0.191} & 97            & 39/25$^*$         & 795/813$^*$     & \underline{0.143} & 23.14             & 0.857             & 165           \\
			{}                                        & {DETRIS (Ours)}                                             & \textbf{0.206}                            & \underline{24.28}                        & \textbf{0.725}                            & \textbf{0.155}                       & \textbf{31.13}    & \textbf{0.930}    & \textbf{0.183}    & 43            & 55$^*$            & 946$^*$         & \textbf{0.133}    & \underline{23.31} & \underline{0.860} & 79            \\
			\bottomrule
		\end{tabular}
	}
\end{table}

\begin{figure*}[t!]
	\centering
	\newcommand{\fimgq}[1]{\includegraphics[width=\linewidth]{#1}}
	\newcommand{\rlblq}[1]{\rotatebox[origin=c]{90}{\tiny #1}}
	\setlength{\tabcolsep}{0.5pt}
	\renewcommand{\arraystretch}{0}
	\begin{tabular}{@{} >{\centering\arraybackslash}m{5mm} @{\,} >{\centering\arraybackslash}m{0.235\linewidth} @{\,} >{\centering\arraybackslash}m{0.235\linewidth} @{\,} >{\centering\arraybackslash}m{0.235\linewidth} @{\,} >{\centering\arraybackslash}m{0.235\linewidth} @{}}
		                                                      & \scriptsize GT & \scriptsize TS~\cite{held2025trianglesplatting} & \scriptsize 3DGS~\cite{kerbl20233dgs} & \scriptsize Ours \\[2pt]
		\rlblq{Counter}                                       &
		\fimgq{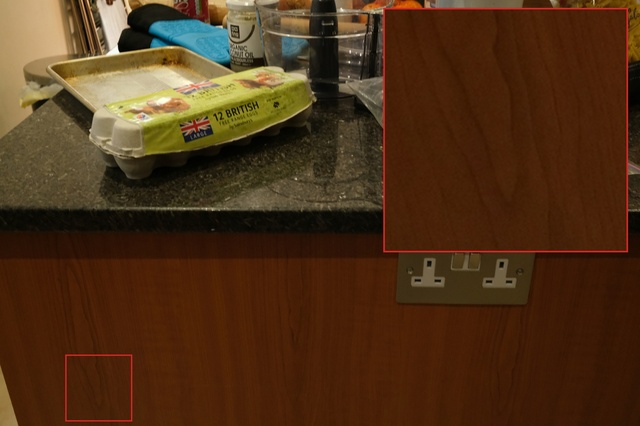}  &
		\fimgq{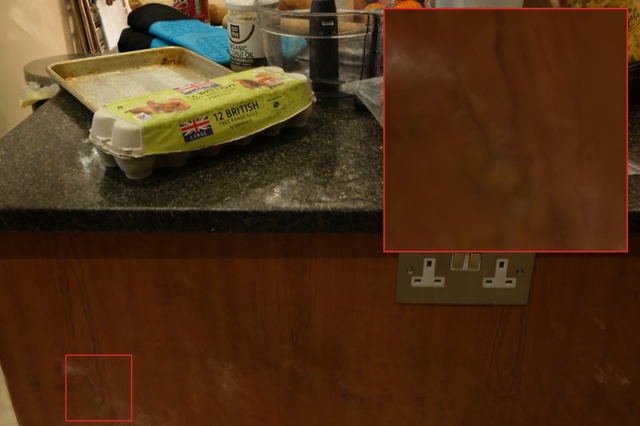}  &
		\fimgq{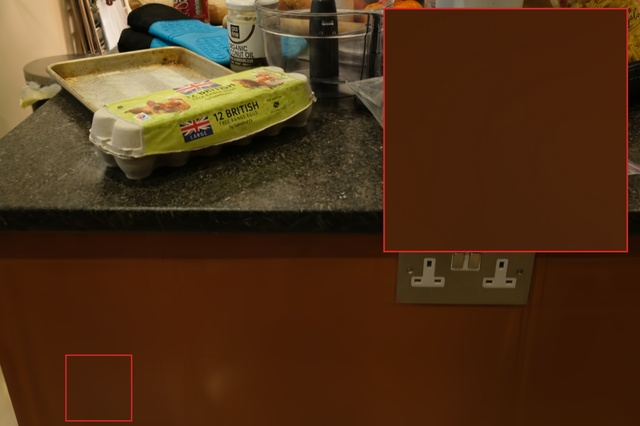}  &
		\fimgq{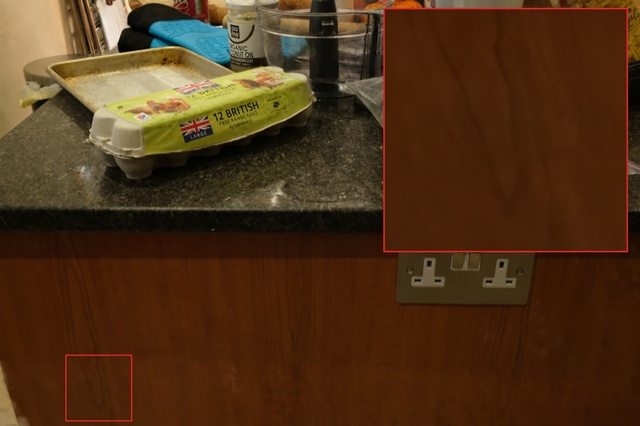}                                                                                                                              \\
		\rlblq{Stump}                                         &
		\fimgq{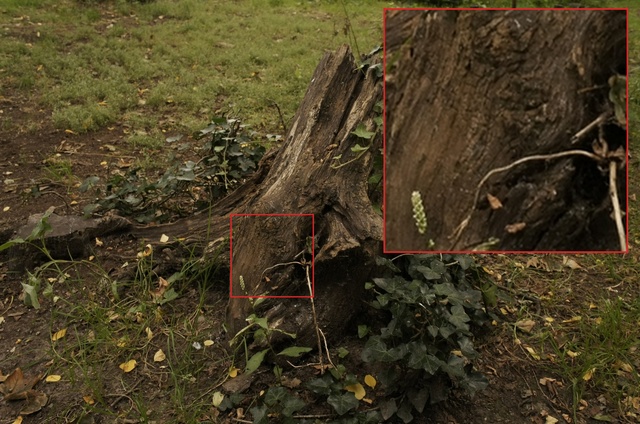}    &
		\fimgq{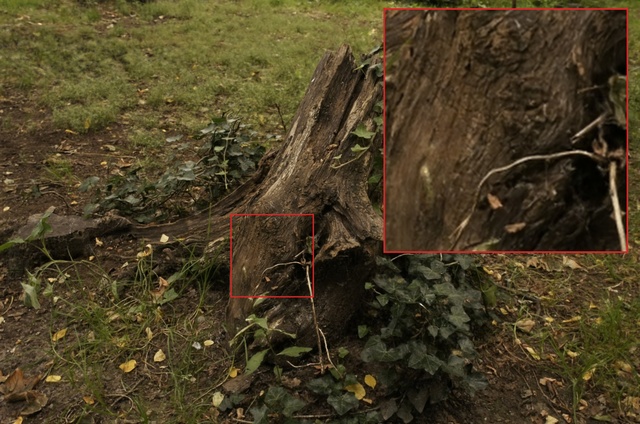}    &
		\fimgq{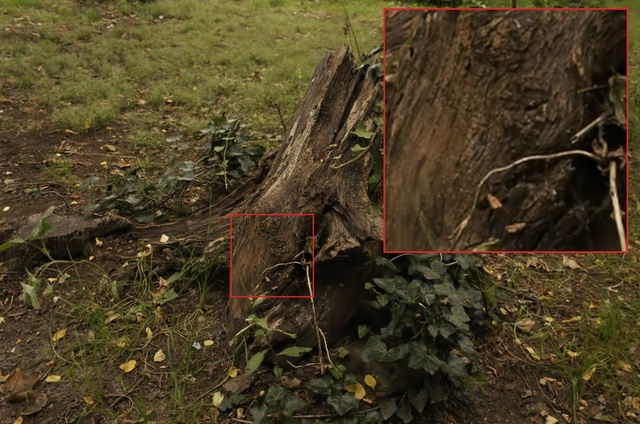}    &
		\fimgq{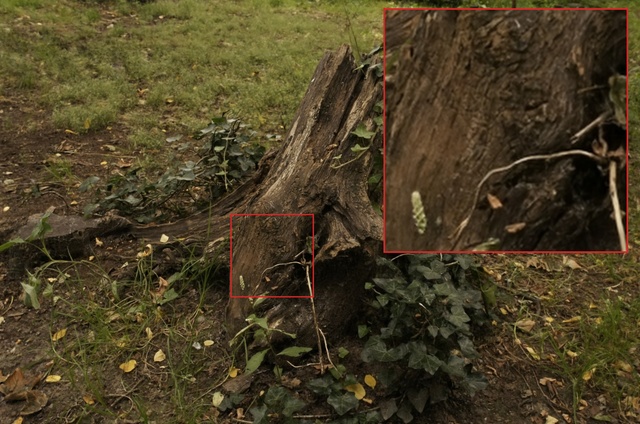}                                                                                                                                \\
		\rlblq{Bicycle}                                       &
		\fimgq{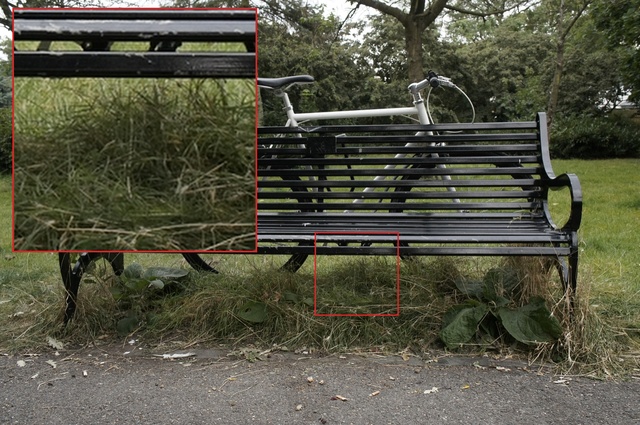}  &
		\fimgq{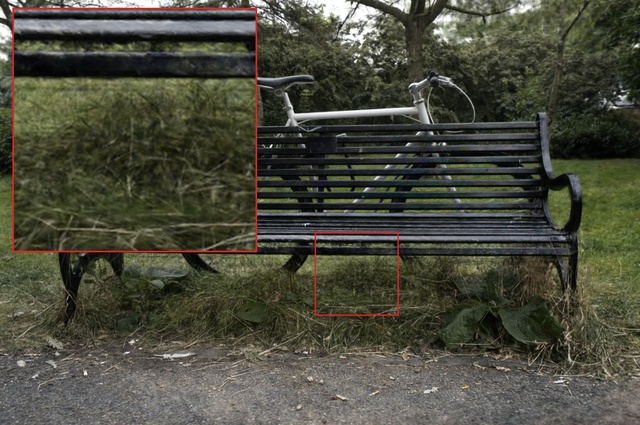}  &
		\fimgq{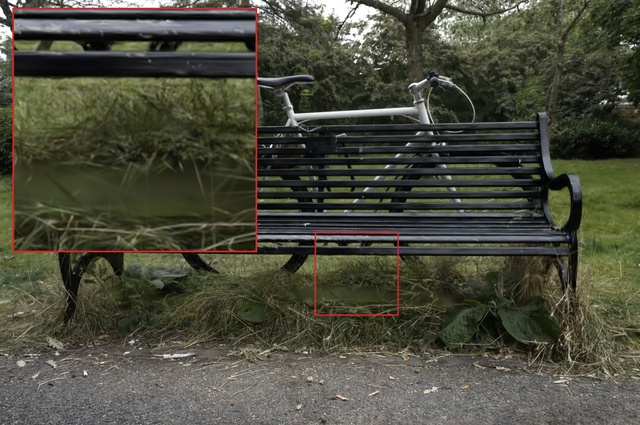}  &
		\fimgq{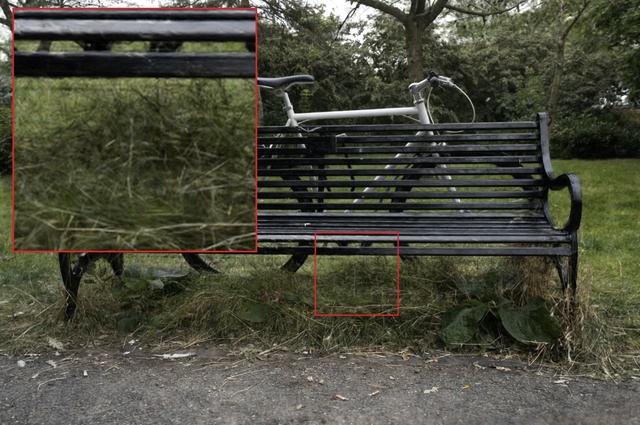}                                                                                                                              \\
		\rlblq{Treehill}                                      &
		\fimgq{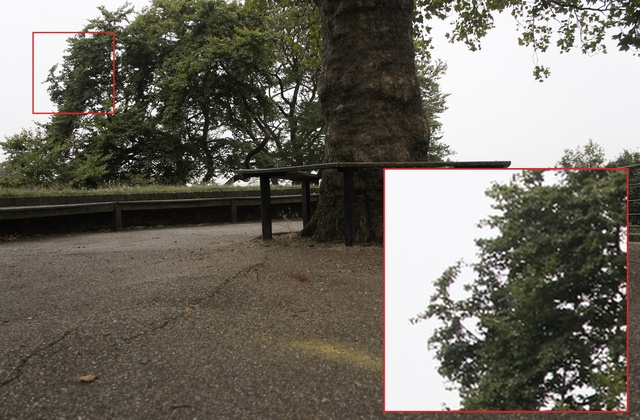} &
		\fimgq{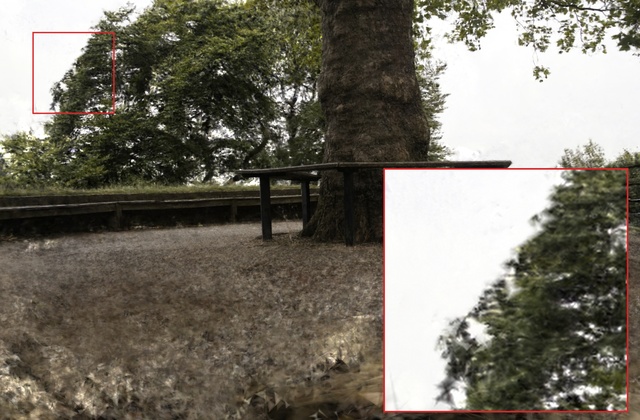} &
		\fimgq{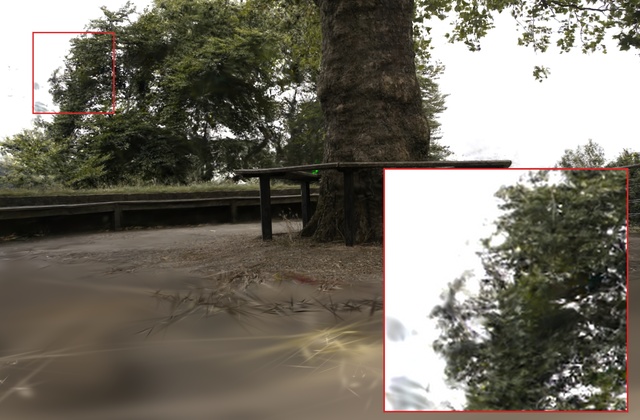} &
		\fimgq{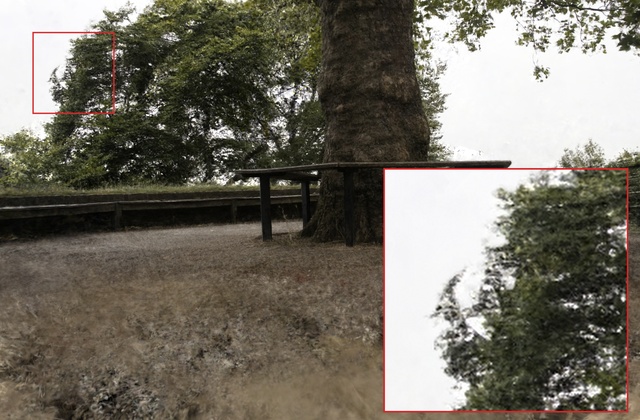}                                                                                                                             \\
	\end{tabular}
	\caption{\textbf{Qualitative comparison on real scenes.} Each row shows a test view from a different scene. Each column shows the ground truth, TS~\cite{held2025trianglesplatting}, 3DGS~\cite{kerbl20233dgs}, and ours.}
	\label{fig:qualitative}
\end{figure*}

Figure~\ref{fig:qualitative} compares rendered test views on four scenes spanning indoor and outdoor settings. Our deformable primitives better resolve fine spoke geometry (Bicycle), sharper object edges (Counter), and crisper foliage (Stump, Treehill), whereas TS~\cite{held2025trianglesplatting} exhibits tessellation artifacts and 3DGS~\cite{kerbl20233dgs} produces soft edges due to its Gaussian falloff. These gains are consistent with the per-scene LPIPS improvements provided in the supplementary.

\begin{figure*}[t!]
	\centering
	\newcommand{\fimgd}[1]{\includegraphics[width=\linewidth]{#1}}
	\newcommand{\rlbld}[1]{\rotatebox[origin=c]{90}{\tiny #1}}
	\setlength{\tabcolsep}{0.5pt}
	\renewcommand{\arraystretch}{0}
	\begin{tabular}{@{} >{\centering\arraybackslash}m{5mm} @{\,}
		>{\centering\arraybackslash}m{0.235\linewidth} @{\,}
		>{\centering\arraybackslash}m{0.235\linewidth} @{\,}
		>{\centering\arraybackslash}m{0.235\linewidth} @{\,}
		>{\centering\arraybackslash}m{0.235\linewidth} @{}}
		                                                           & \scriptsize GT & \scriptsize Full ($K{=}3$) & \scriptsize Rigid ($d{=}0$) & \scriptsize $|\text{Full}{-}\text{Rigid}|$ \\[2pt]
		\rlbld{Counter}                                                          &
		\fimgd{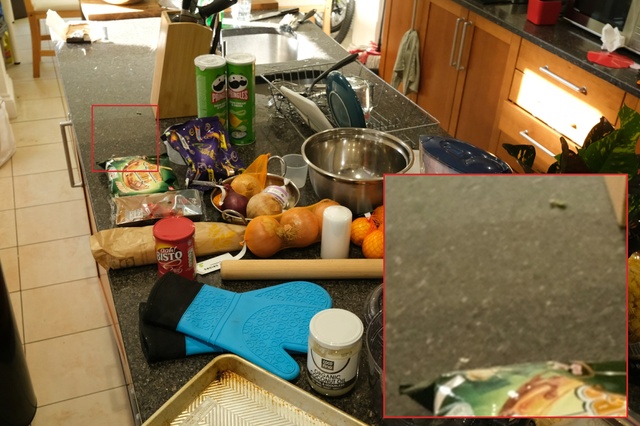}    &
		\fimgd{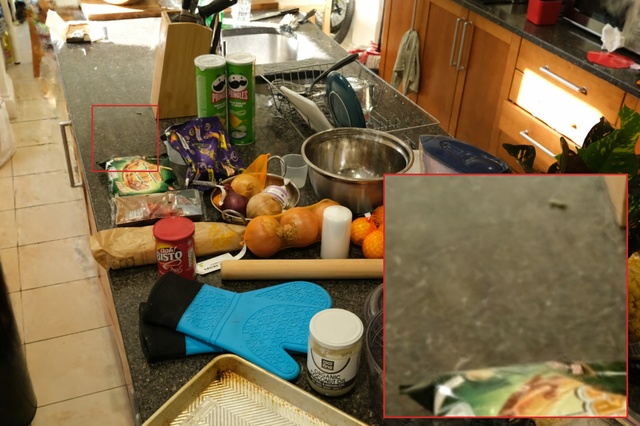}  &
		\fimgd{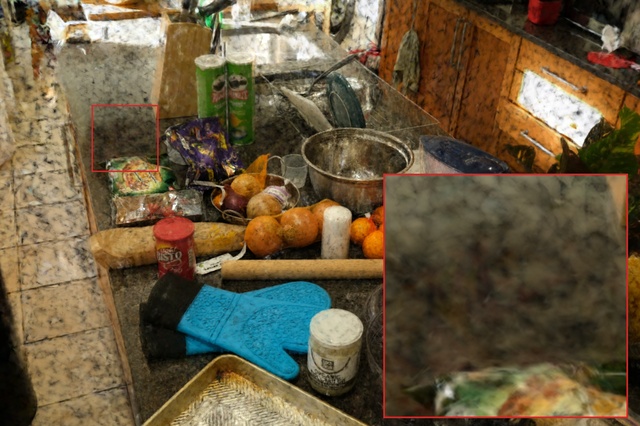} &
		\fimgd{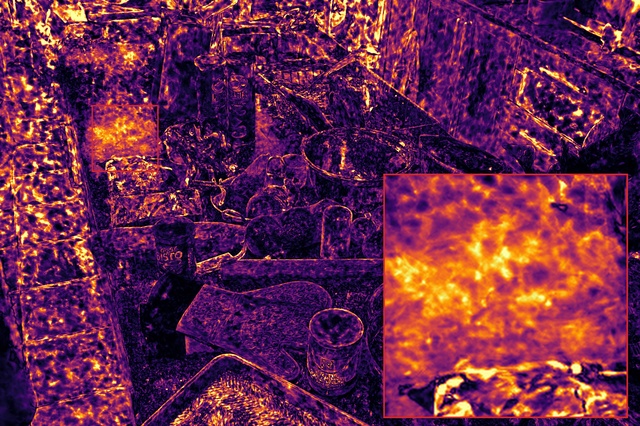}                                                                                                          \\[2pt]
		\rlbld{Truck}                                                            &
		\fimgd{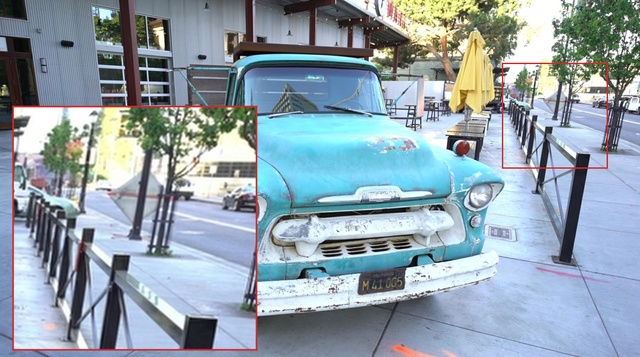}      &
		\fimgd{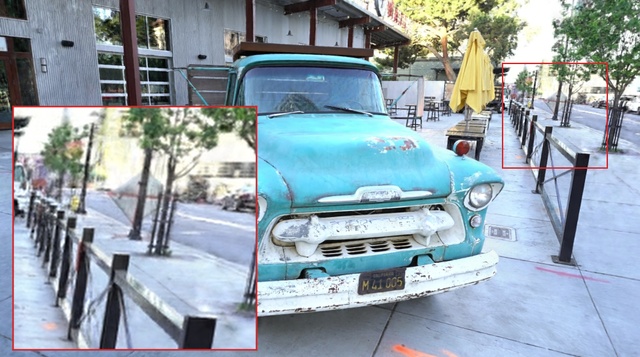}    &
		\fimgd{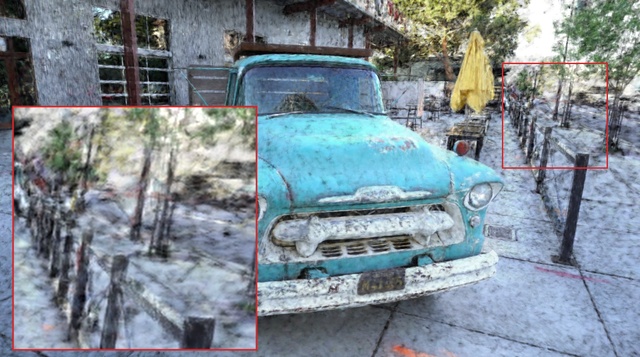}   &
		\fimgd{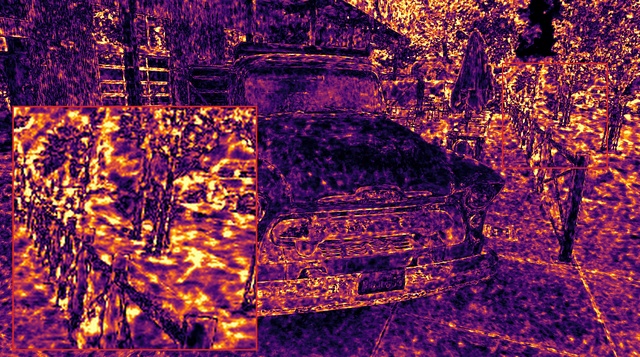}                                                                                                           \\
	\end{tabular}
	\caption{\textbf{Deformation contribution maps.} For each scene we render the trained model with learned control-point displacements active (\emph{Full}) and with all displacements and corner smoothness $\delta$ zeroed (\emph{Rigid}, $d{=}0$), keeping all other parameters unchanged. The rightmost column shows the per-pixel absolute difference $|\text{Full}{-}\text{Rigid}|$ (inferno colormap, percentile-normalized), isolating the visual contribution of the learned boundary deformations from vertex positions, opacities, and colors. Red insets highlight the patch of highest deformation activity. The effect concentrates at fine boundary structures---thin object edges in Counter (top), cabin edges and wheel wells in Truck (bottom)---the same regions that drive our LPIPS improvement over TS~\cite{held2025trianglesplatting}.}
	\label{fig:deformation_contribution}
\end{figure*}
To understand \emph{where} the learned deformations contribute, Figure~\ref{fig:deformation_contribution} compares each model rendered with and without control-point displacements (see caption for more details). The deformation signal concentrates at fine boundary structures, such as thin object edges in Counter and cabin edges and wheel wells in Truck, precisely where LPIPS improves most over TS~\cite{held2025trianglesplatting}. Quantitatively, zeroing the control points increases the mean absolute error vs.\ ground truth by $+0.052$ on Bonsai and $+0.029$ on Bicycle (shown in the supplementary), confirming that the learned deformations reduce reconstruction error at boundaries.

Table~\ref{tab:main_results} also reports rendering speed, training time, and model size. The additional per-pixel distance computations for $M$ boundary segments reduce FPS compared to TS~\cite{held2025trianglesplatting}, though our throughput remains comparable to 3DCS~\cite{held2025convex} and BBSplat~\cite{svitov2024bbsplat}; hierarchical culling or approximate distance fields could close this gap. Training takes roughly $2{\times}$ longer than TS~\cite{held2025trianglesplatting} on the same hardware (NVIDIA RTX\,5090), a trade-off we consider justified by the consistent quality gains. Regarding model size, each deformable primitive carries $3K{+}1{=}10$ extra floats (9~CP displacements $+$ 1~corner smoothness $\delta$), a $+17\%$ per-primitive overhead. Since the optimizer converges to a comparable primitive count, the net storage increase comes almost entirely from these extra parameters rather than primitive count.

\subsection{Shape Expressiveness}
\begin{figure*}[ht]
	\centering
	\newcommand{\fimgs}[1]{\includegraphics[width=\linewidth]{#1}}
	\newcommand{\rlbls}[1]{\makebox[0pt]{\raisebox{0.75pt}{\smash{\rotatebox[origin=c]{90}{\scalebox{0.65}{\tiny #1}}}}}}
	\newcommand{\clbl}[1]{\scalebox{0.7}{\tiny #1}}
	\newcommand{\mlbl}[1]{\smash{\raisebox{1.5pt}{\scalebox{0.7}{\tiny #1}}}}
	\setlength{\tabcolsep}{0.3pt}
	\renewcommand{\arraystretch}{0}
	\begin{minipage}[t]{0.49\linewidth}
		\centering
		\begin{tabular}{@{}
			>{\centering\arraybackslash}m{2.5mm} @{\,}
			>{\centering\arraybackslash}m{0.112\linewidth} @{\,}
			>{\centering\arraybackslash}m{0.112\linewidth} @{\,}
			>{\centering\arraybackslash}m{0.112\linewidth} @{\,}
			>{\centering\arraybackslash}m{0.112\linewidth} @{\,}
			>{\centering\arraybackslash}m{0.112\linewidth} @{\,}
			>{\centering\arraybackslash}m{0.112\linewidth} @{\,}
			>{\centering\arraybackslash}m{0.112\linewidth} @{}}
			                                                        &               & \multicolumn{2}{c}{\mlbl{Ours}} & \multicolumn{2}{c}{\mlbl{TS~\cite{held2025trianglesplatting}}} & \multicolumn{2}{c}{\mlbl{3DGS~\cite{kerbl20233dgs}}}                                                    \\[0pt]
			                                                        & \clbl{Target} & \clbl{$K{=}3$}                  & \clbl{$K{=}7$}                                                 & \clbl{$N{=}1$}                                       & \clbl{$N{=}3$} & \clbl{$N{=}1$} & \clbl{$N{=}8$} \\[1pt]
			\rlbls{Circle}                                          &
			\fimgs{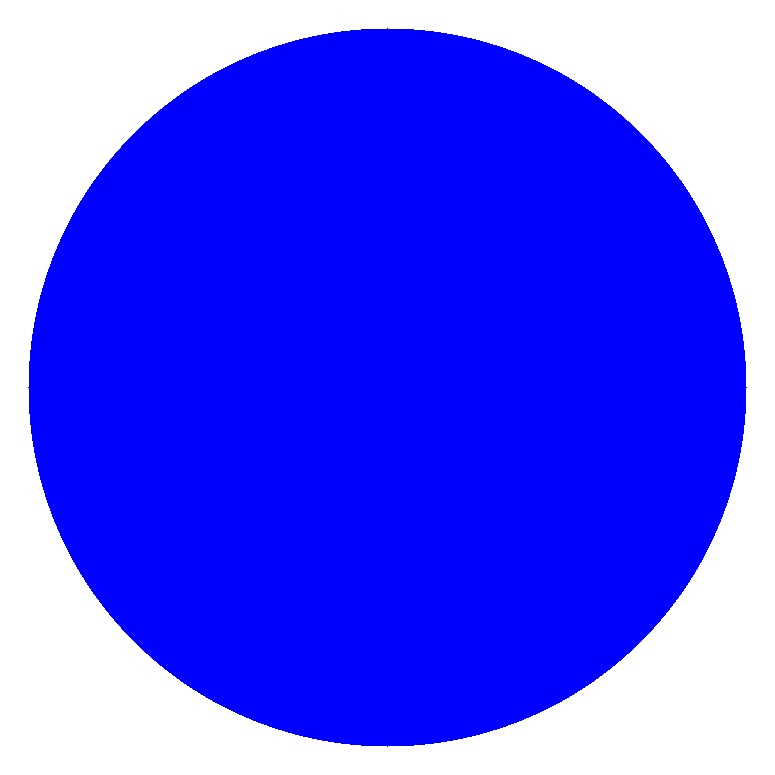}           &
			\fimgs{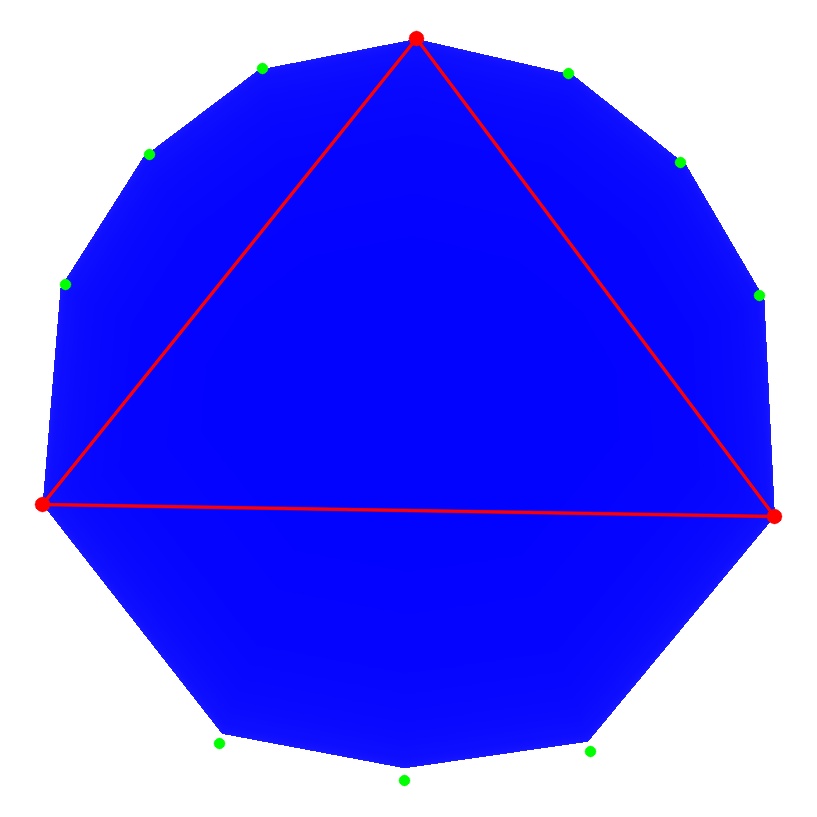}          &
			\fimgs{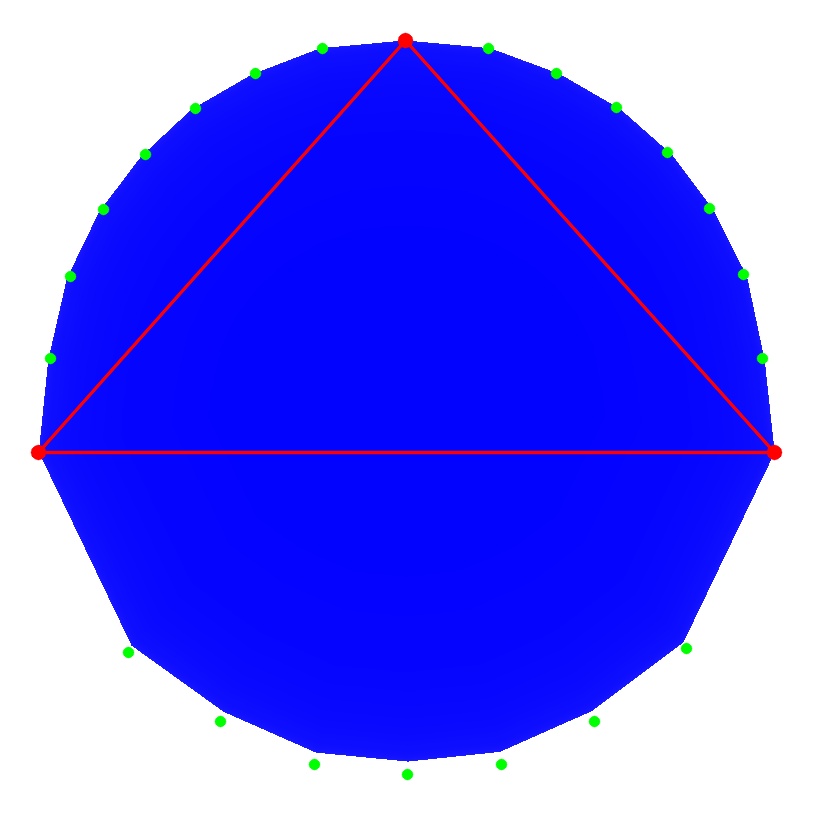}          &
			\fimgs{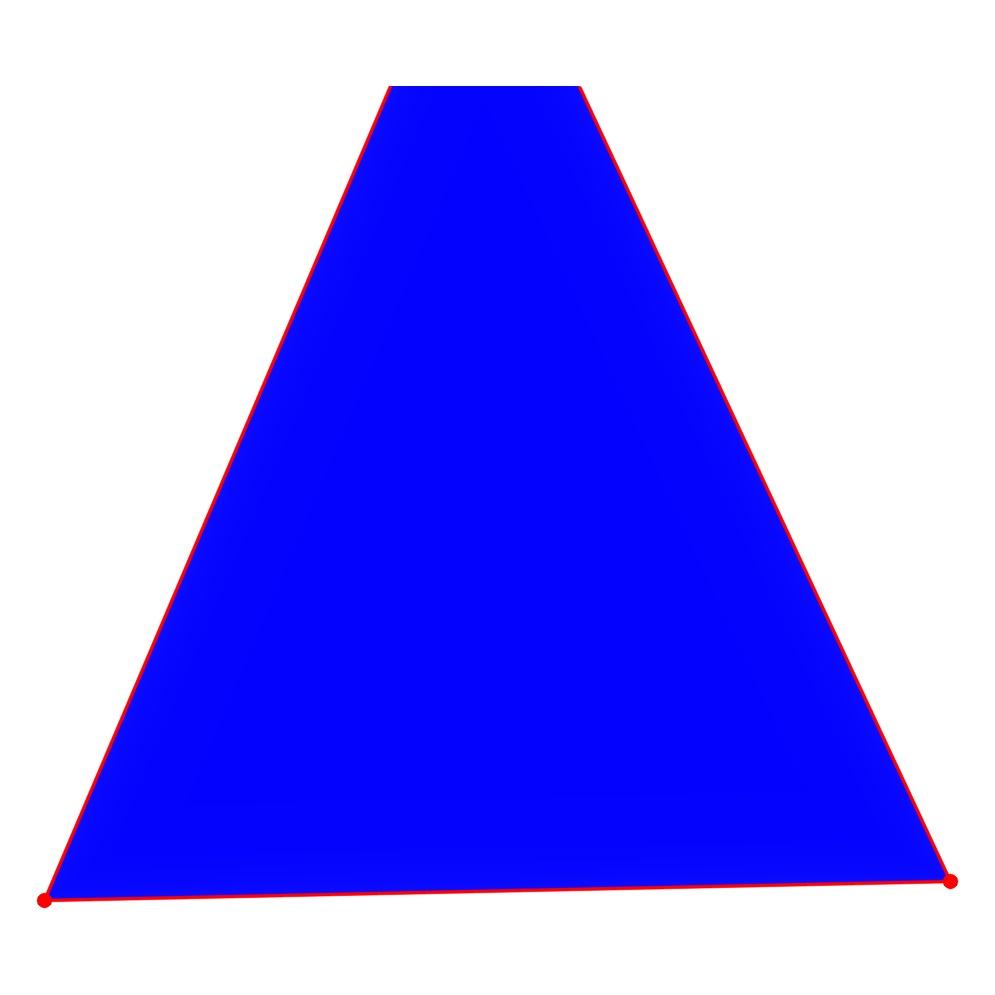}             &
			\fimgs{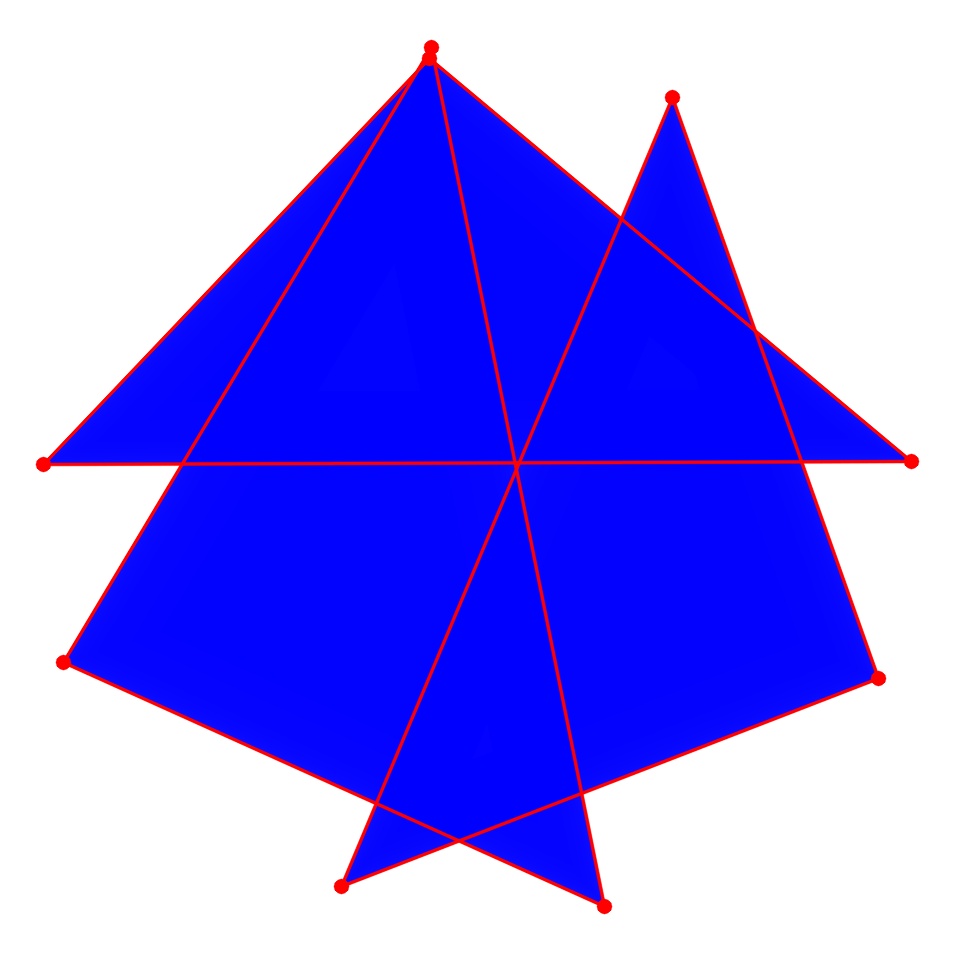}             &
			\fimgs{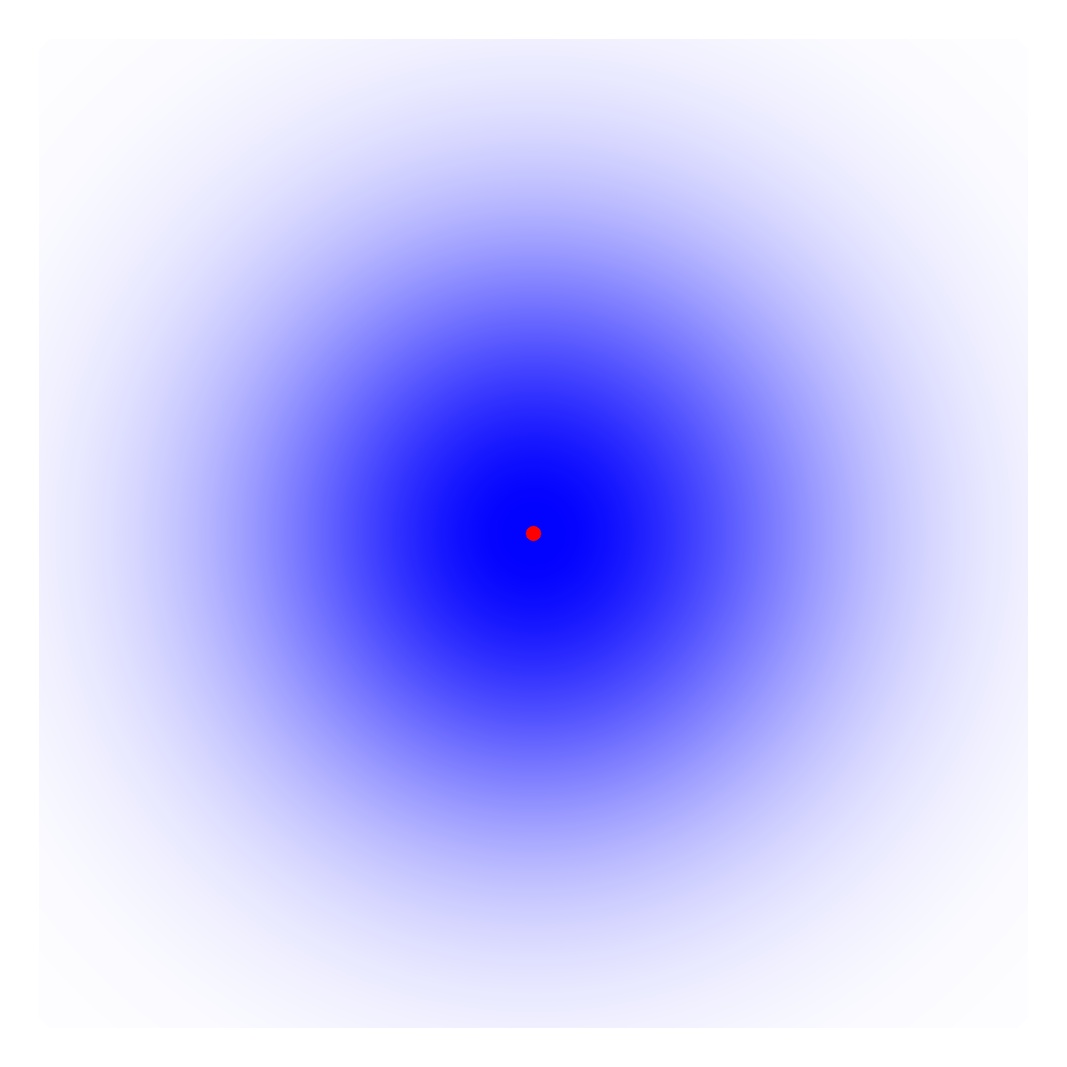}             &
			\fimgs{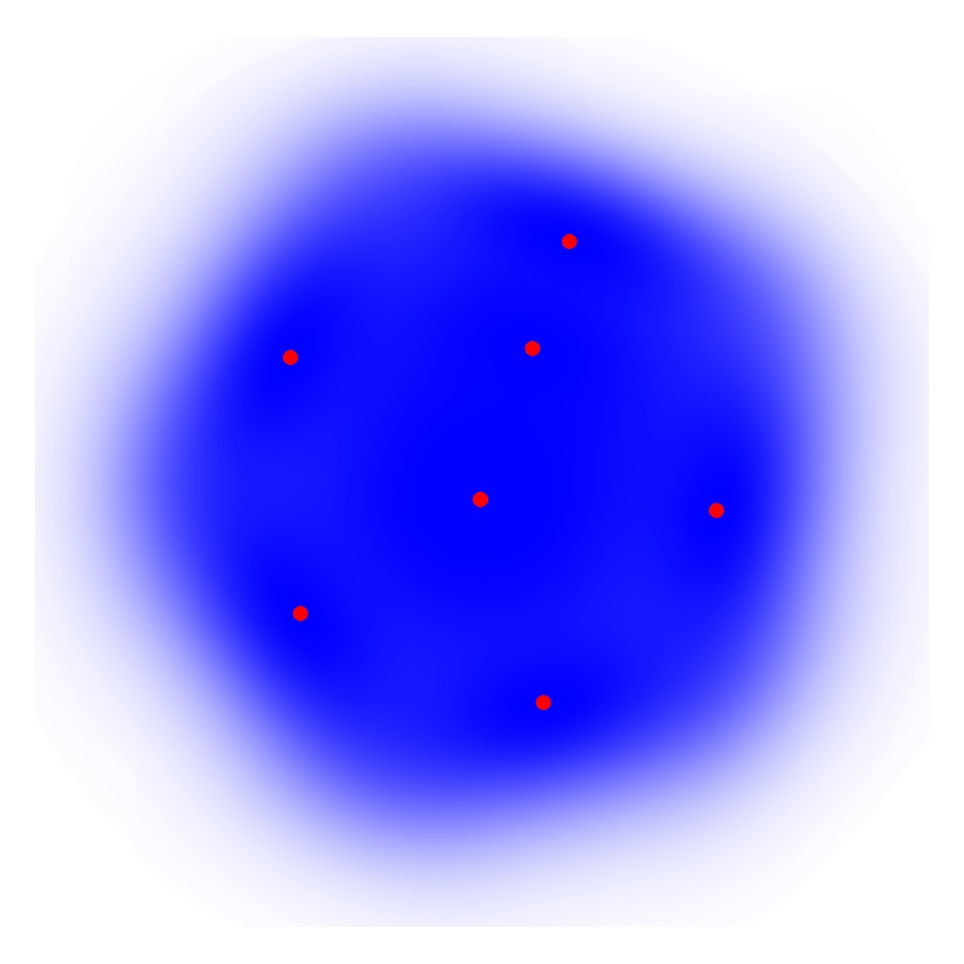}                                                                                                                                                                                                                                          \\
			\rlbls{Triangle}                                        &
			\fimgs{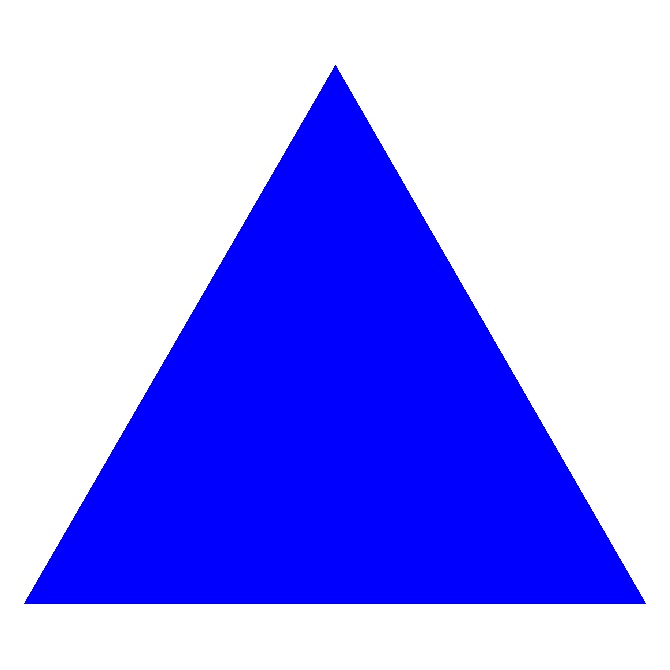}         &
			\fimgs{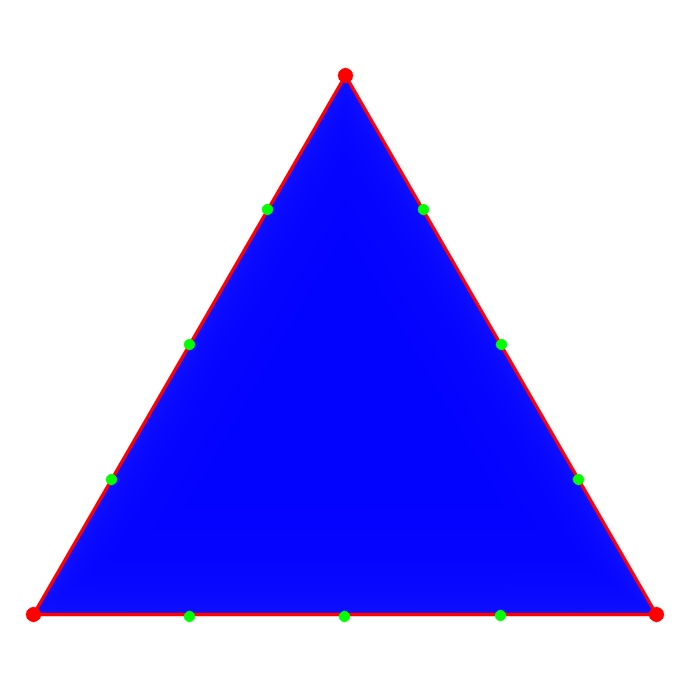}        &
			\fimgs{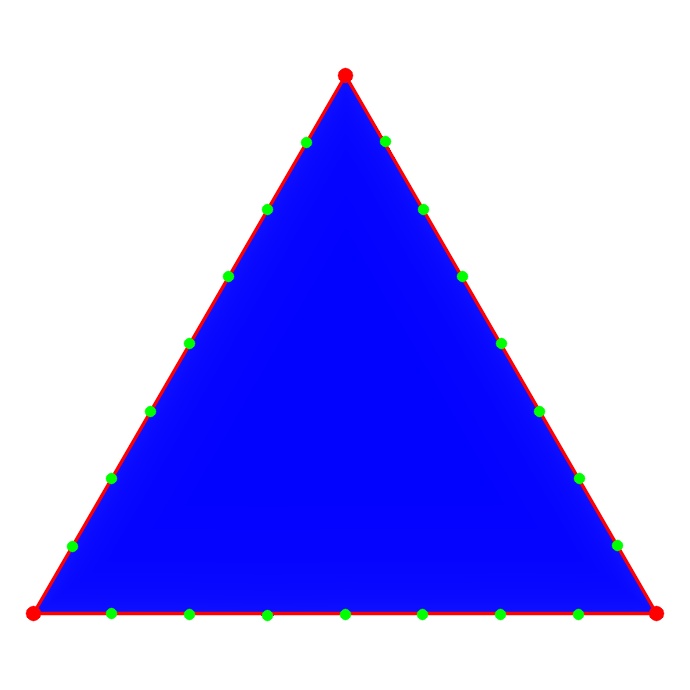}        &
			\fimgs{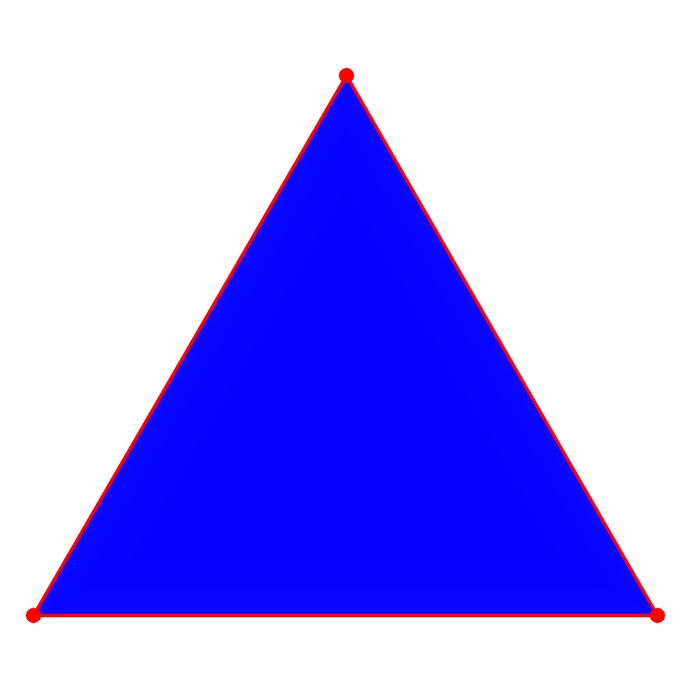}           &
			\fimgs{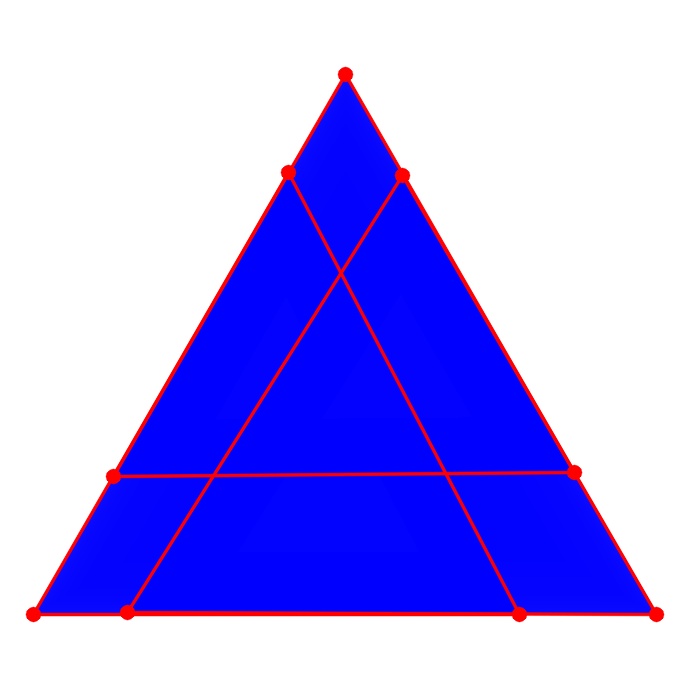}           &
			\fimgs{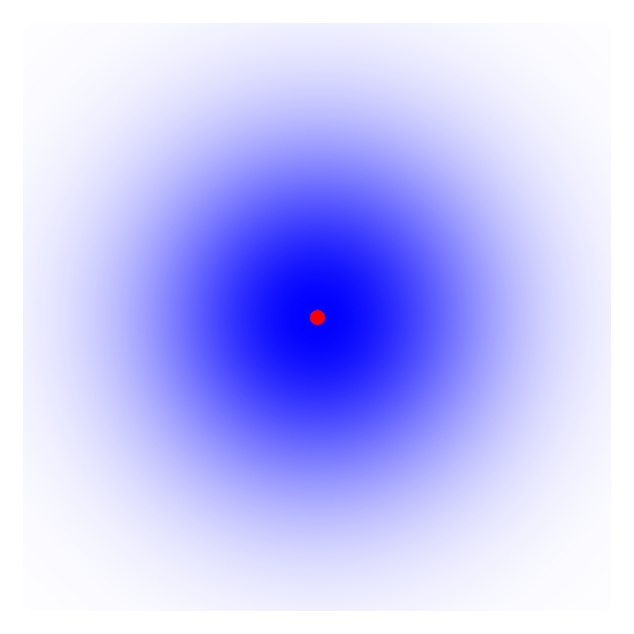}           &
			\fimgs{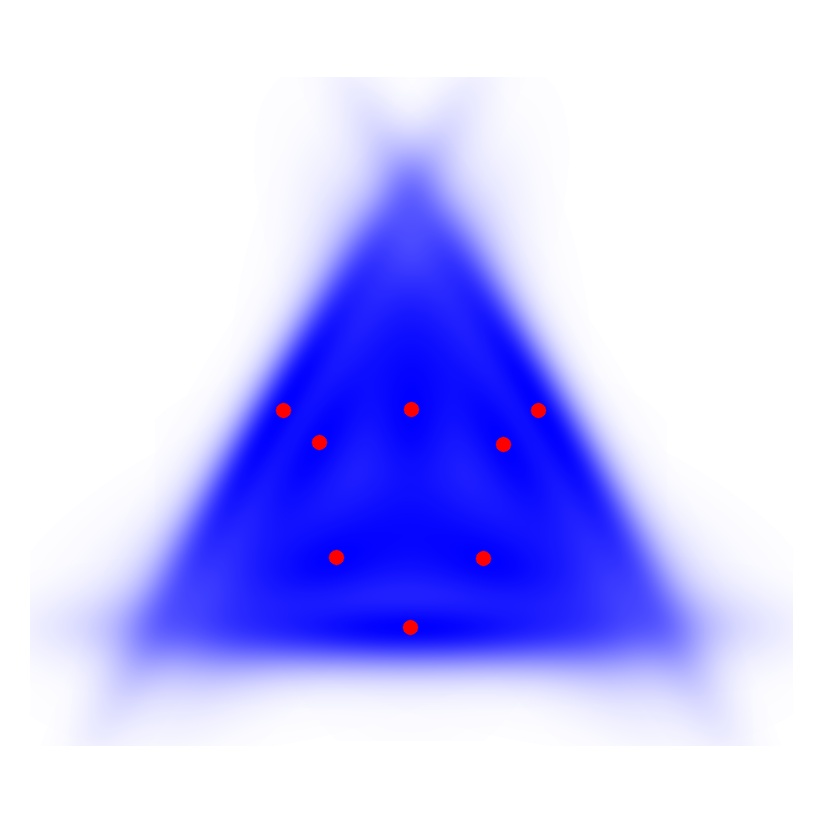}                                                                                                                                                                                                                                        \\
			\rlbls{Square}                                          &
			\fimgs{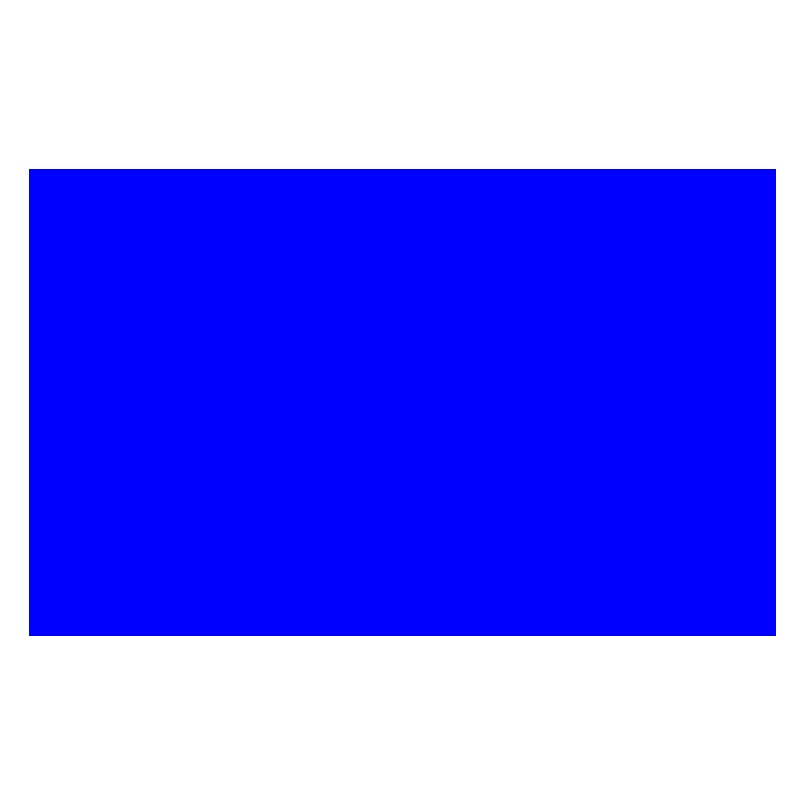}           &
			\fimgs{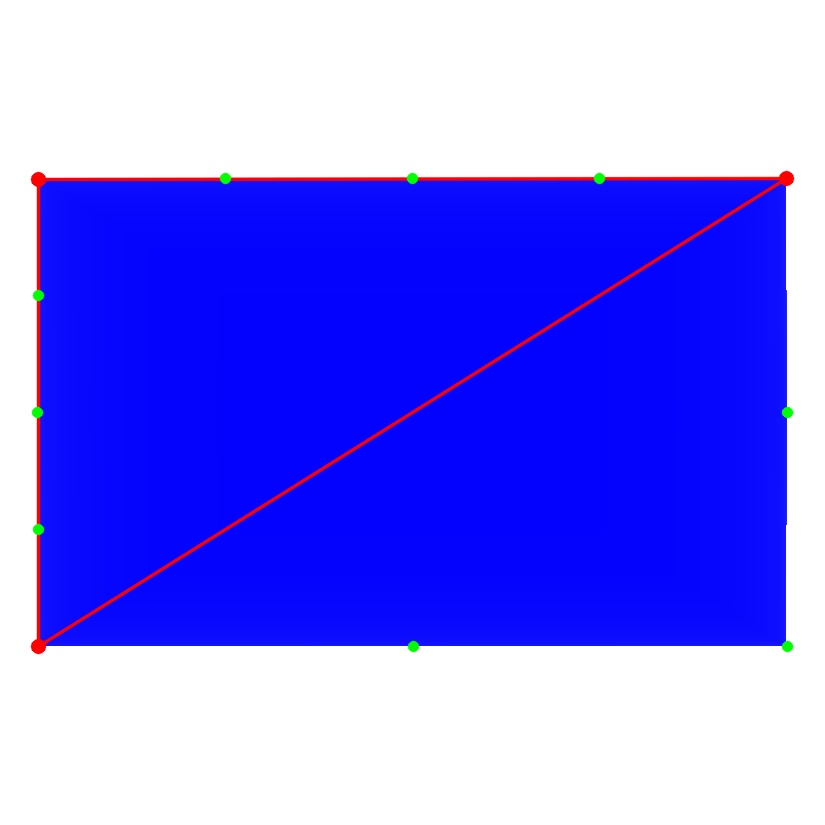}          &
			\fimgs{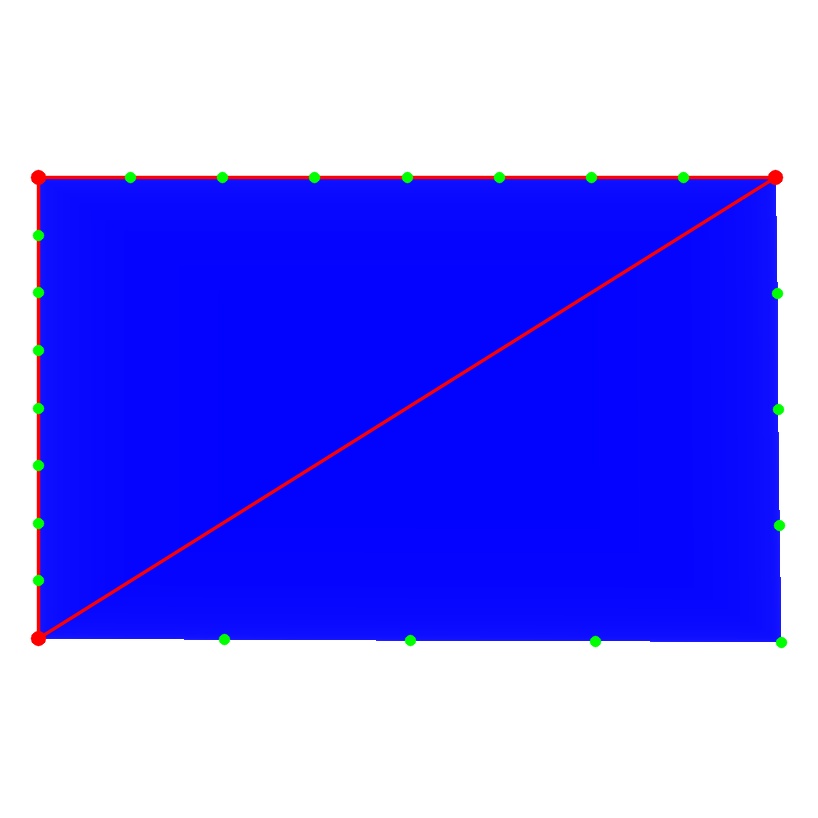}          &
			\fimgs{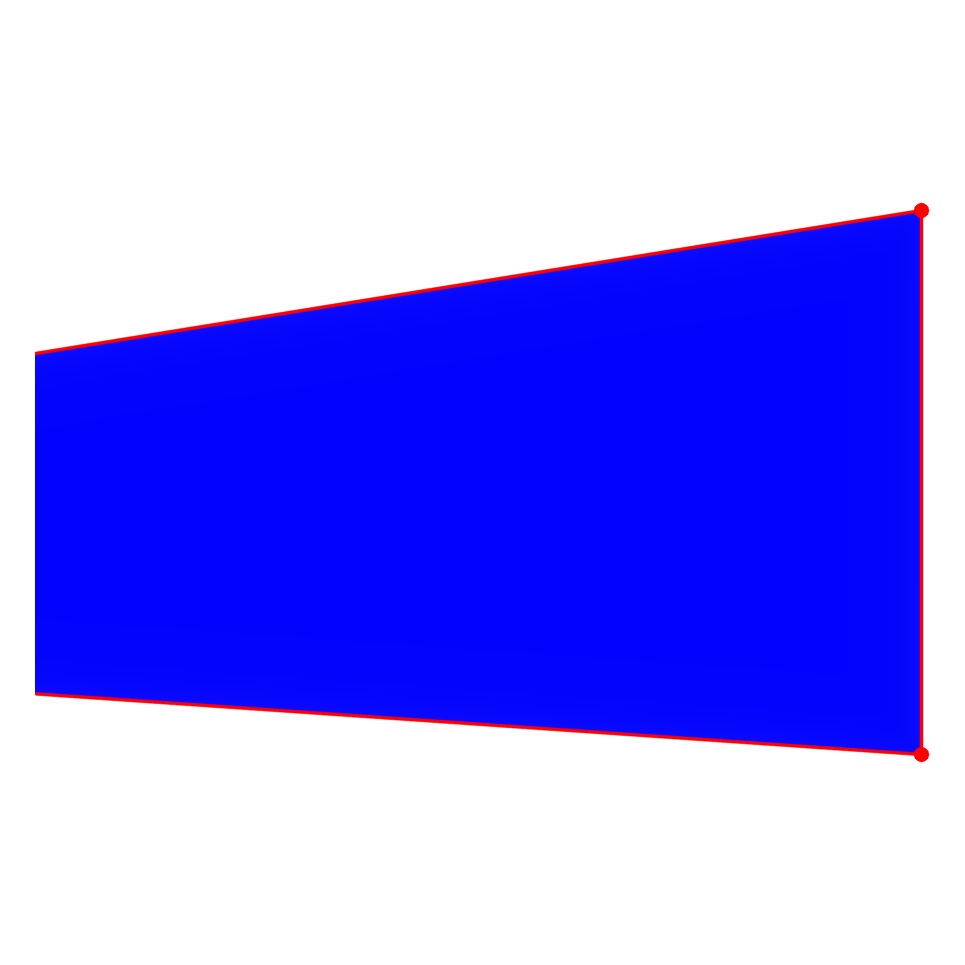}             &
			\fimgs{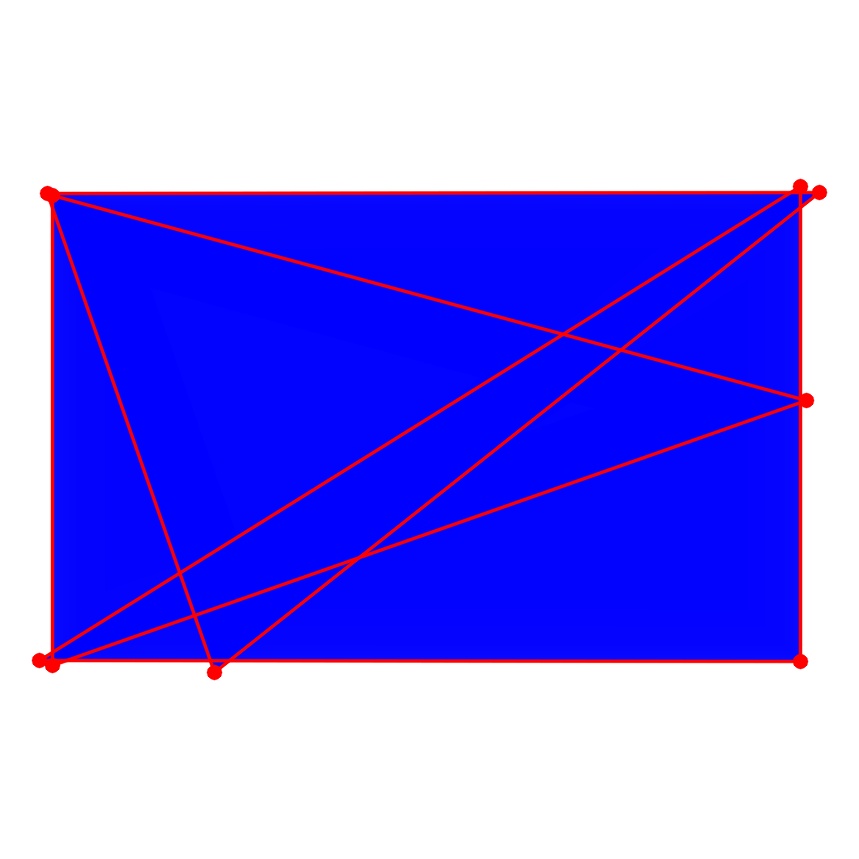}             &
			\fimgs{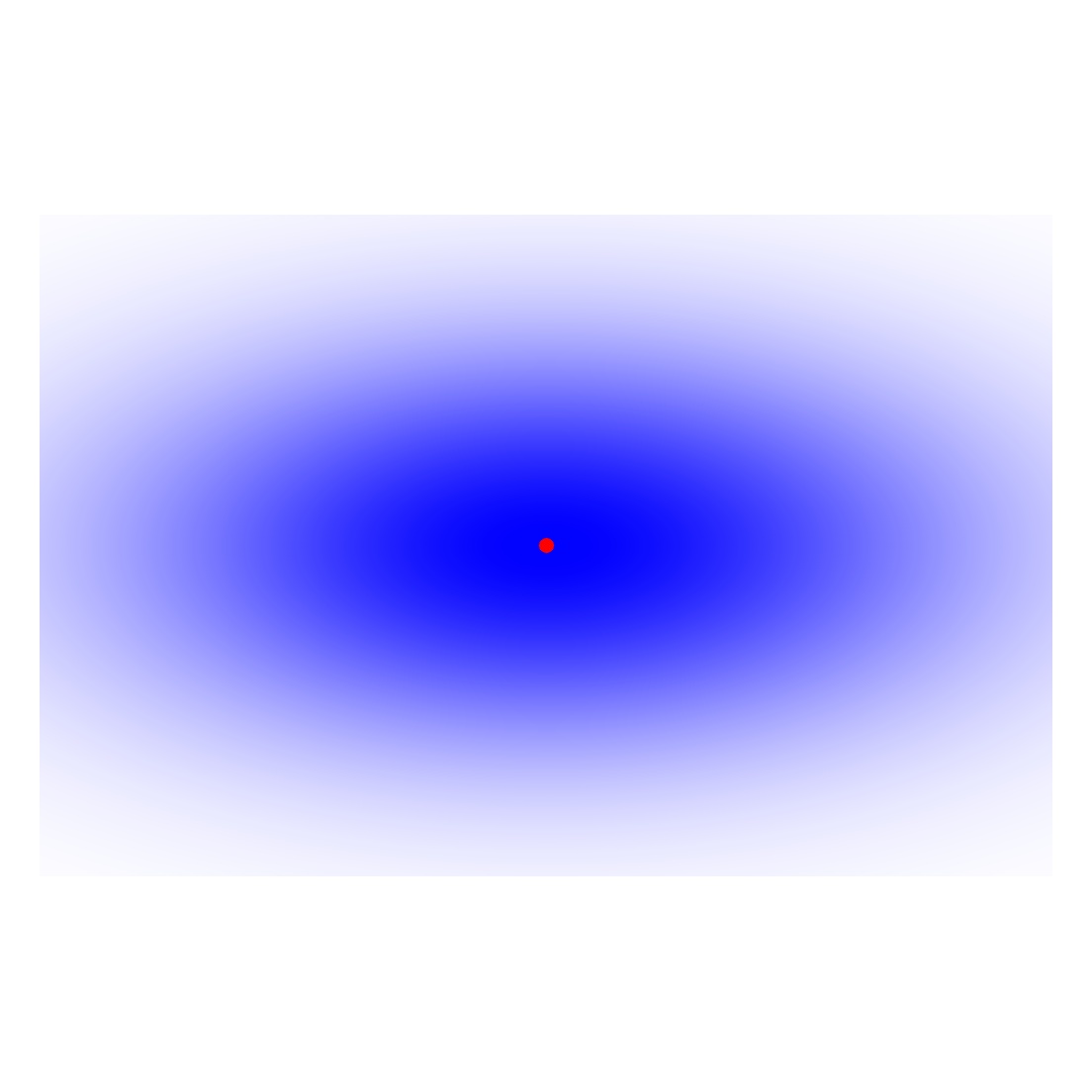}             &
			\fimgs{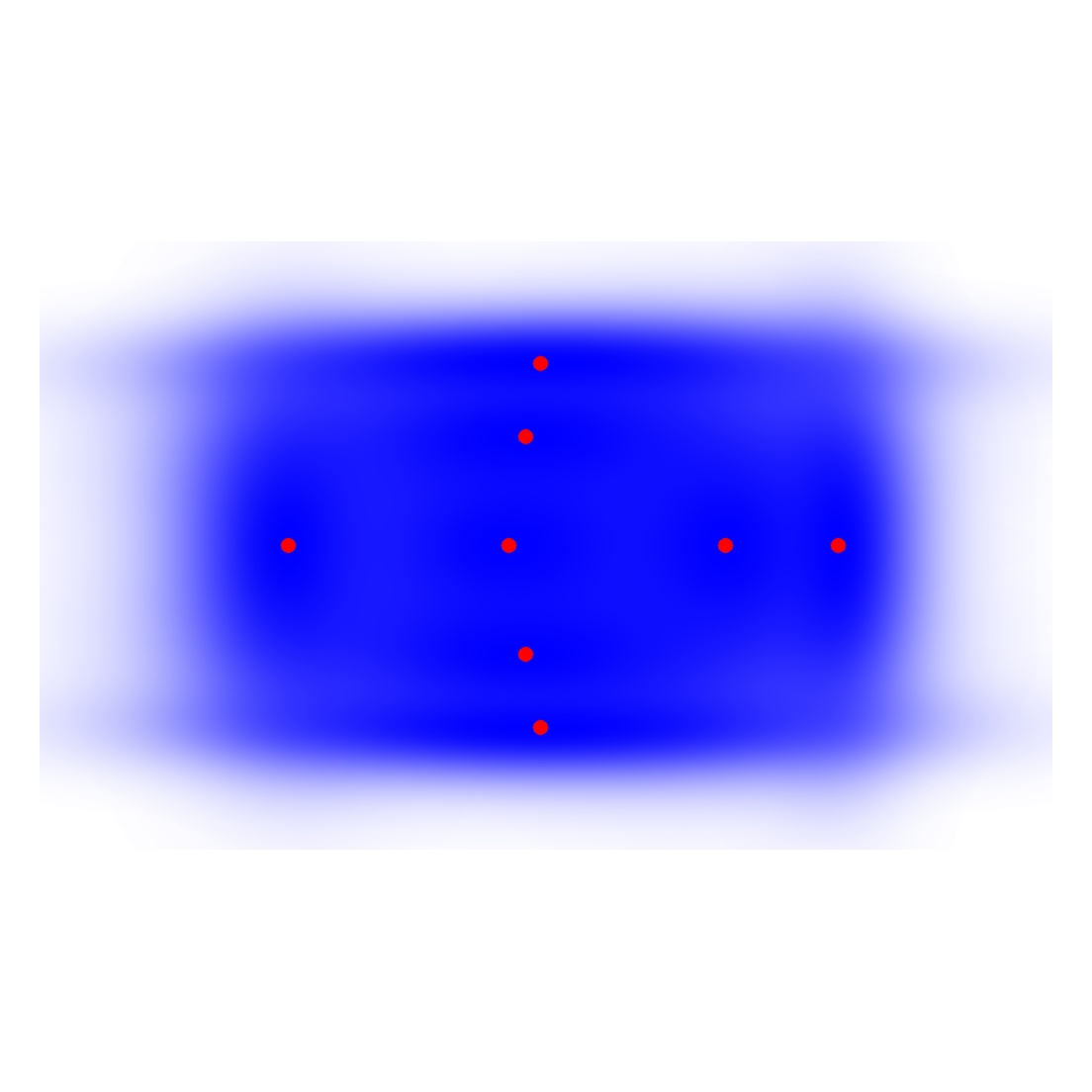}                                                                                                                                                                                                                                          \\
			\rlbls{Gaussian}                                        &
			\fimgs{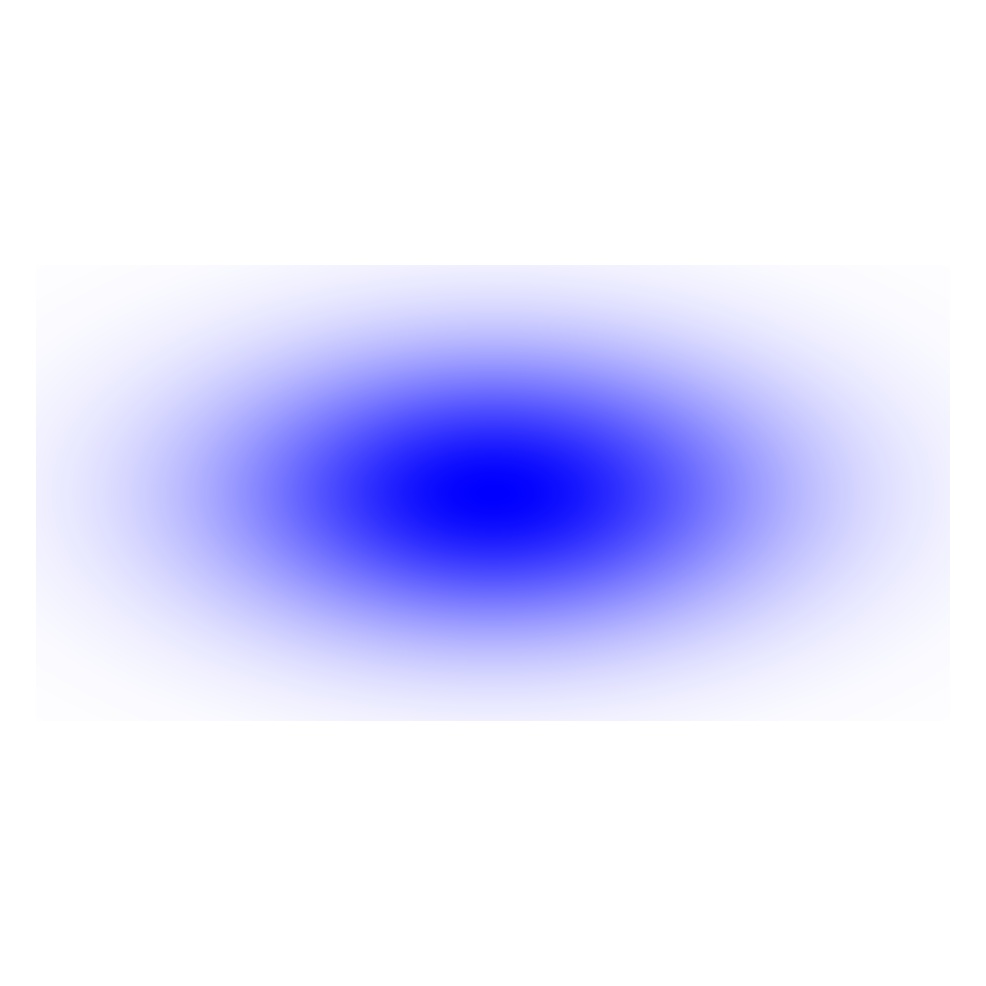}  &
			\fimgs{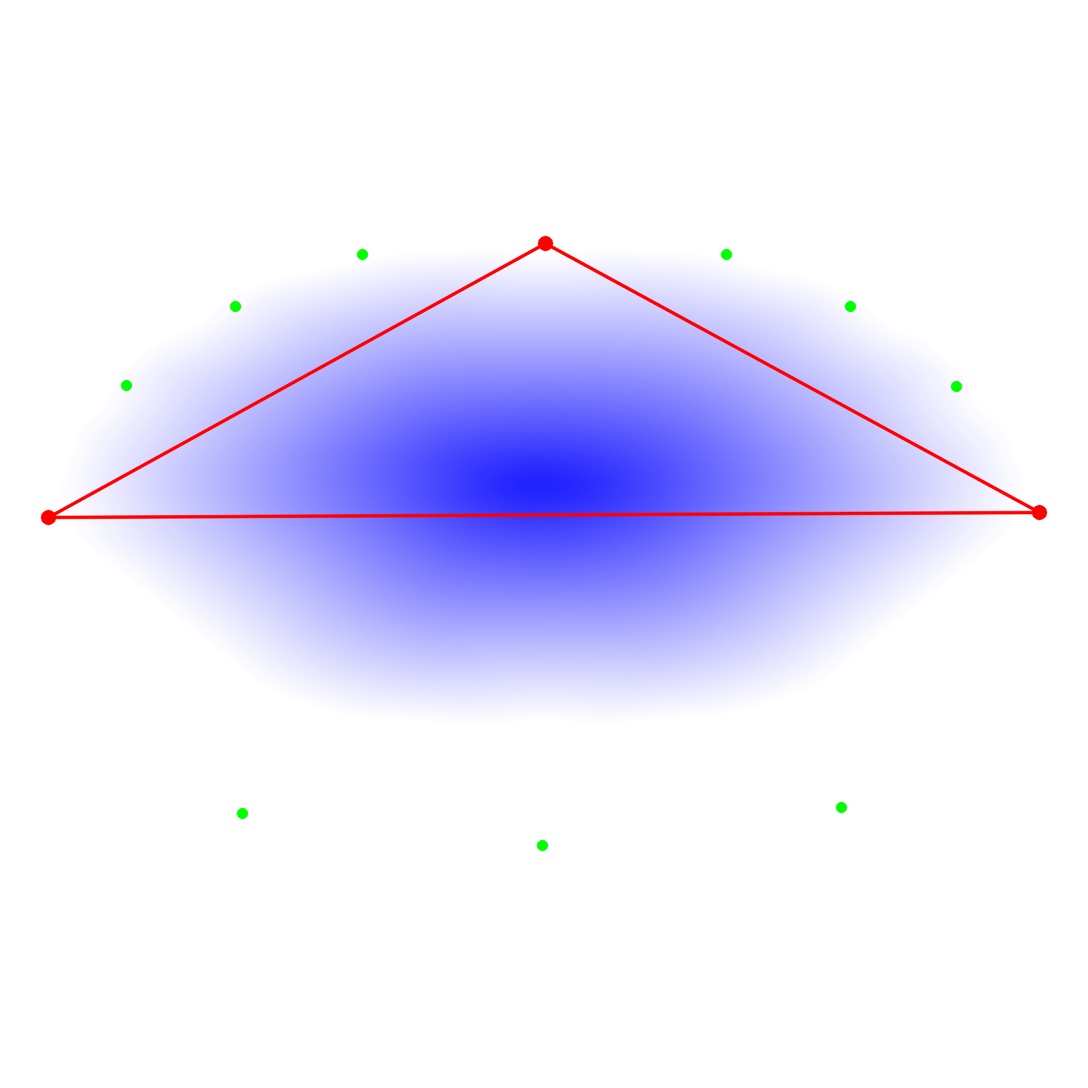} &
			\fimgs{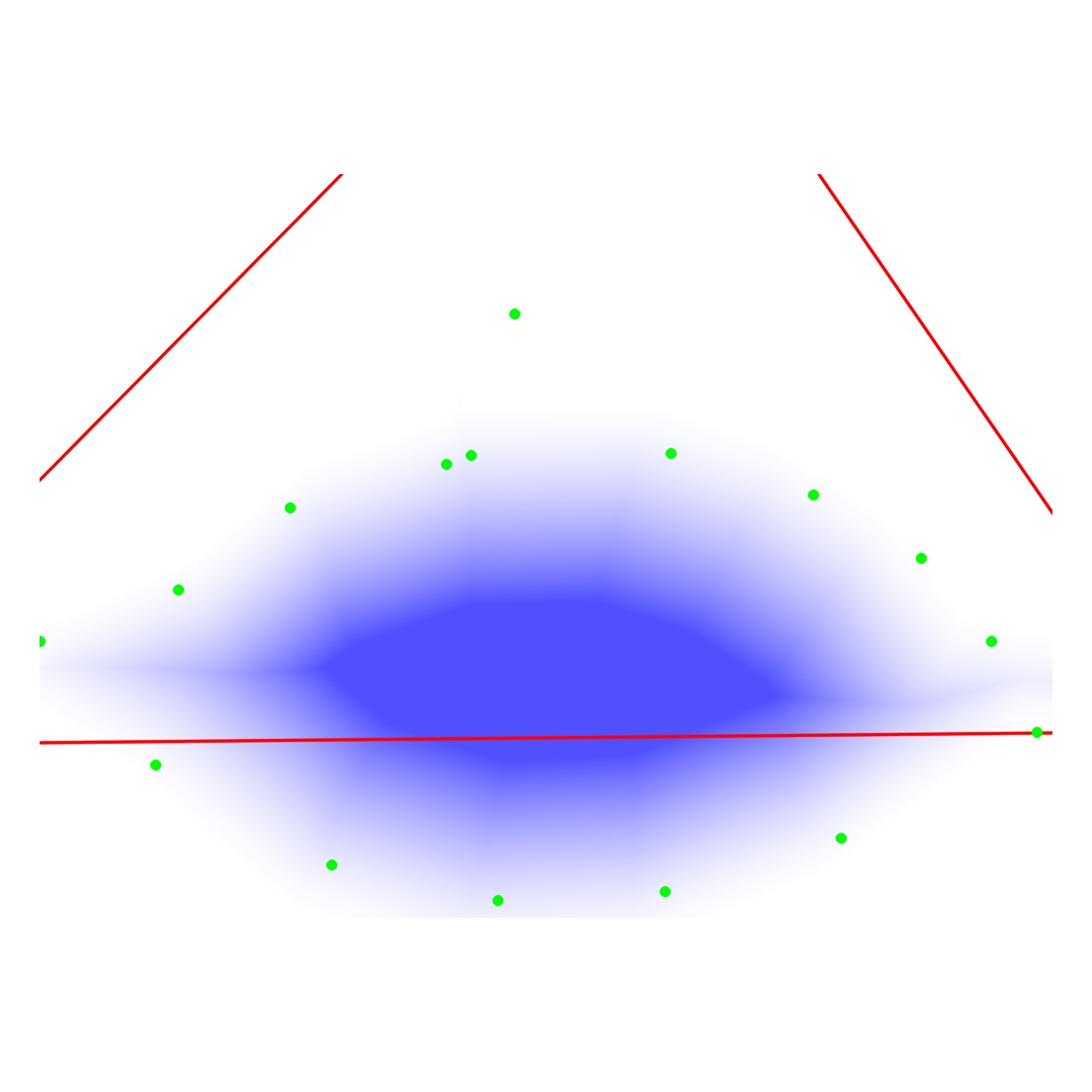} &
			\fimgs{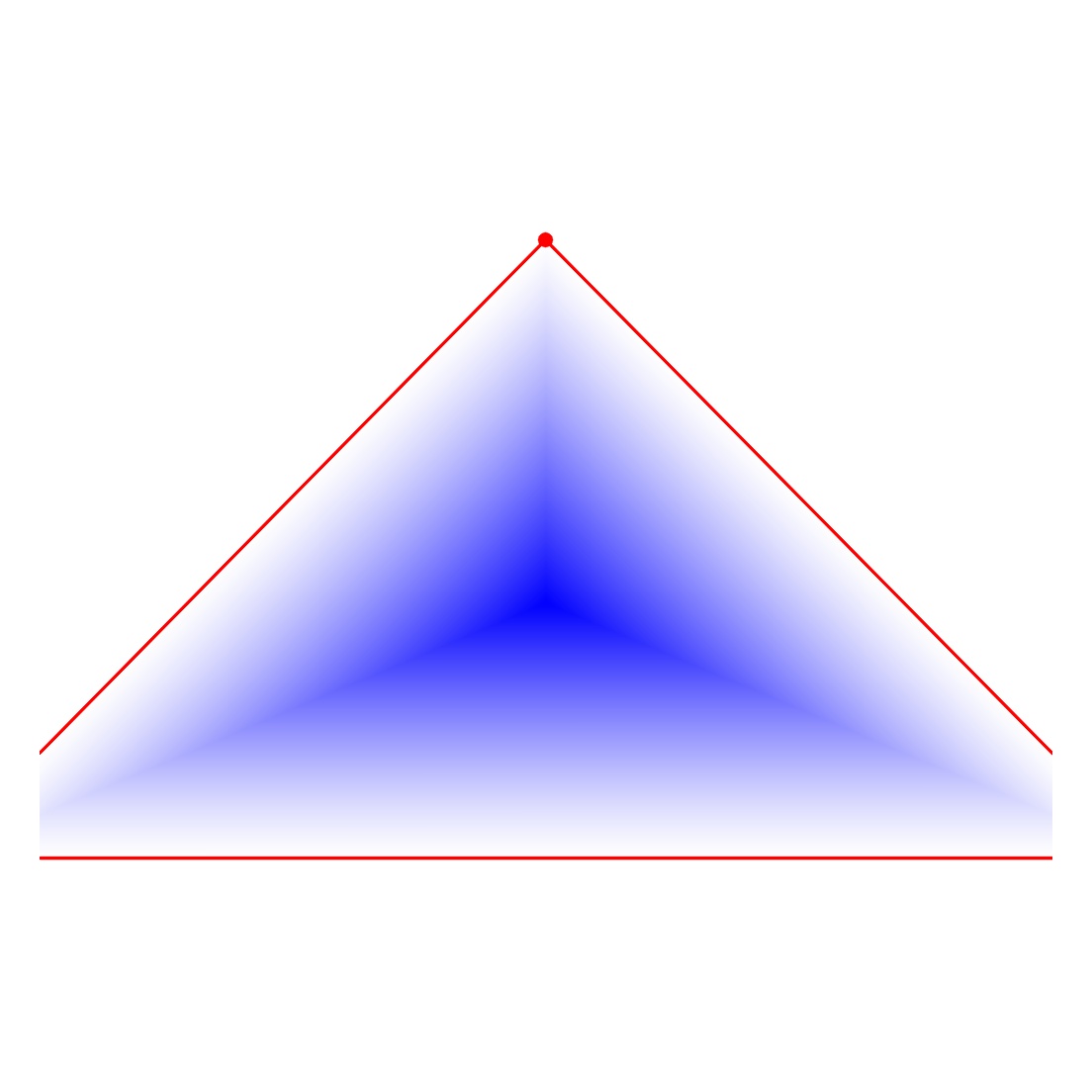}    &
			\fimgs{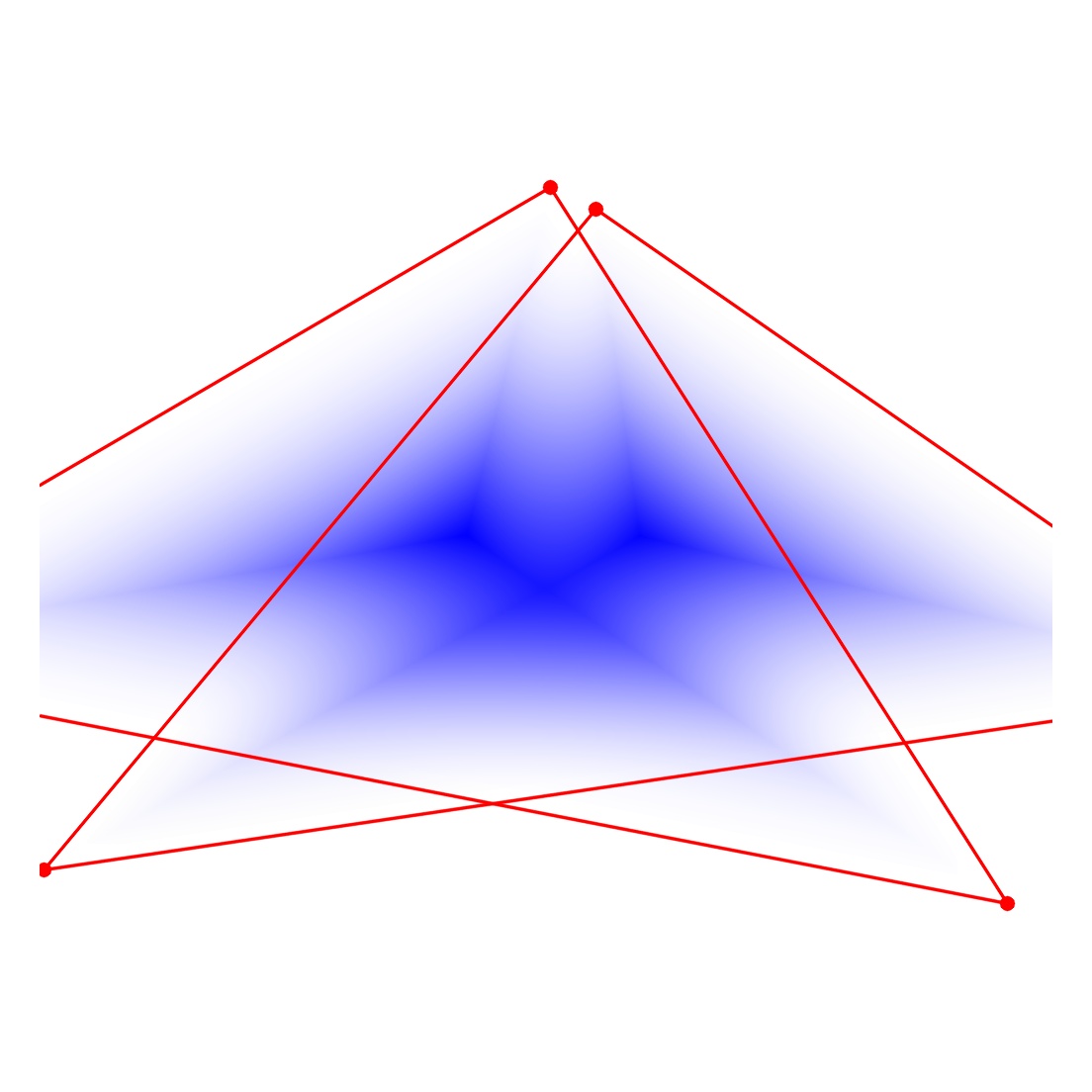}    &
			\fimgs{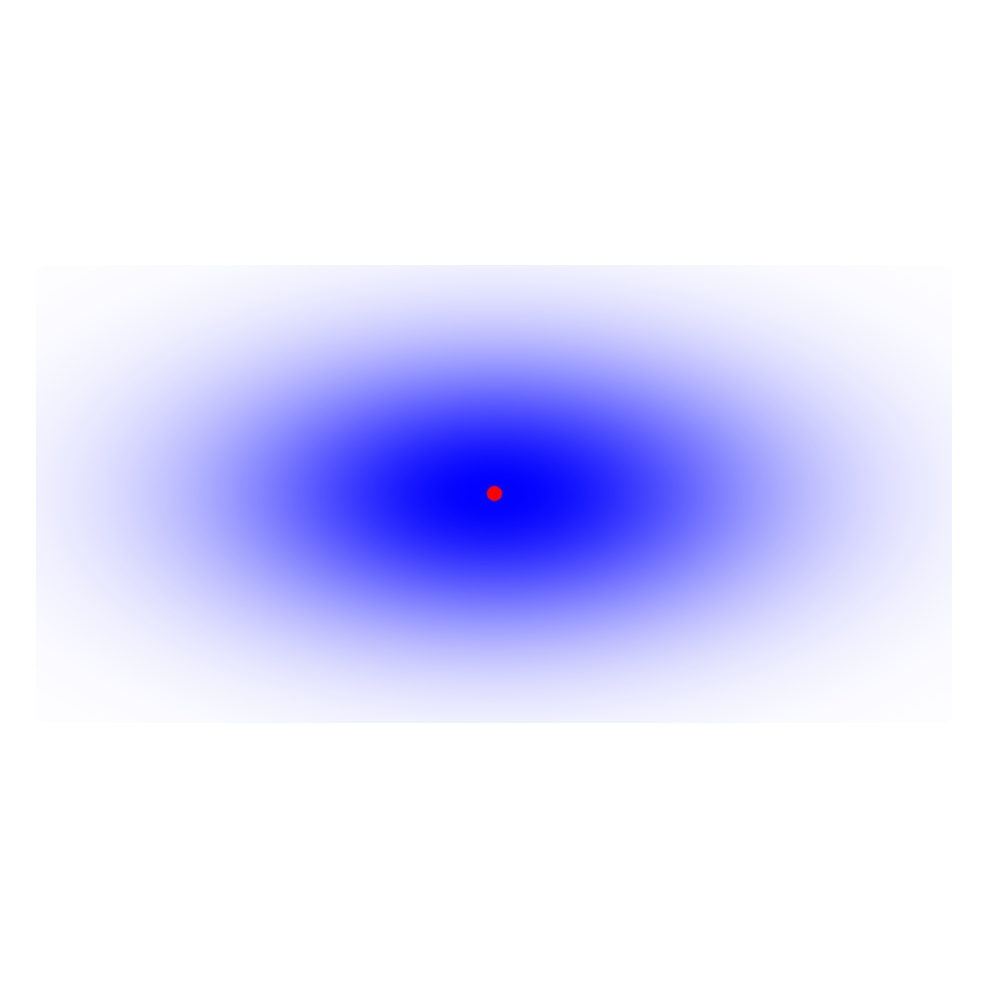}    &
			\fimgs{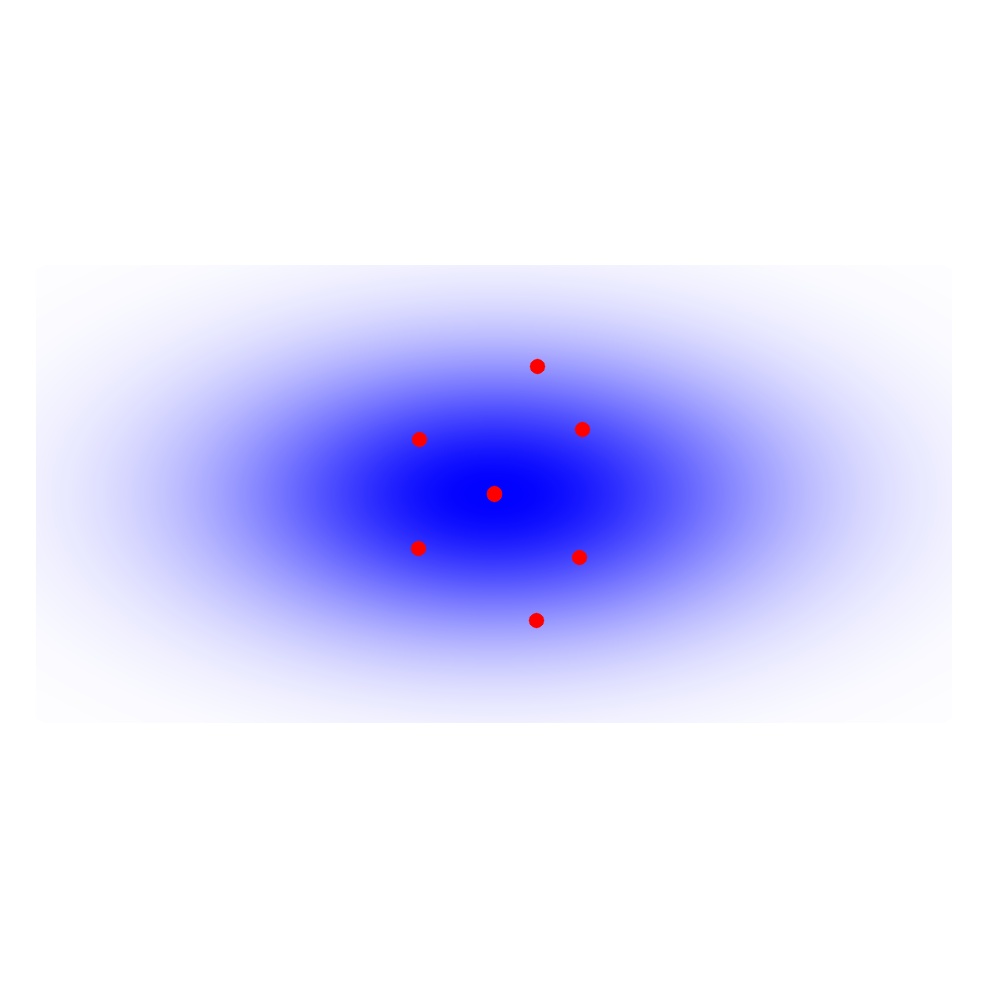}                                                                                                                                                                                                                                 \\
		\end{tabular}
	\end{minipage}\hfill
	\begin{minipage}[t]{0.49\linewidth}
		\centering
		\begin{tabular}{@{}
			>{\centering\arraybackslash}m{2.5mm} @{\,}
			>{\centering\arraybackslash}m{0.112\linewidth} @{\,}
			>{\centering\arraybackslash}m{0.112\linewidth} @{\,}
			>{\centering\arraybackslash}m{0.112\linewidth} @{\,}
			>{\centering\arraybackslash}m{0.112\linewidth} @{\,}
			>{\centering\arraybackslash}m{0.112\linewidth} @{\,}
			>{\centering\arraybackslash}m{0.112\linewidth} @{\,}
			>{\centering\arraybackslash}m{0.112\linewidth} @{}}
			                                                &               & \multicolumn{2}{c}{\mlbl{Ours}} & \multicolumn{2}{c}{\mlbl{TS~\cite{held2025trianglesplatting}}} & \multicolumn{2}{c}{\mlbl{3DGS~\cite{kerbl20233dgs}}}                                                    \\[0pt]
			                                                & \clbl{Target} & \clbl{$K{=}3$}                  & \clbl{$K{=}7$}                                                 & \clbl{$N{=}1$}                                       & \clbl{$N{=}3$} & \clbl{$N{=}1$} & \clbl{$N{=}8$} \\[1pt]
			\rlbls{Clover}                                  &
			\fimgs{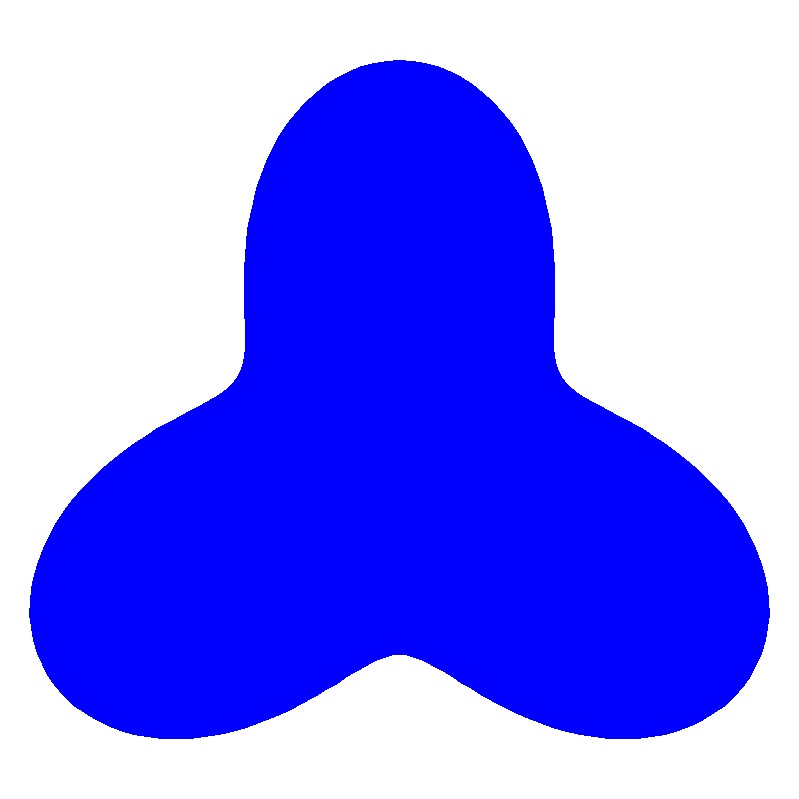}  &
			\fimgs{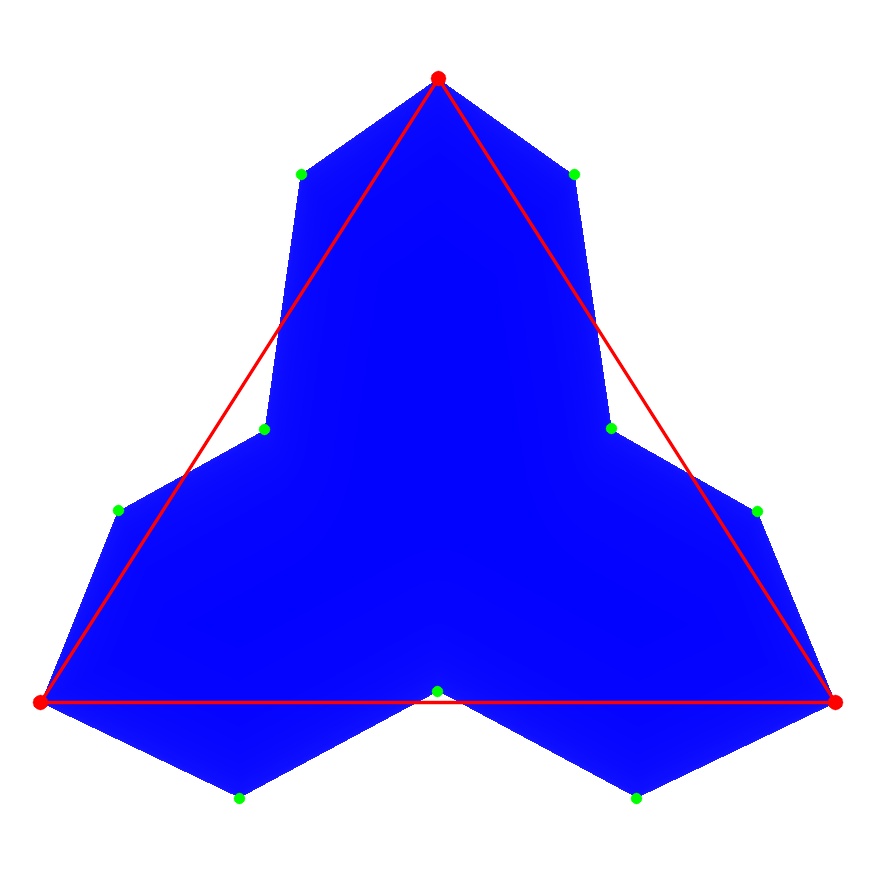} &
			\fimgs{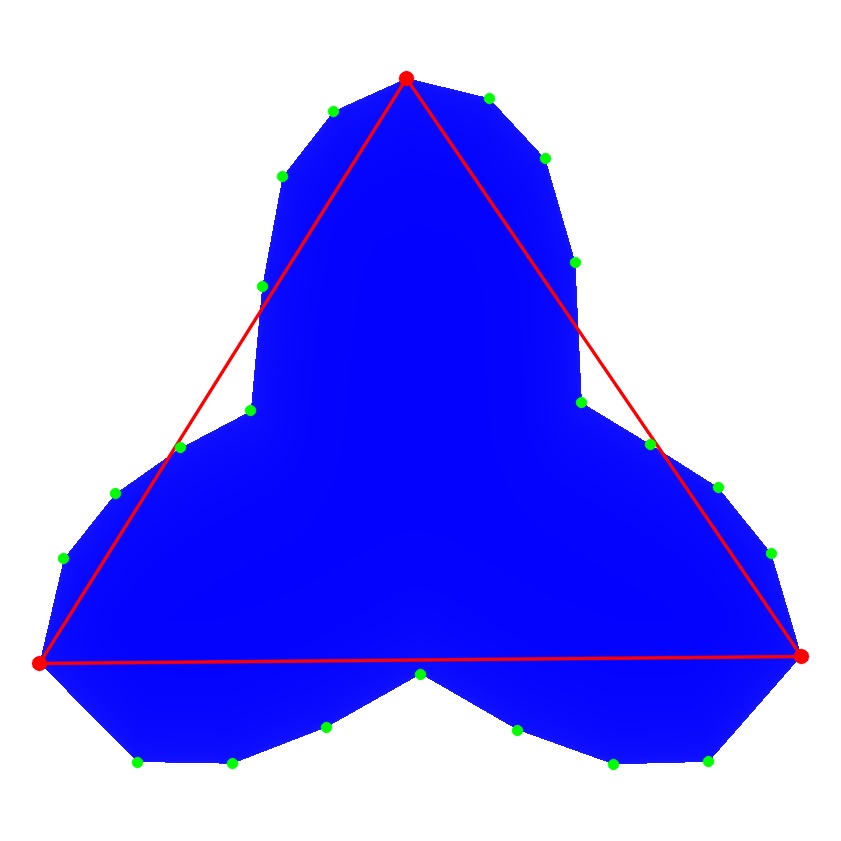} &
			\fimgs{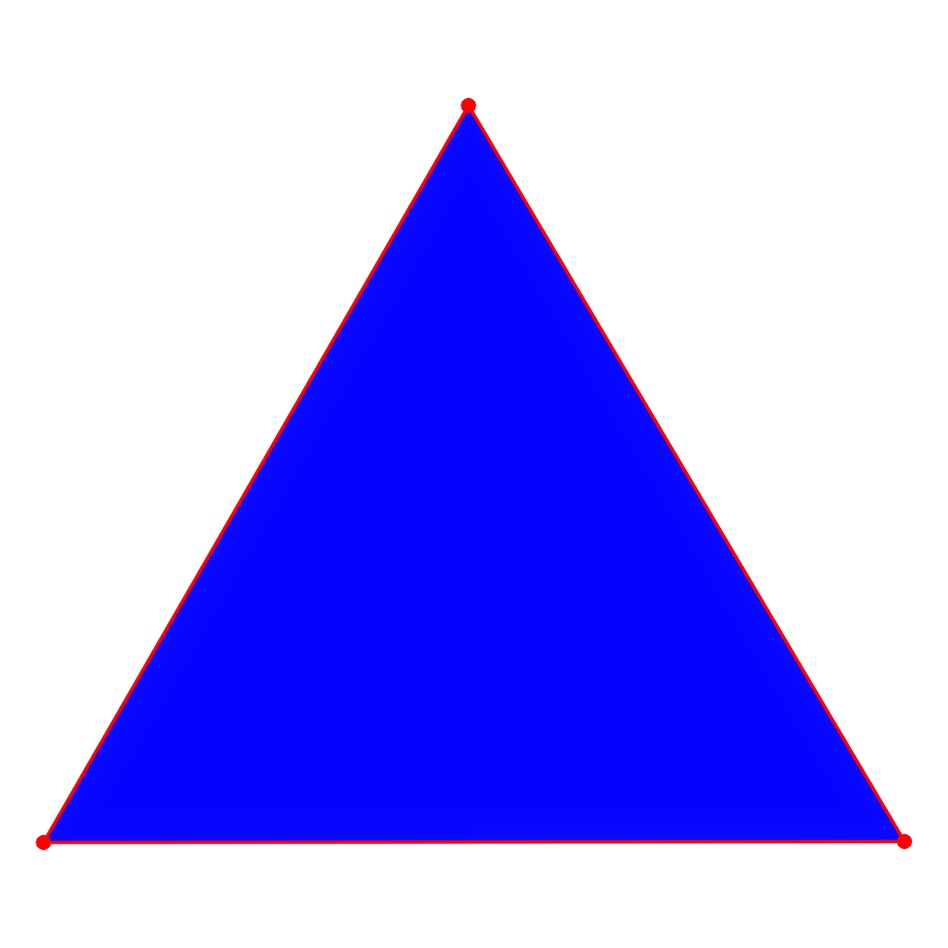}    &
			\fimgs{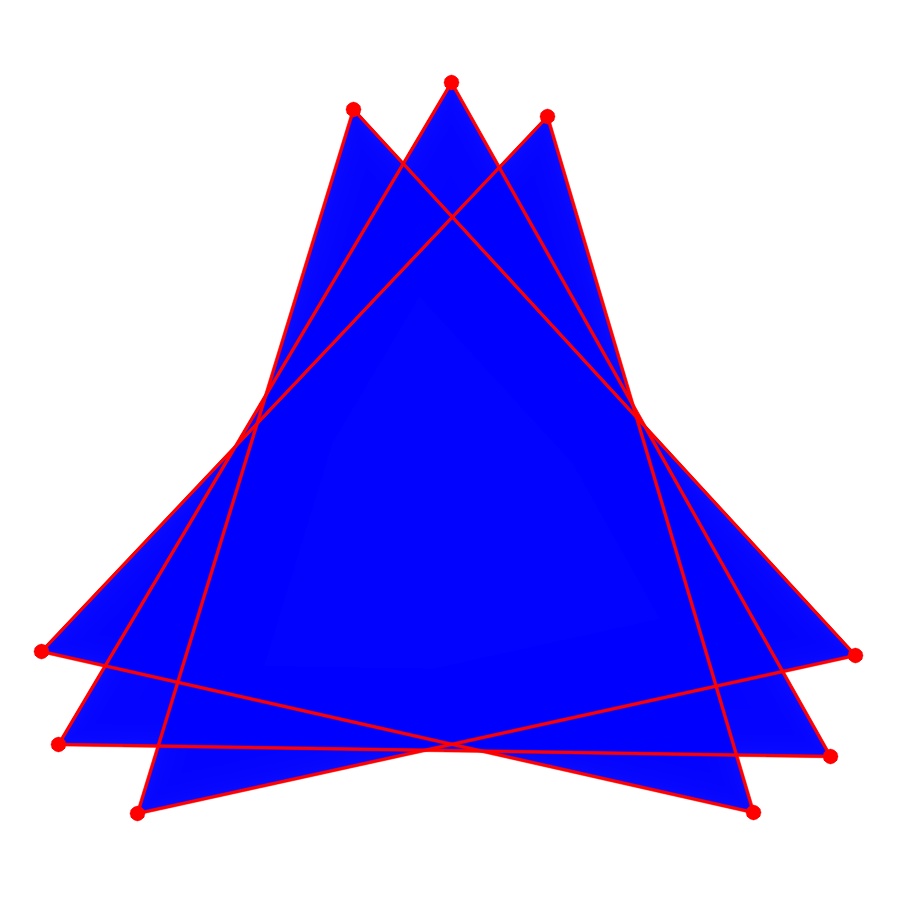}    &
			\fimgs{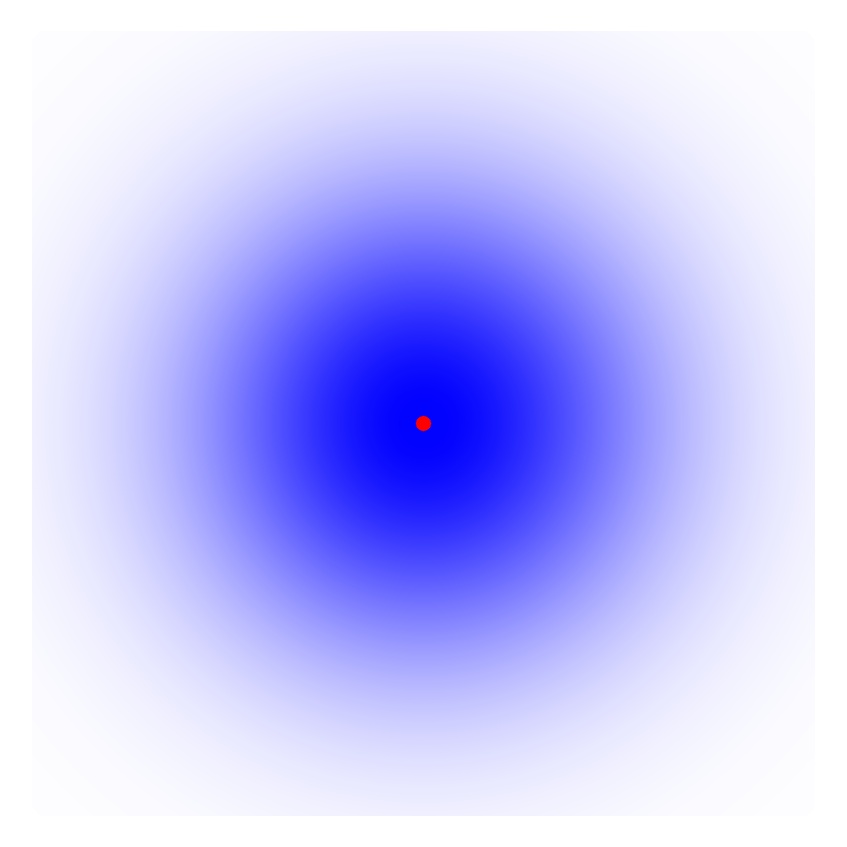}    &
			\fimgs{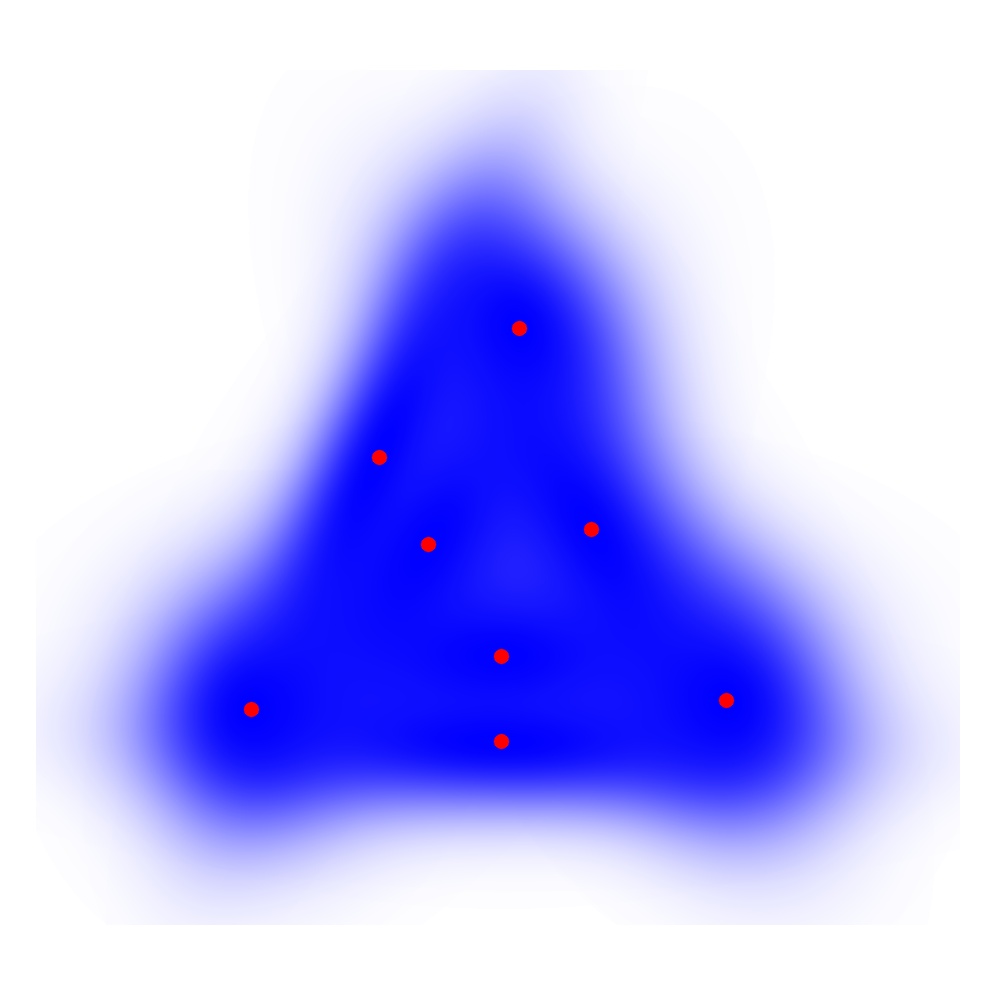}                                                                                                                                                                                                                                 \\
			\rlbls{Arrow}                                   &
			\fimgs{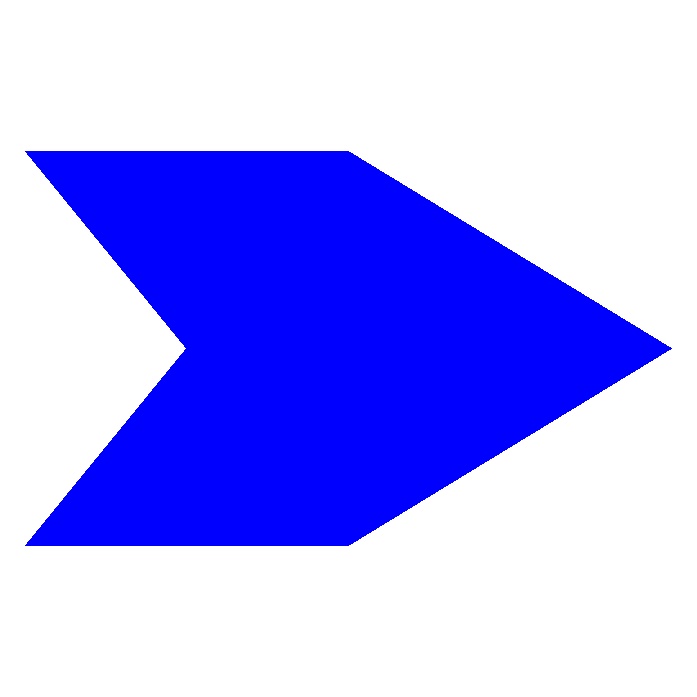}    &
			\fimgs{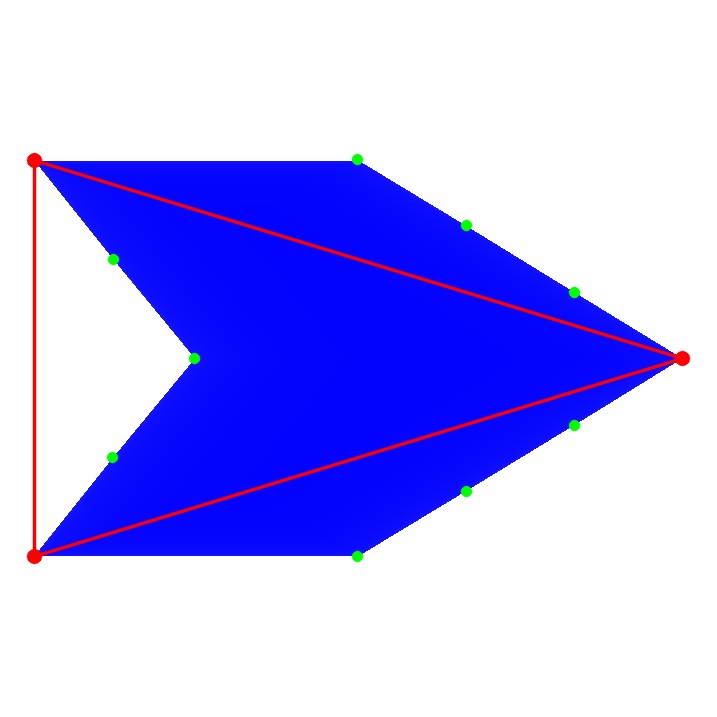}   &
			\fimgs{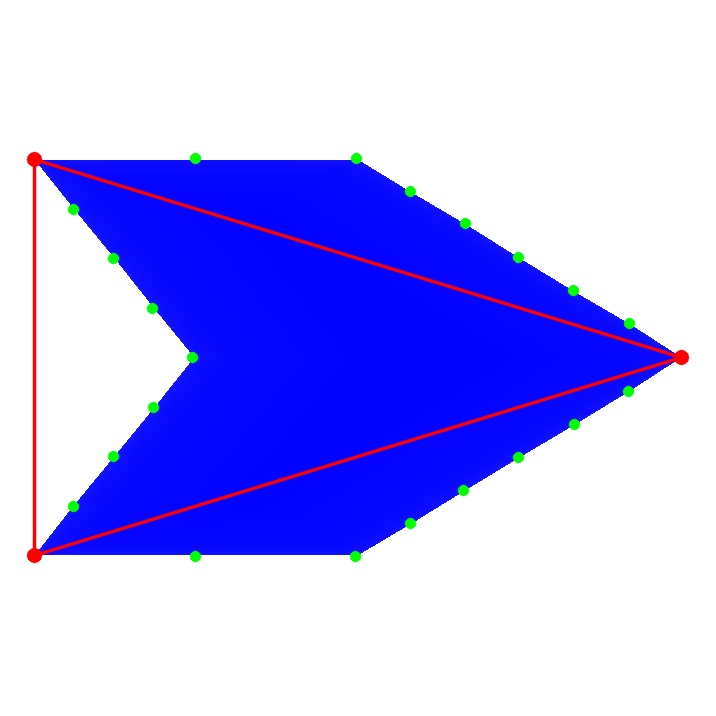}   &
			\fimgs{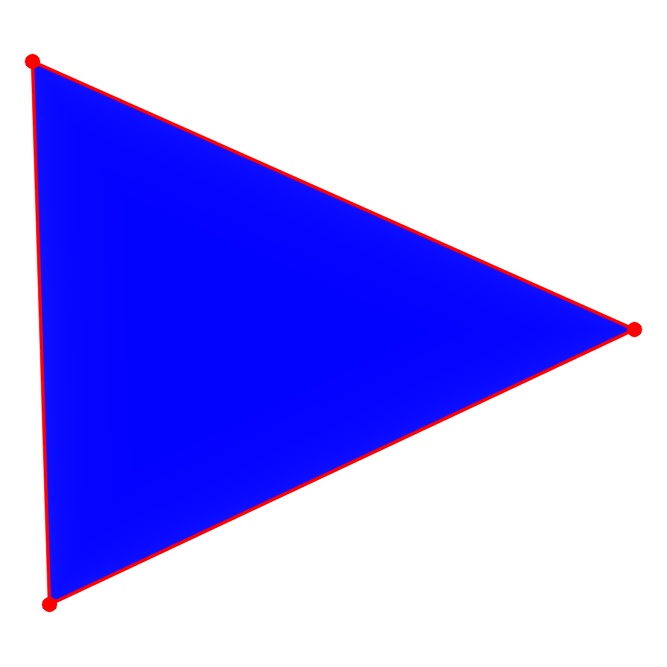}      &
			\fimgs{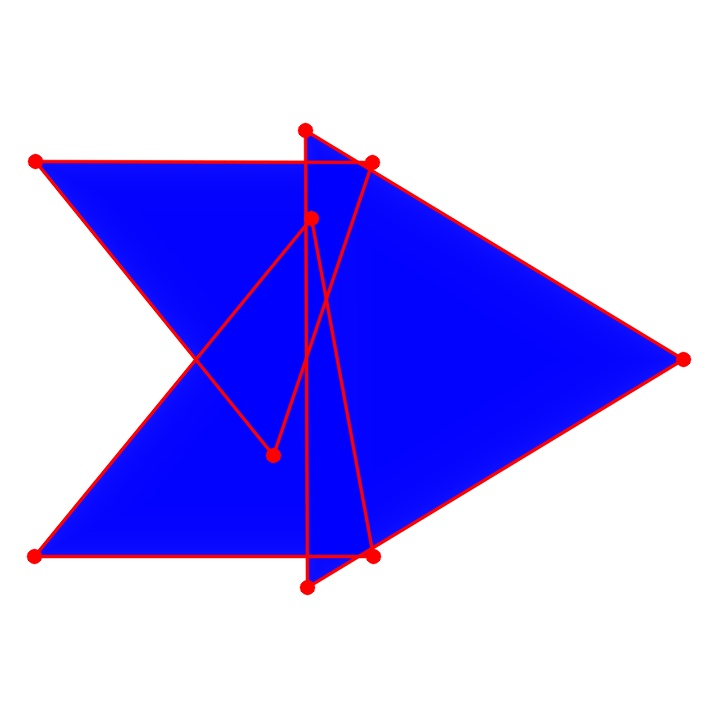}      &
			\fimgs{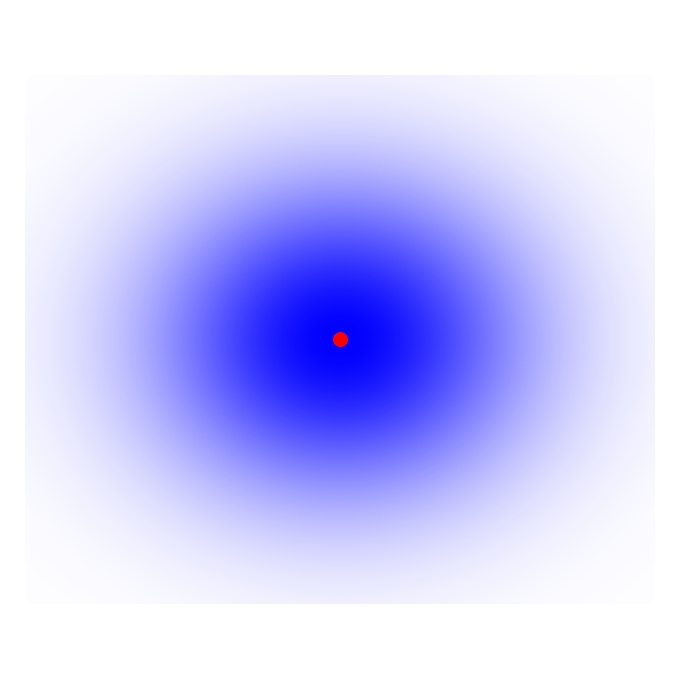}      &
			\fimgs{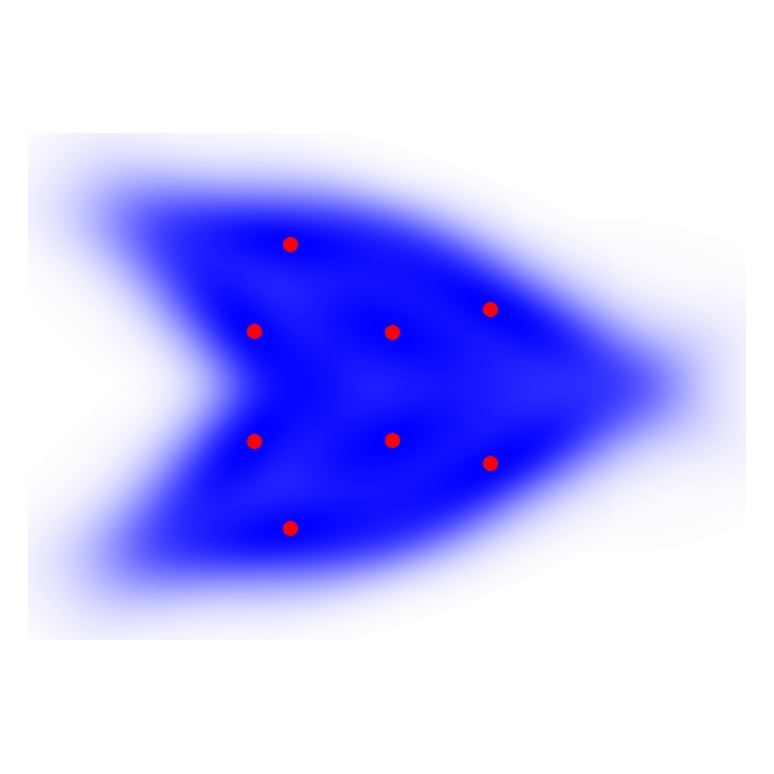}                                                                                                                                                                                                                                   \\
			\rlbls{Pac-Man}                                 &
			\fimgs{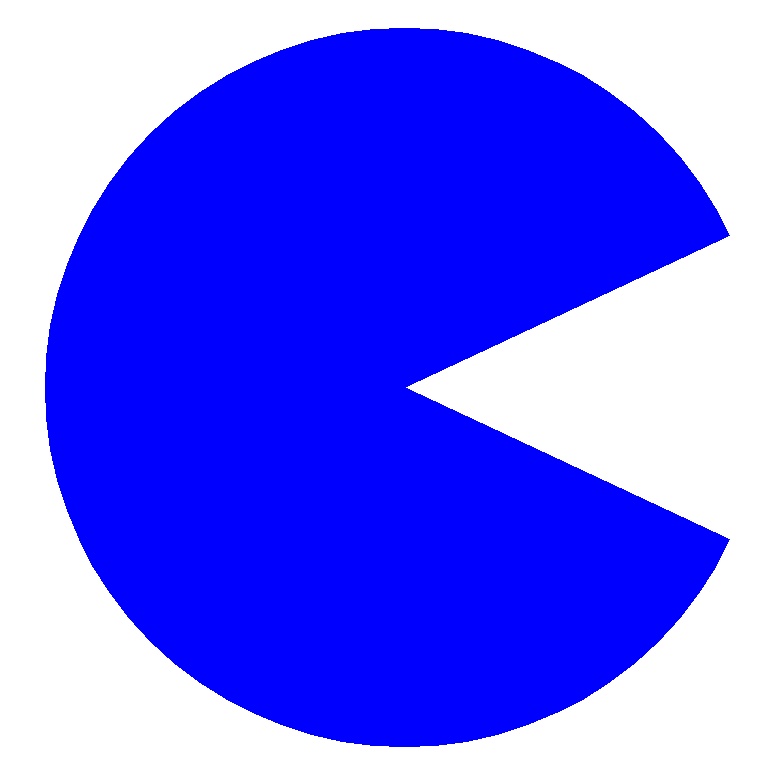}   &
			\fimgs{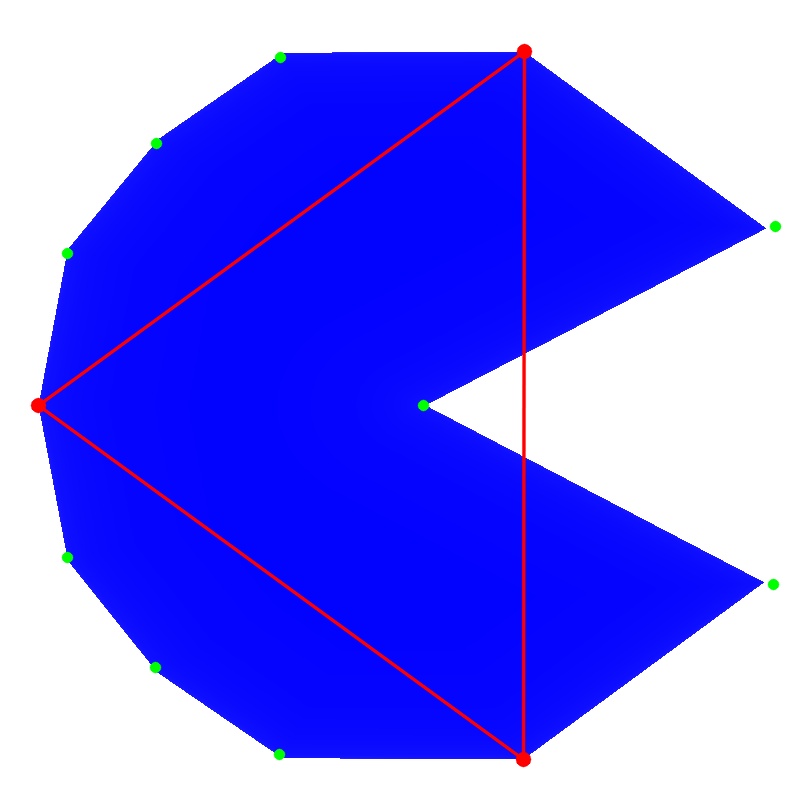}  &
			\fimgs{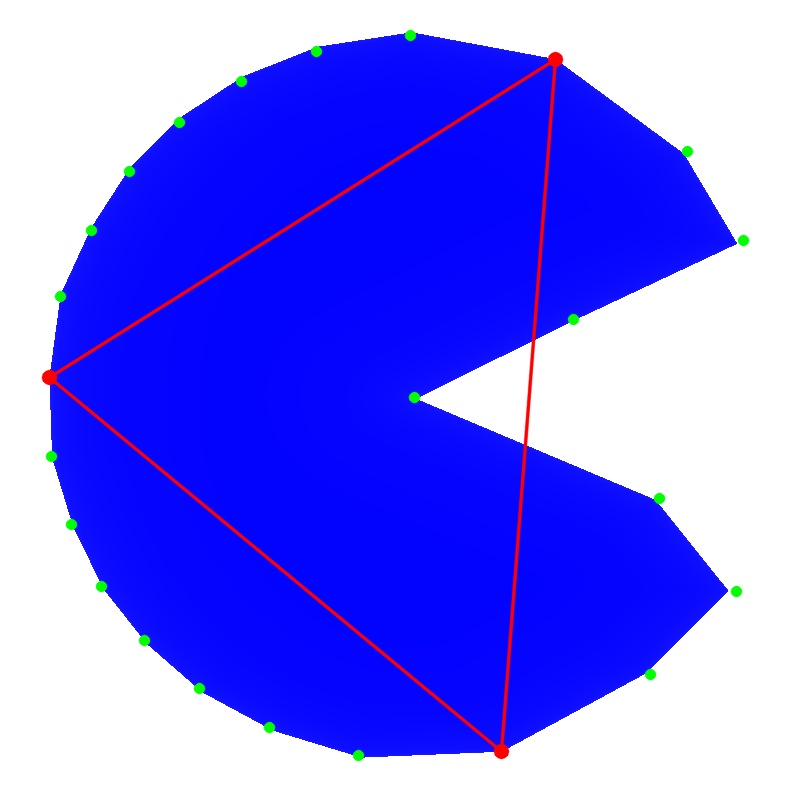}  &
			\fimgs{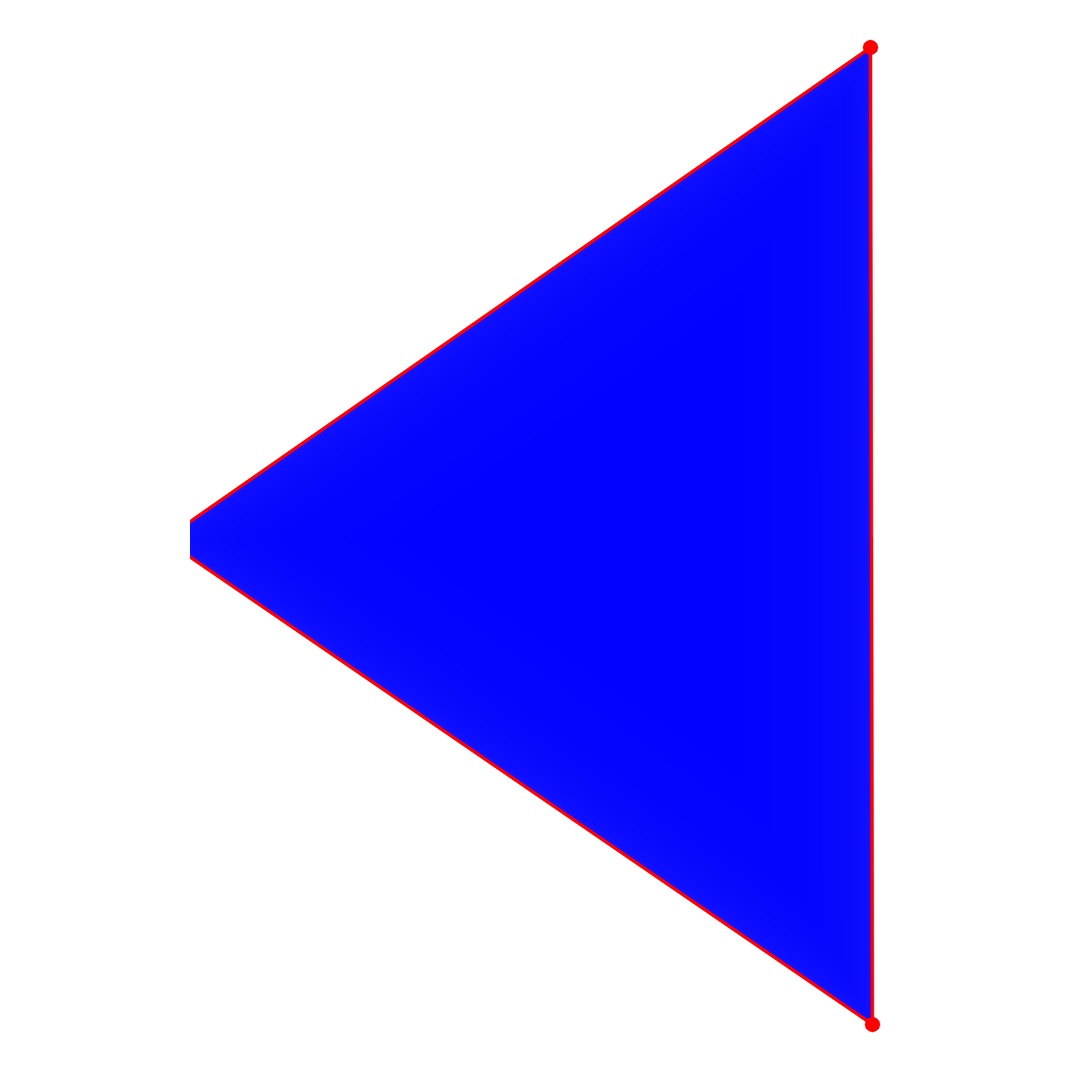}     &
			\fimgs{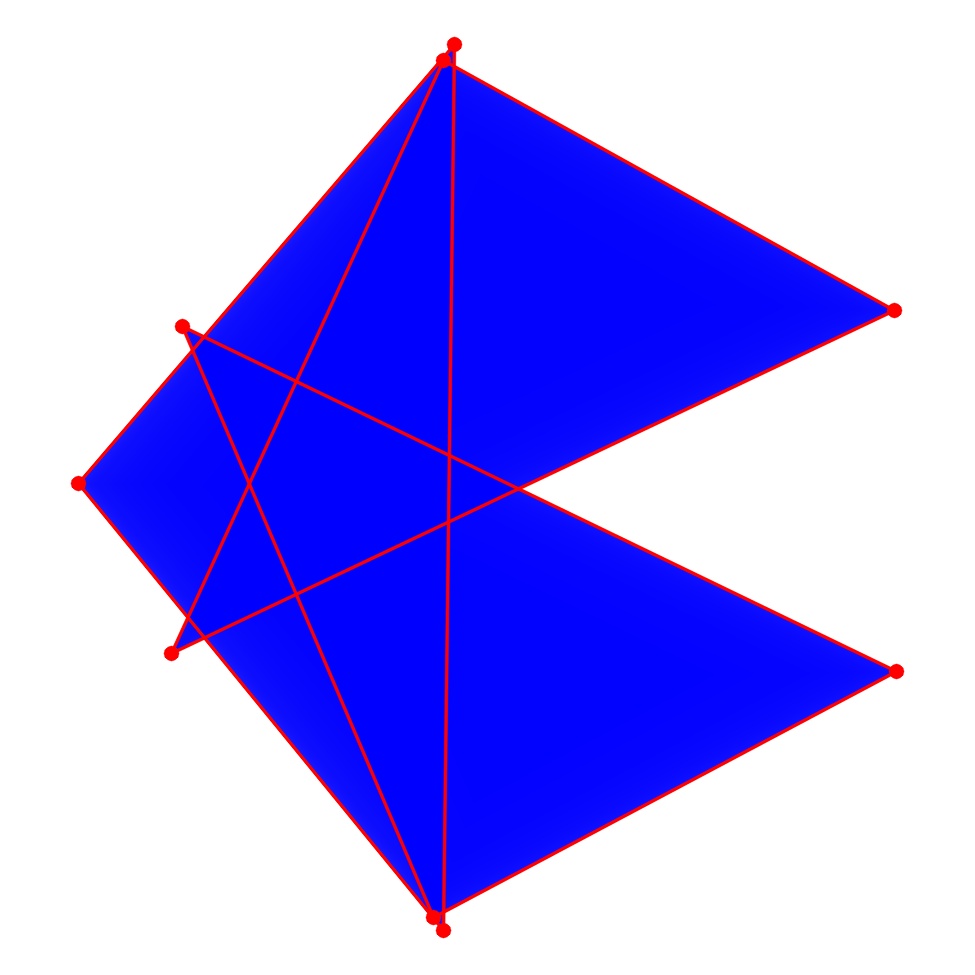}     &
			\fimgs{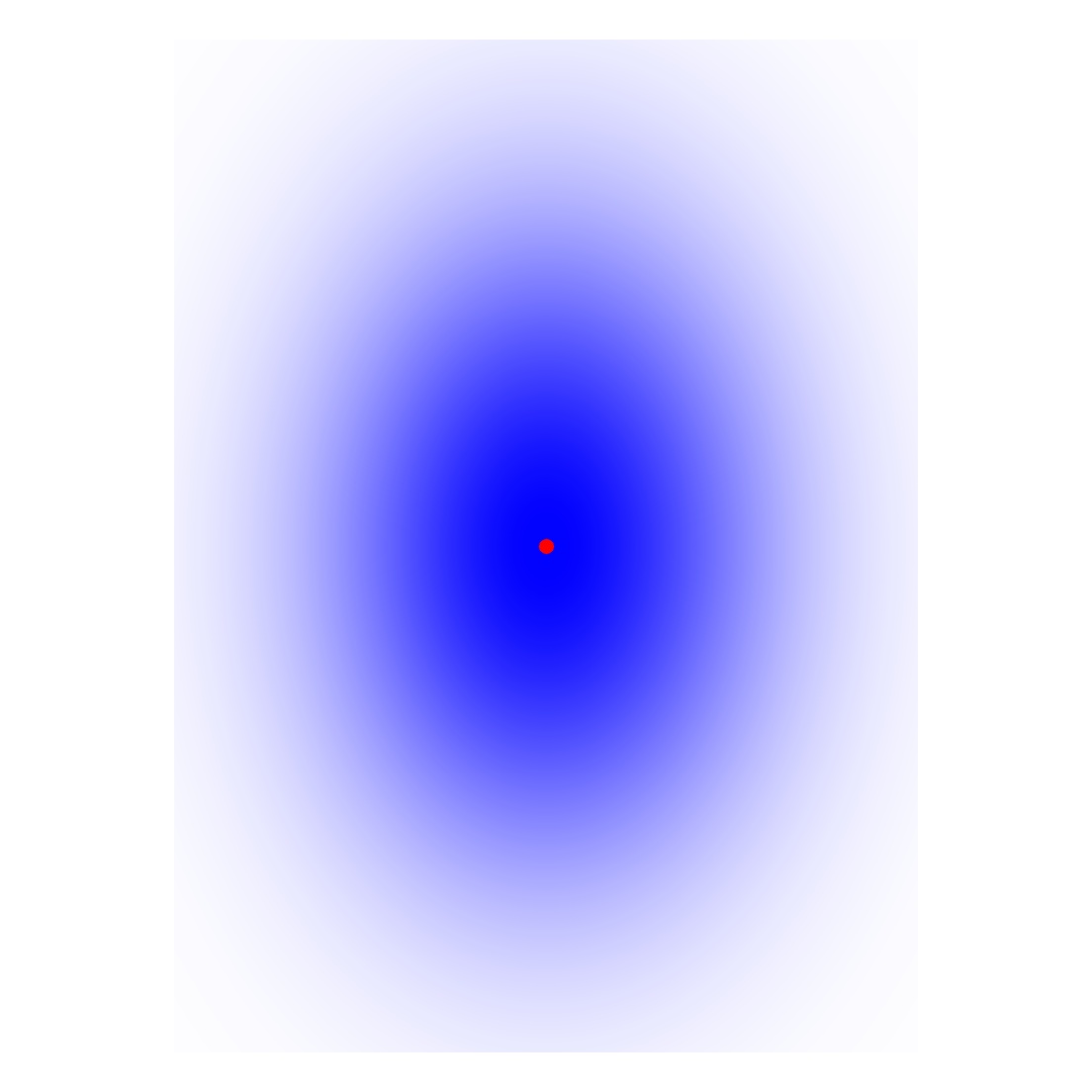}     &
			\fimgs{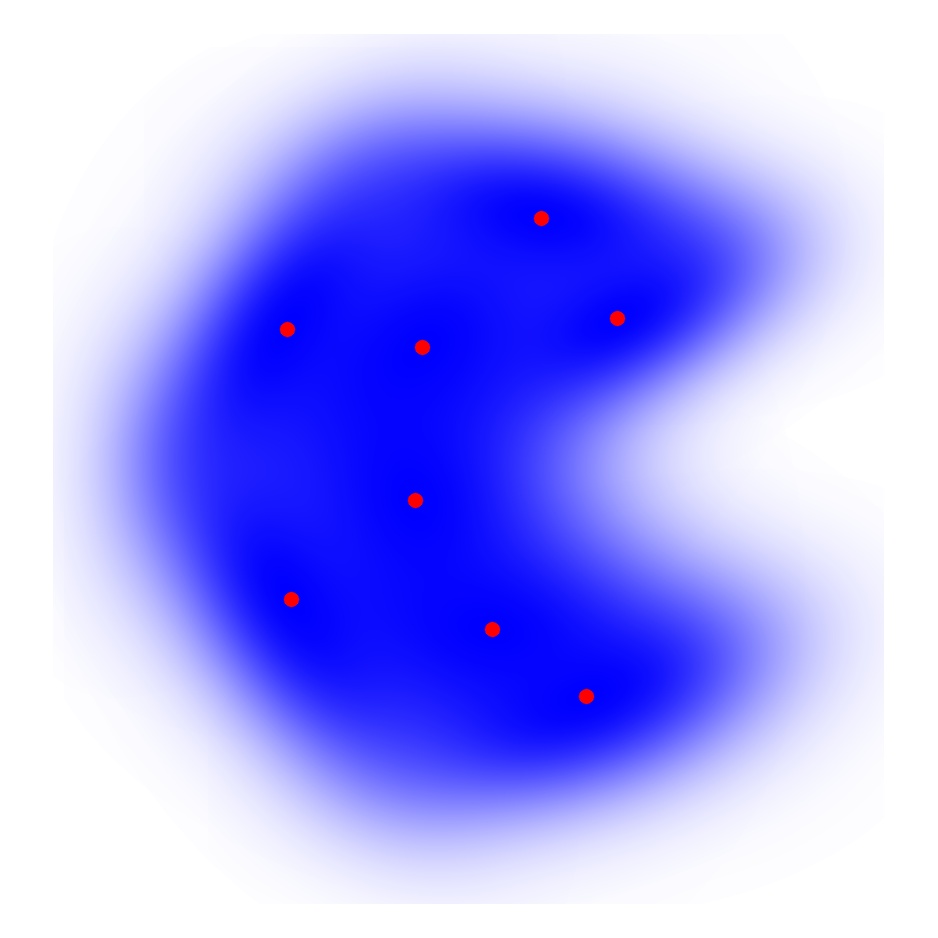}                                                                                                                                                                                                                                  \\
			\rlbls{Heart}                                   &
			\fimgs{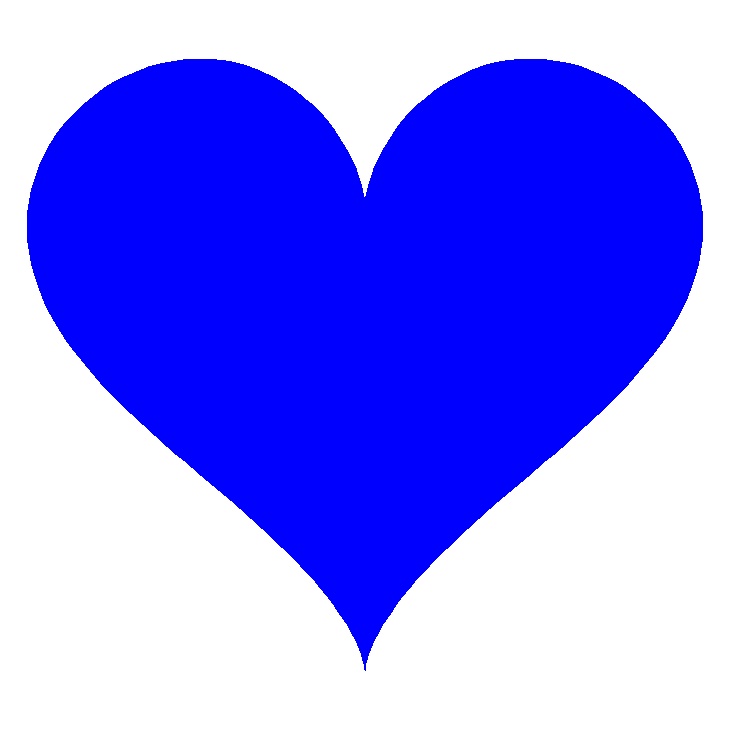}    &
			\fimgs{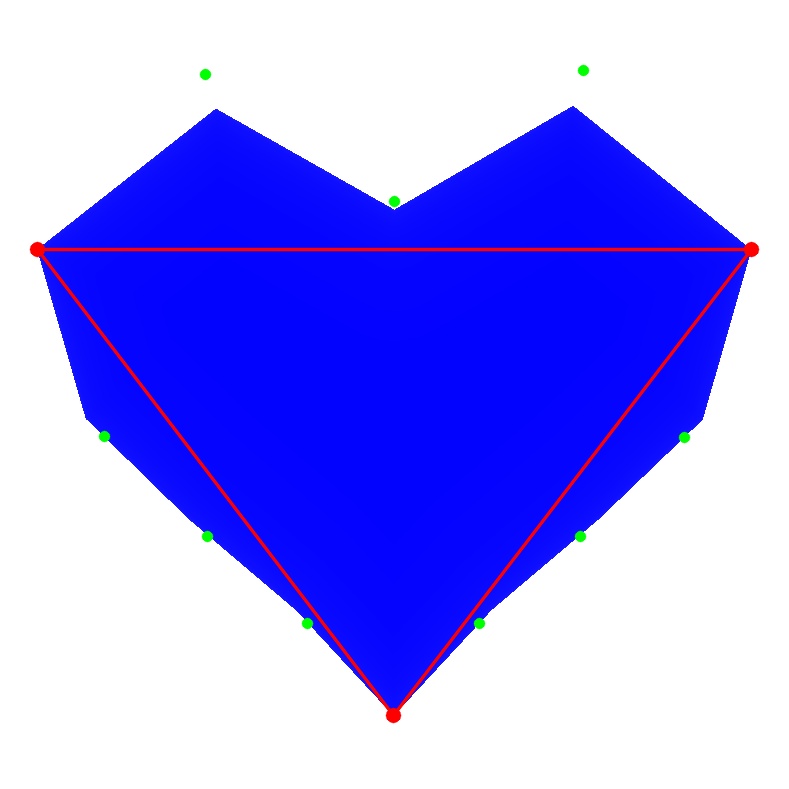}   &
			\fimgs{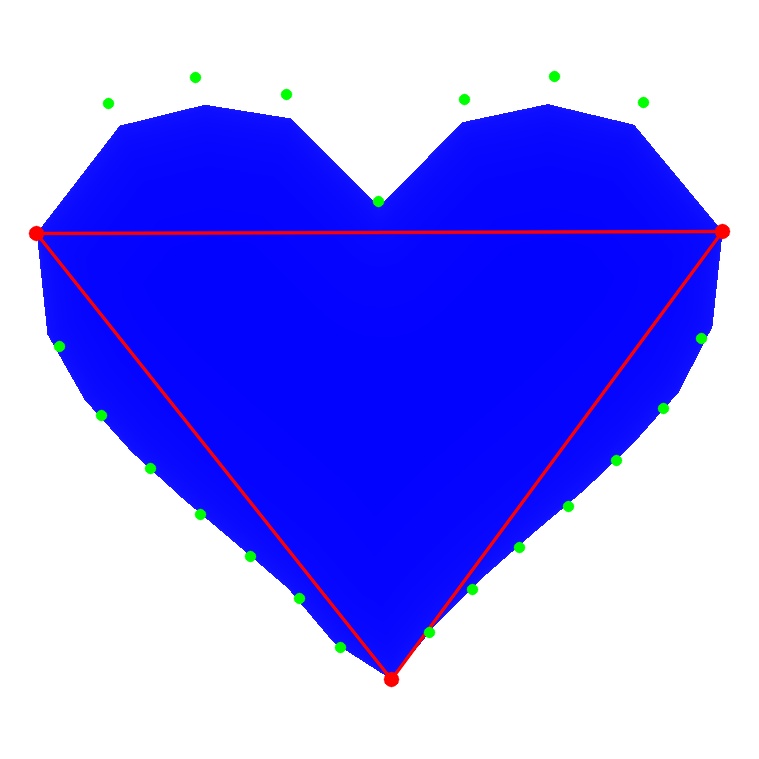}   &
			\fimgs{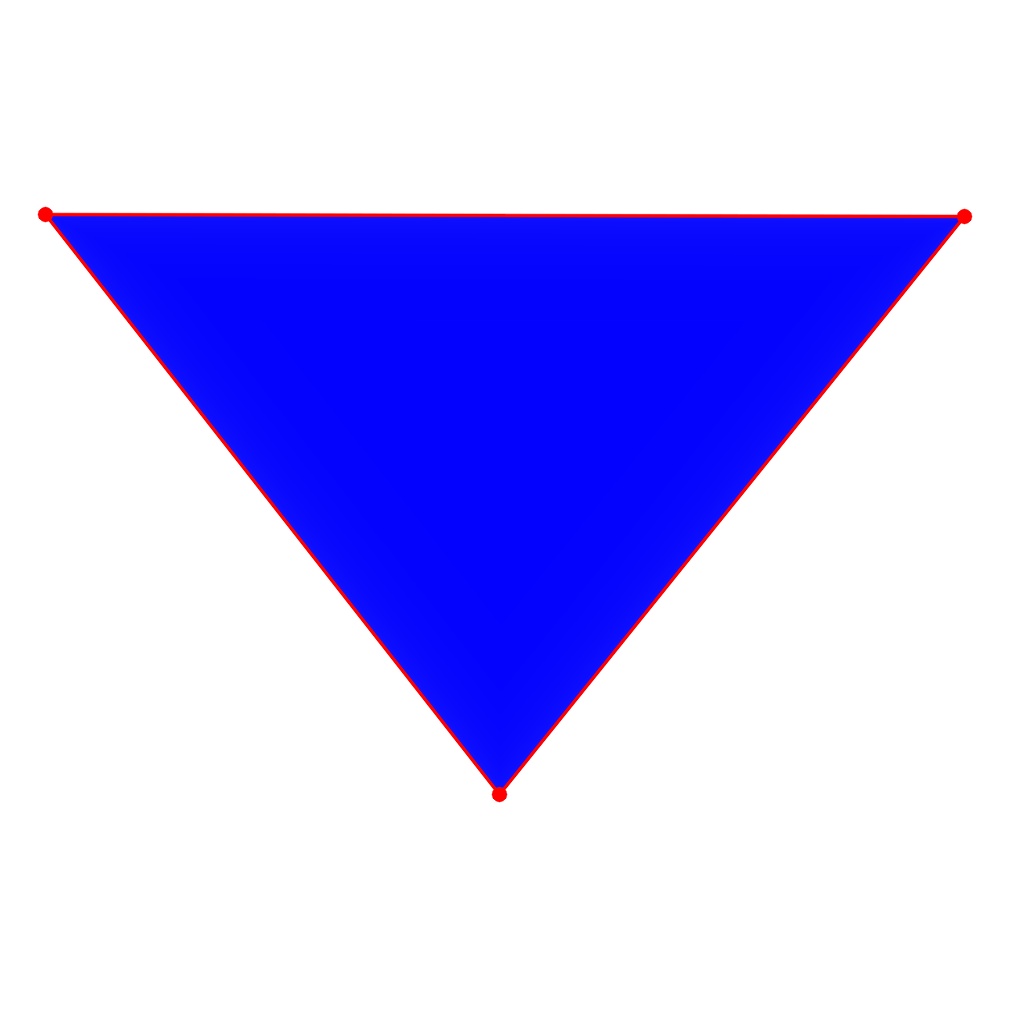}      &
			\fimgs{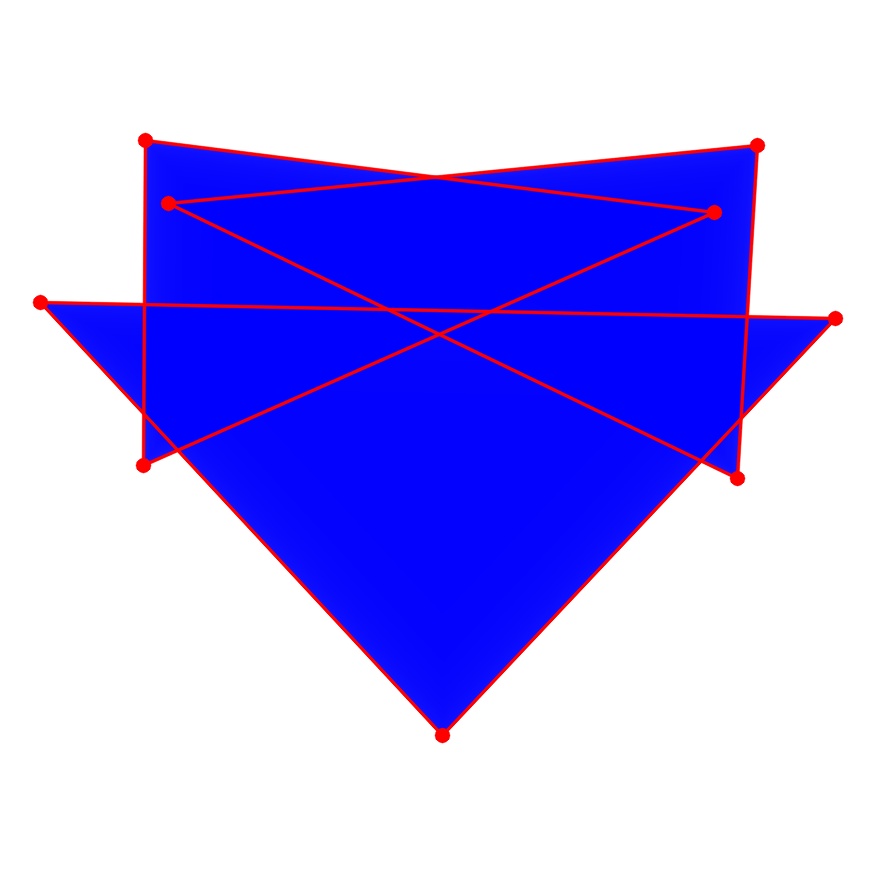}      &
			\fimgs{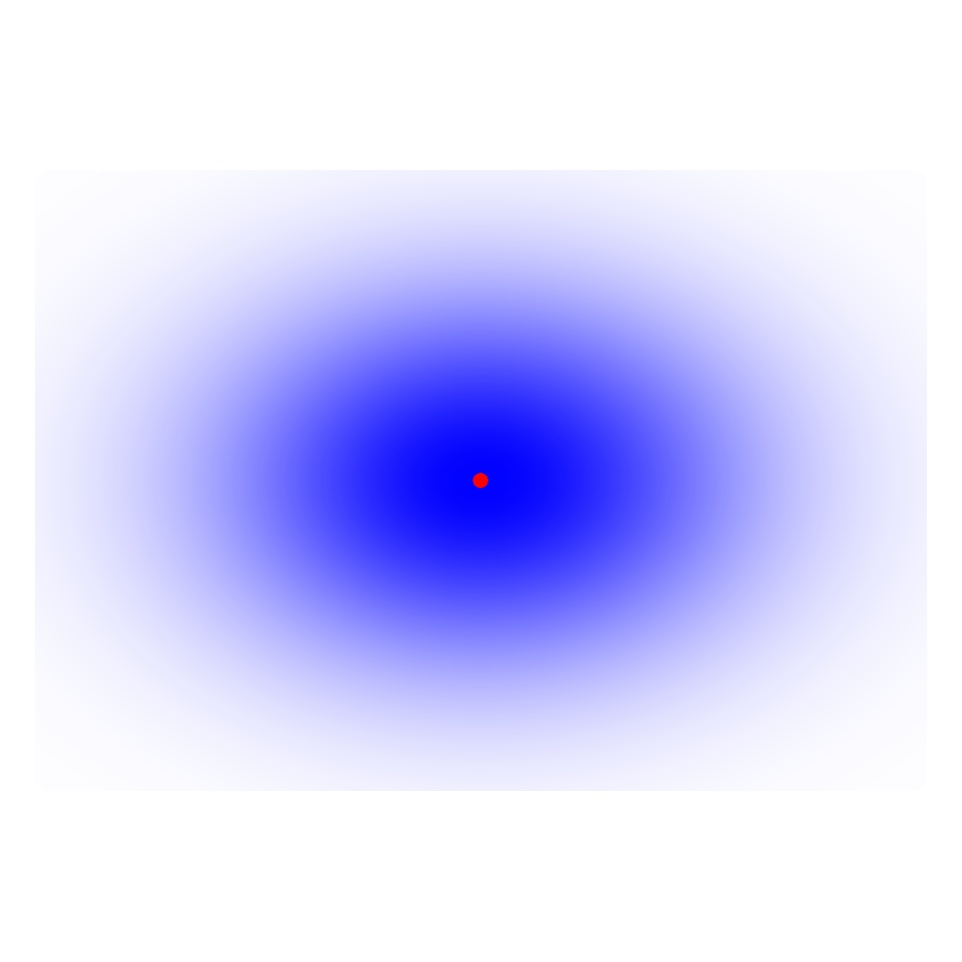}      &
			\fimgs{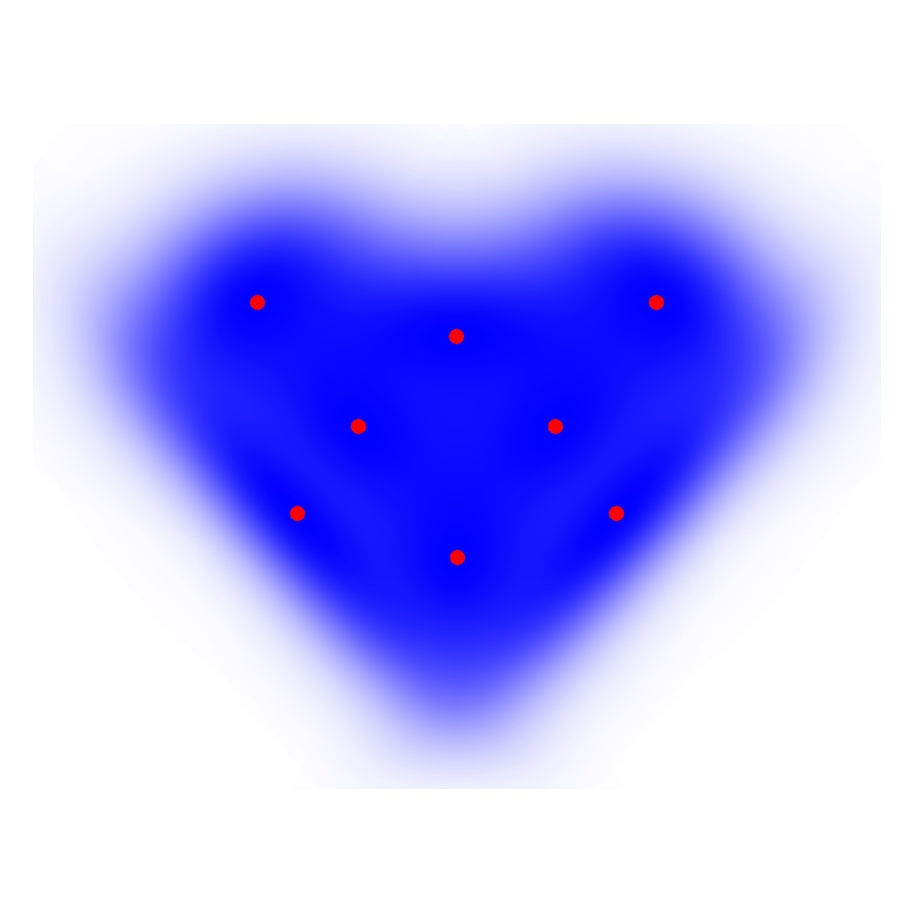}                                                                                                                                                                                                                                   \\
		\end{tabular}
	\end{minipage}
	\caption{\textbf{Single-primitive shape fitting.} In both sides, the same information is displayed. Each row represents a target shape and columns compare our method ($K{=}3$ and $K{=}7$ control points per edge) with TS~\cite{held2025trianglesplatting} and 3DGS~\cite{kerbl20233dgs}. $N$ denotes the number of primitives, being $N{=}1$ in our case. Red dots and lines show triangle vertices and edges; green dots mark the learned control points along edges; for 3DGS~\cite{kerbl20233dgs}, red dots indicate Gaussian centers. \textbf{Left:} Convex shapes. \textbf{Right:} Non-convex shapes.}
	\label{fig:shapes}
\end{figure*}

To illustrate the representational advantage of our deformable primitives, Fig.~\ref{fig:shapes} compares our method against TS~\cite{held2025trianglesplatting} and 3DGS~\cite{kerbl20233dgs} on synthetic 2D shape-fitting tasks. Each method optimizes from the same initial configuration to match a target shape. With a single primitive, our deformable triangles closely approximate circles, rectangles, Gaussians, and non-convex shapes such as a 3-clover, arrow, Pac-Man, and heart. In contrast, TS~\cite{held2025trianglesplatting}'s rigid boundary limits it to crude triangular approximations even with three primitives ($3\times$ more parameters), while 3DGS~\cite{kerbl20233dgs} fundamentally cannot produce sharp edges even with eight ($8\times$), yielding only soft blobs. Our method handles both convexity and concavity within a single deformable primitive.

\subsection{Ablation Study}
\label{sec:ablation}

To validate our design, we first analyze learned control-point statistics, then ablate key components.

\begin{figure*}[t!]
	\centering
	\begin{minipage}[t]{0.27\textwidth}
		\centering
		\includegraphics[width=\linewidth]{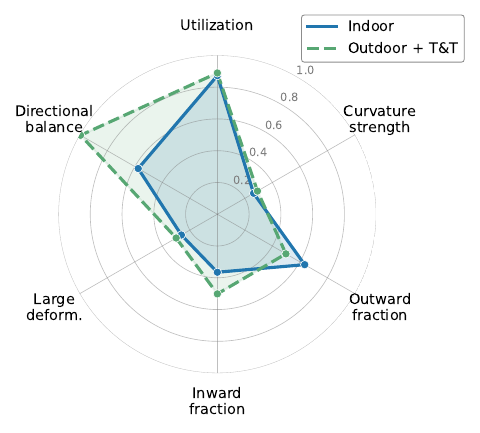}
	\end{minipage}\hfill
	\begin{minipage}[t]{0.34\textwidth}
		\centering
		\includegraphics[width=\linewidth]{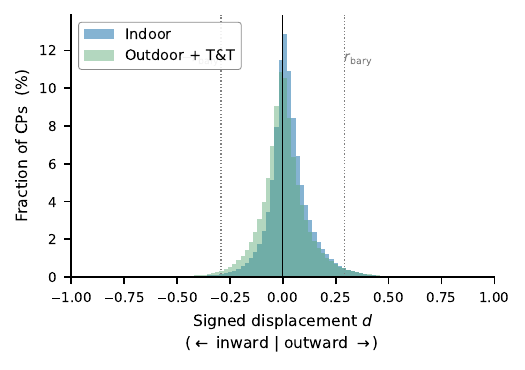}
	\end{minipage}\hfill
	\begin{minipage}[t]{0.34\textwidth}
		\centering
		\includegraphics[width=\linewidth]{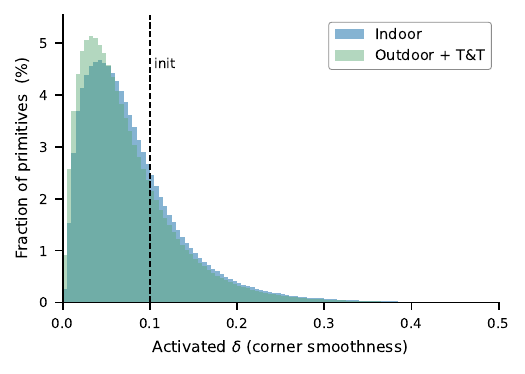}
	\end{minipage}
	\vspace{-0.3cm}
	\caption{\textbf{Learned parameter analysis} (Indoor~\cite{barron2022mipnerf360} vs.\ Outdoor~\cite{barron2022mipnerf360}\,+\,T\&T~\cite{knapitsch2017tanks}). \textbf{Left:} Radar profile of six CP application metrics, each normalized to $[0,1]$. \textbf{Middle:}~Signed CP displacement histogram ($d{<}0$: inward, $d{>}0$: outward); dashed lines mark $\pm r_\text{bary}$. Indoor scenes show a clear outward skew; outdoor ones are nearly symmetric. \textbf{Right:}~Distribution of the corner-smoothness parameter $\delta$ (softplus-activated); the dashed line marks the initialization ($\delta{=}0.1$). Most primitives converge to sharper corners.}
	\label{fig:cp_utilization}
	\vspace{-0.2cm}
\end{figure*}

\textbf{Control-point utilization.}
To verify that the learned control points are not degenerate, we analyze the displacement statistics across all $323$M control points from the $11$ evaluated scenes at convergence.
Each displacement $d$ is a signed scalar in barycentric coordinates: positive values push the boundary outward (away from the opposite vertex), negative ones pull it inward.
We normalize by the inradius $r_\text{bary}$ of the unit barycentric triangle; a displacement of $r_\text{bary}$ moves a control point from the edge midpoint to the centroid. Across all scenes, $88.5\%$ of control points carry meaningful deformation ($|d|{>}0.01$), with a mean magnitude of $\bar{|d|}{=}0.083$ ($28\%$ of $r_\text{bary}$); $29.0\%$ of all control points undergo large deformations ($|d|{>}0.1$).
\Cref{fig:cp_utilization} compares indoor and outdoor scene categories.
Indoor scenes exhibit a strong outward bias ($63.5\%$ outward vs.\ $36.5\%$ inward among active CPs), consistent with primitives expanding to cover smooth walls and flat surfaces. Outdoor~\cite{barron2022mipnerf360} and T\&T~\cite{knapitsch2017tanks} scenes are nearly perfectly balanced ($50\%$/$50\%$), reflecting the need for both convex and concave boundary adjustments to capture complex vegetation and fine structures.

\textbf{Corner smoothness.}
The learned $\delta$ has a global mean of $0.075$ (median $0.061$), with the majority of primitives falling below the $0.1$ initialization (see Figure~\ref{fig:cp_utilization}-right), indicating that the optimizer favors crisper corners while keeping the parameter well-distributed rather than collapsing it to zero.

\textbf{Component ablation.}
Table~\ref{tab:ablation_design} evaluates three key components averaged over all scenes. Among the three factors, edge deformation has the largest impact: removing it entirely (\emph{w/o\,Def.}) produces the biggest drop in PSNR ($-0.09$\,dB) and the largest LPIPS increase ($+0.006$), confirming that the learnable control points are the primary driver of reconstruction improvement (baseline wins on 9 of 11 scenes). Disabling the curvature regularizer \emph{w/o\,$\mathcal{L}_\text{curv}$}, ($\lambda_6{=}0$ in Eq.~\eqref{eq:loss}) has a minor effect on average metrics, with the baseline winning on 7 of 11 scenes. However, analyzing the learned control points reveals that without $\mathcal{L}_\text{curv}$, the fraction of CPs with large deformations ($|d|{>}0.1$) rises from $29.0\%$ to $36.0\%$ and the mean displacement magnitude increases from $0.083$ to $0.102$ ($+23\%$). This confirms that the regularizer acts as a control mechanism that prevents unconstrained boundary growth without significantly affecting reconstruction quality. Finally, removing the control point warm-up delay (\emph{w/o\,Delay}, $T_{\text{delay}}{=}0$) reduces PSNR by $0.04$\,dB, with the baseline winning on 8 of 11 scenes, the largest single-scene drop being $+0.23$\,dB on Kitchen, confirming that letting vertex positions settle before activating boundary deformation improves convergence.

\begin{table}[H]
	\centering
	\caption{\textbf{Ablation study} averaged over all scenes. \textbf{Bold} means best per metric.}
    \vspace{-0.2cm}
	\label{tab:ablation_design}
	\begin{tabular}{l|ccc}
		\toprule
		Configuration                 & PSNR$\uparrow$    & SSIM$\uparrow$    & LPIPS$\downarrow$ \\
		\midrule
		\textbf{Ours} (baseline)      & \textbf{26.588} & \textbf{0.824} & \textbf{0.174} \\
		w/o Deformation               & 26.500          & 0.822          & 0.180          \\
		w/o $\mathcal{L}_\text{curv}$ & 26.549          & 0.823          & 0.175          \\
		w/o Delay                     & 26.545          & 0.823          & 0.175          \\
		\bottomrule
	\end{tabular}
\end{table}

\section{Conclusion}
\label{sec:conclusion}

We have presented DETRIS, a method that augments triangle primitives' capabilities with $K$ learnable control points per edge to represent shapes as a soup of non-uniform primitives. By performing all per-pixel deformations and distance computations in a canonical barycentric space, our formulation ensures these boundaries remain strictly view-independent within a fully differentiable rendering pipeline. Our approach also introduces a ray-casting winding number test for robust inside/outside classification, a polynomial smooth-minimum distance field with controllable corner rounding, and a power-law window function for opacity falloff. A simple learning rate delay suffices for the effective use of control points without requiring explicit gradient scheduling. All our claims have been experimentally validated on real-world scenes, providing competitive visual fidelity and rendering speed among non-volumetric methods. Furthermore, our method demonstrates better adaptability and versatility by learning specialized primitive shapes directly from data, as opposed to current methods that rely on fixed geometric shapes, which limits their expressiveness or drastically increases the number of primitives required to model a scene. Future work will explore multi-layer primitives for handling semi-transparent and thin structures.

\section*{Acknowledgements}
This work has been supported by the project GRAVATAR PID2023-151184OB-I00 funded by MCIU/AEI/10.13039/501100011033 and by ERDF, UE.

\bibliographystyle{splncs04}
\bibliography{main}

\clearpage
\raggedbottom
\setcounter{section}{0}
\setcounter{subsection}{0}
\setcounter{figure}{0}
\setcounter{table}{0}
\setcounter{equation}{0}
\renewcommand{\thesection}{S\arabic{section}}
\renewcommand{\thesubsection}{S\arabic{section}.\arabic{subsection}}
\renewcommand{\thefigure}{S\arabic{figure}}
\renewcommand{\thetable}{S\arabic{table}}
\renewcommand{\theequation}{S\arabic{equation}}
\renewcommand{\theHsection}{S\arabic{section}}
\renewcommand{\theHsubsection}{S\arabic{section}.\arabic{subsection}}
\renewcommand{\theHsubsubsection}{S\arabic{section}.\arabic{subsection}.\arabic{subsubsection}}
\renewcommand{\theHfigure}{S\arabic{figure}}
\renewcommand{\theHtable}{S\arabic{table}}
\renewcommand{\theHequation}{S\arabic{equation}}

\begin{center}
	{\Large\bfseries Supplementary Material\par}
	\vspace{3pt}
	{\large Deformable Triangle Splatting:\\ Flexible Primitives for Real-Time Radiance Field Rendering\par}
\end{center}
\vspace{0.8em}

\section{Initialization \& Hyperparameters}
\label{sec:hyperparams}

\noindent\textbf{Initialization.}
We inherit the SfM-based~\cite{schonberger2016sfmrevisited} triangle initialization from Triangle Splatting~\cite{held2025trianglesplatting}: each SfM point $\mathbf{q}$ spawns a triangle whose vertices $\mathbf{v}_i = \mathbf{q} + 2.23\, r \, \mathbf{u}_i$ are placed on the unit sphere and scaled by the nearest-neighbor distance $r$.
Each primitive is assigned an initial opacity of $0.28$, falloff exponent $\sigma{=}1.16$, corner rounding $\delta{=}0.1$, and all edge control-point displacements $d_{i,k}{=}0$.
These values were determined empirically to ensure stable gradient flow from the first iteration.

\noindent\textbf{Densification.}
Densification is performed every 500 iterations, from iteration 500 to 25\,000.
At each step the triangle count is increased by $30\%$ via midpoint subdivision, following the Markov Chain Monte Carlo (MCMC) sampling strategy of TS~\cite{held2025trianglesplatting}, alternating between $\sigma^{-1}$ and opacity for Bernoulli sampling.

\noindent\textbf{Hyperparameters.}
Table~\ref{tab:hyperparams} lists all learning rates, loss weights, and adaptive-density thresholds used in our experiments.
Compared to TS, our method introduces three additional parameters: the control-point learning rate (\texttt{lr\_ctrl\_pts}), the corner-rounding learning rate (\texttt{lr\_delta}), and the curvature regularization weight ($\lambda_\text{curv}$).
The control-point learning rate is delayed to zero for the first 3\,000 iterations (\texttt{deform\_delay}), allowing the vertex positions to settle before enabling boundary deformations.
Following TS~\cite{held2025trianglesplatting}, outdoor and Tanks\,\&\,Temples scenes use $l_2$ photometric loss, while indoor scenes use $l_1$; all other parameters are shared.
For reference, the loss weights in the table correspond to the main paper's notation as follows: $\lambda_\text{dssim} \!=\! \lambda_1$, $\lambda_\text{opacity} \!=\! \lambda_2$, $\lambda_\text{dist} \!=\! \lambda_3$, $\lambda_\text{normals} \!=\! \lambda_4$, $\lambda_\text{size} \!=\! \lambda_5$, $\lambda_\text{curv} \!=\! \lambda_6$.

\begin{table}[t]
	\centering
	\caption{\textbf{Hyperparameters used for all scenes.} Parameters marked with $\dagger$ are new compared to TS~\cite{held2025trianglesplatting}; all others follow the same values.}
	\label{tab:hyperparams}
	\footnotesize
	\setlength{\tabcolsep}{4pt}
	\renewcommand{\arraystretch}{0.8}
	\begin{tabular}{lc@{\qquad}lc}
		\toprule
		Parameter                                             & Value                                                       & Parameter                       & Value              \\
		\midrule
		\multicolumn{2}{l}{\textit{Learning rates}}           & \multicolumn{2}{l}{\textit{Loss weights}}                                                                          \\
		\texttt{feature\_lr}                                  & 0.0025                                                      & $\lambda_\text{dssim}$          & 0.2                \\
		\texttt{opacity\_lr}                                  & 0.014                                                       & $\lambda_\text{opacity}$        & 0.0055             \\
		\texttt{lr\_vertices\_init}                           & 0.0018                                                      & $\lambda_\text{normals}$        & $3{\times}10^{-3}$ \\
		\texttt{lr\_sigma}                                    & 0.0008                                                      & $\lambda_\text{size}$           & $10^{-8}$          \\
		\texttt{lr\_ctrl\_pts}$^\dagger$                      & 0.008                                                       & $\lambda_\text{curv}$$^\dagger$ & $10^{-3}$          \\
		\texttt{lr\_delta}$^\dagger$                          & 0.0015                                                      & $\lambda_\text{dist}$           & 0                  \\
			\midrule
		\multicolumn{2}{l}{\textit{Adaptive density control}} & \multicolumn{2}{l}{\textit{Deformation schedule}$^\dagger$}                                                        \\
		\texttt{max\_noise\_factor}                           & 1.5                                                         & \texttt{deform\_delay}          & 3\,000             \\
		\texttt{opacity\_dead}                                & 0.014                                                       & $K$ (ctrl pts per edge)         & 3                  \\
		\texttt{split\_size}                                  & 24                                                          &                                 &                    \\
		\texttt{importance\_threshold}                        & 0.022                                                       &                                 &                    \\
		\bottomrule
	\end{tabular}
\end{table}

\section{Methodology Details}
\label{sec:method_details}

This section provides additional details on the differentiable rendering pipeline that complement the main paper.
We elaborate on the full backward-pass gradient flow through barycentric space (Sec.~\ref{sec:backprop_bary}), the role of curvature regularization (Sec.~\ref{sec:curv_reg}), and tile assignment for deformed primitives (Sec.~\ref{sec:tile_assign}).

\subsection{Backpropagation Through Barycentric Space}
\label{sec:backprop_bary}

The rendering loss $\mathcal{L}$ is evaluated in pixel space, but all signed distance field (SDF) computations occur in barycentric $(u,v)$ space.
Gradients must therefore flow from $\partial\mathcal{L}/\partial\alpha$ back to two sets of parameters: (a)~the 3D vertex positions $\mathbf{v}_i \in \mathbb{R}^3$, and (b)~the scalar control-point displacements $d_{i,k} \in \mathbb{R}$.
Below we trace the full chain.

\paragraph{Step 1: From loss to alpha.}
Standard alpha-compositing backward~\cite{kerbl20233dgs} gives $\partial\mathcal{L}/\partial\alpha$ per pixel per primitive, accumulated over all contributing primitives in back-to-front order.

\paragraph{Step 2: From alpha to $\phi$.}
With $C_x = \phi^{\sigma}$ and $\alpha = \min(0.99,\; o \cdot C_x)$ (Eq.~(6) of the main paper), we use a straight-through estimator for the 0.99 clamp to maintain gradient flow through opaque primitives.
This yields:
\begin{equation}
	\frac{\partial\mathcal{L}}{\partial C_x} = o \;\frac{\partial\mathcal{L}}{\partial\alpha}, \qquad
	\frac{\partial\mathcal{L}}{\partial S} = \frac{\partial\mathcal{L}}{\partial C_x} \cdot \sigma \,\frac{C_x}{\phi} \cdot \frac{1}{R},
	\label{eq:dL_dphi}
\end{equation}
where $R = 2A/P$ is the polygon inradius (Eq.~(5) of the main paper) and $\phi = \min(1, S/R)$.

\paragraph{Step 3: Smooth-minimum fold.}
The gradient $\partial\mathcal{L}/\partial S$ is propagated through the sequential fold
$S_m {=} \mathrm{smin}(S_{m-1}, D_m; g)$
to obtain $\partial\mathcal{L}/\partial D_m$ for each segment~$m$, using the piecewise derivatives of Eq.~(3) of the main paper applied in reverse order.

\paragraph{Step 4: From segment distances to deformed vertices and pixel coordinates.}
Each $D_m$ is the point-to-segment distance from pixel $\mathbf{b} = (u,v)$ to edge $(\mathbf{p}_m, \mathbf{p}_{m+1})$ in barycentric space (Eq.~(2) of the main paper).
Its gradient has two components:
\begin{itemize}
	\item \emph{Vertex component:} $\partial D_m / \partial \mathbf{p}_m$ and $\partial D_m / \partial \mathbf{p}_{m+1}$, distributed by the closest-point parameter $t_m$ as:
	      \begin{equation}
		      \frac{\partial\mathcal{L}}{\partial \mathbf{p}_m} = \frac{\partial\mathcal{L}}{\partial D_m} \cdot (1{-}t_m)\,\hat{\mathbf{r}}_m, \qquad
		      \frac{\partial\mathcal{L}}{\partial \mathbf{p}_{m+1}} = \frac{\partial\mathcal{L}}{\partial D_m} \cdot t_m\,\hat{\mathbf{r}}_m,
		      \label{eq:dL_dvert}
	      \end{equation}
	      where $\hat{\mathbf{r}}_m$ is the unit vector from the pixel to the closest point on segment $m$ and $t_m \in [0,1]$ is the clamped projection parameter.
	\item \emph{Pixel component:} $\partial D_m / \partial \mathbf{b} = -\hat{\mathbf{r}}_m \cdot (\partial\mathcal{L}/\partial D_m)$, accumulated over all $M$ segments into gradients $(\partial\mathcal{L}/\partial u,\; \partial\mathcal{L}/\partial v)$.
\end{itemize}

\paragraph{Step 5, Branch 1: Control-point gradients.}
For each subdivided vertex $\mathbf{p}_m$ (a control point on edge $i$), the gradient accumulated in Eq.~\eqref{eq:dL_dvert} is projected onto the precomputed outward normal $\hat{\mathbf{n}}_i$ of the parent edge (Eq.~(1) of the main paper):
\begin{equation}
	\frac{\partial\mathcal{L}}{\partial d_{i,k}} = \frac{\partial\mathcal{L}}{\partial \mathbf{p}_m} \cdot \hat{\mathbf{n}}_i .
	\label{eq:dL_dcp}
\end{equation}
This is the key simplification enabled by the one-dimensional displacement parameterization: each control point moves only along a fixed direction $\hat{\mathbf{n}}_i$ in barycentric space, so the full 2D vertex gradient reduces to a scalar dot product.

\paragraph{Step 5, Branch 2: 3D vertex gradients.}
The barycentric-coordinate gradients $(\partial\mathcal{L}/\partial u,\; \partial\mathcal{L}/\partial v)$ accumulated from the pixel component are propagated to screen-space vertex positions through the Jacobian of $\mathbf{T}^{-1}$.
Let $\Delta = \det(\mathbf{T}) = e_{1,x}\, e_{2,y} - e_{1,y}\, e_{2,x}$, where $\mathbf{e}_1 = \mathbf{p}_2 - \mathbf{p}_1$ and $\mathbf{e}_2 = \mathbf{p}_3 - \mathbf{p}_1$.
The derivatives of $(u,v)$ with respect to each vertex $\mathbf{p}_j$ involve both the inverse-map coefficients and their dependence on $\Delta$; for $\mathbf{p}_2$ (and analogously $\mathbf{p}_3$):
\begin{equation}
	\frac{\partial u}{\partial \mathbf{p}_{2}} = -\frac{u}{\Delta}\,\frac{\partial \Delta}{\partial \mathbf{p}_2},
	\qquad
	\frac{\partial v}{\partial \mathbf{p}_{2}} = \frac{1}{\Delta}\!\left(\mathbf{e}_p^{\perp} - v\,\frac{\partial \Delta}{\partial \mathbf{p}_2}\right),
	\label{eq:jacobian_v2}
\end{equation}
where $\mathbf{e}_p = \mathbf{p} - \mathbf{p}_1$ and $\perp$ denotes the appropriate signed rotation.
Rather than computing the analogous expressions for $\mathbf{p}_1$ (which depend on \emph{all} three vertices plus the pixel coordinate), we exploit a \textbf{force-balance identity}.
Because barycentric coordinates are translation-invariant (shifting $\mathbf{p}_1, \mathbf{p}_2, \mathbf{p}_3$ and $\mathbf{p}$ by the same vector $\boldsymbol{\epsilon}$ leaves $(u,v)$ unchanged), the total gradient with respect to a uniform translation must vanish:
\begin{equation}
	\frac{\partial\mathcal{L}}{\partial \mathbf{p}_1}
	= -\frac{\partial\mathcal{L}}{\partial \mathbf{p}}
	- \frac{\partial\mathcal{L}}{\partial \mathbf{p}_2}
	- \frac{\partial\mathcal{L}}{\partial \mathbf{p}_3} .
	\label{eq:force_balance}
\end{equation}
This eliminates the most complex Jacobian term and reduces register usage in the CUDA kernel.
Finally, each image-space gradient $\partial\mathcal{L}/\partial \mathbf{p}_j^{\text{screen}}$ is mapped to $\partial\mathcal{L}/\partial \mathbf{v}_j^{3\mathrm{D}}$ via the standard perspective-projection backward pass~\cite{kerbl20233dgs}.

\subsection{Curvature Regularization}
\label{sec:curv_reg}

The curvature loss (Eq.~(7) of the main paper, $\mathcal{L}_{\text{curv}} = \overline{\|\mathbf{d}\|^2}$) is a simple $\ell^2$ penalty on all control-point displacements.
Its role is to prevent a positive-feedback loop: as displacements grow, the deformed polygon's area $A$ increases, inflating the normalization radius $R = 2A/P$; a larger $R$ makes $\phi = S/R$ smaller in the interior, which the optimizer compensates by pushing displacements even further out.
The $\ell^2$ penalty counteracts this by penalizing large $|d_{i,k}|$ directly.
Without it, the fraction of control points with $|d|{>}0.1$ rises from 29\% to 36\% and the mean displacement magnitude increases by 23\%, as reported in the ablation study of the main paper (see Table~3).

\subsection{Tile Assignment for Deformed Primitives}
\label{sec:tile_assign}

Tile-based rasterizers~\cite{kerbl20233dgs} assign each primitive to screen tiles via a bounding circle, typically defined by the circumradius of the projected triangle.
When control points deform the boundary beyond the base triangle, this region may miss affected pixels.

Our solution computes a tight conservative bound during preprocessing: each of the $M = 3(K{+}1)$ deformed polygon vertices $\mathbf{b}_m$ is converted to screen space via
$\mathbf{p}_m^{\text{scr}} = \mathbf{p}_1 + u_m(\mathbf{p}_2 {-} \mathbf{p}_1) + v_m(\mathbf{p}_3 {-} \mathbf{p}_1)$,
and a running axis-aligned bounding box (AABB) is maintained.
We then compute the maximum distance from the centroid to any deformed screen-space vertex:
\begin{equation}
	r_{\text{def}} = \max_{m=\{1,\ldots,M\}} \left\| \mathbf{p}_m^{\text{scr}} - \bar{\mathbf{p}} \right\|,
	\label{eq:tile_radius}
\end{equation}
where $\bar{\mathbf{p}}$ is the image-space centroid.
The tile-assignment radius is then
\(
r_{\text{tile}} = \max(r_{\text{circ}},\; r_{\text{def}})
\),
ensuring all deformed pixels are covered; the cost is $\mathcal{O}(M)$ per primitive during preprocessing.

\section{Robustness and Efficiency Analysis}
\label{sec:robustness}

This section complements the main paper with three analyses: the per-pixel arithmetic cost of our rasterizer (Sec.~\ref{sec:flops}), stability under self-intersecting boundaries (Sec.~\ref{sec:selfint}), and the ability to fit asymmetric shapes for both even and odd $K$ (Sec.~\ref{sec:evenodd}).

\begin{figure}[t!]
	\centering
	\setlength{\tabcolsep}{1pt}
	\renewcommand{\arraystretch}{0.5}
	\newcommand{\rimg}[1]{\includegraphics[width=0.11\linewidth]{#1}}
	\begin{tabular}{@{}cccc@{\hspace{7mm}}cccc@{}}
		\scriptsize Target & \scriptsize $K{=}2$ & \scriptsize $K{=}3$ & \scriptsize $K{=}4$ &
		\scriptsize Target & \scriptsize iter\,0 & \scriptsize iter\,1k & \scriptsize iter\,5k \\
		\rimg{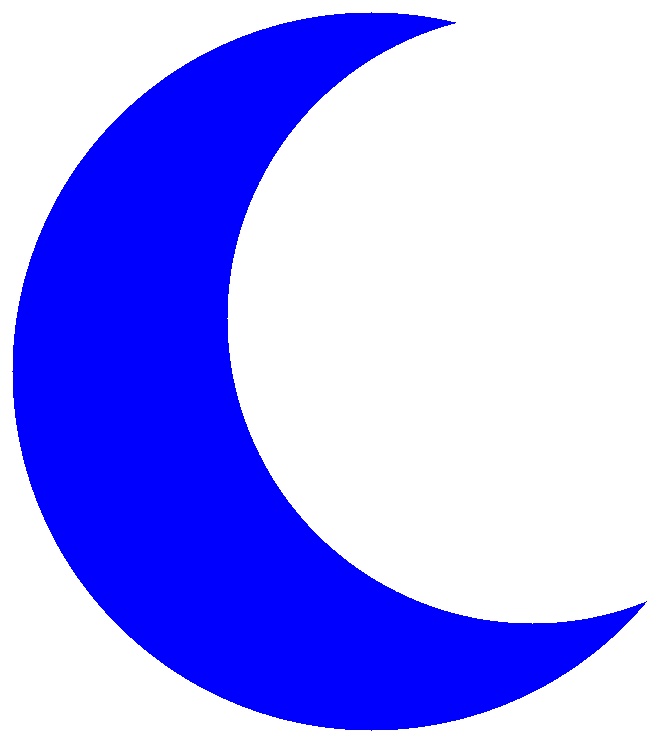} &
		\rimg{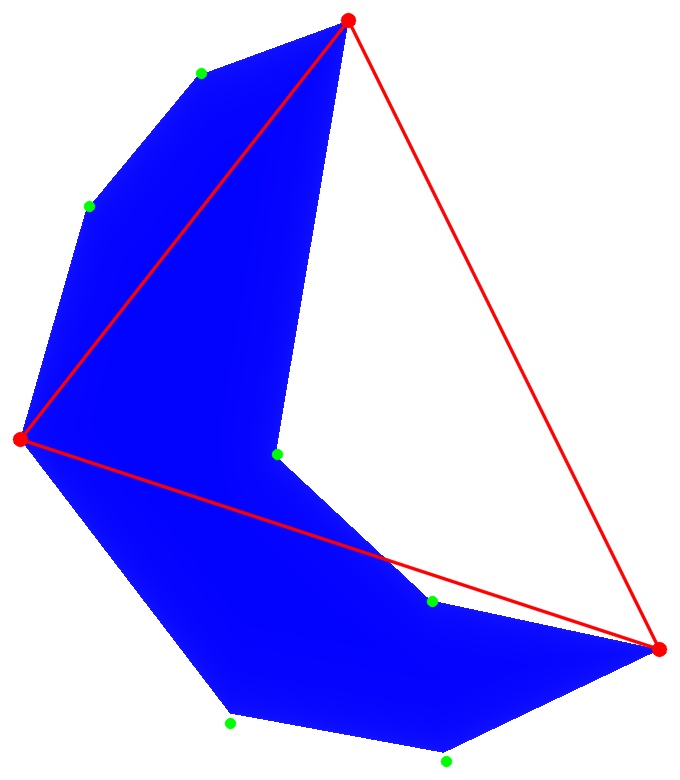} &
		\rimg{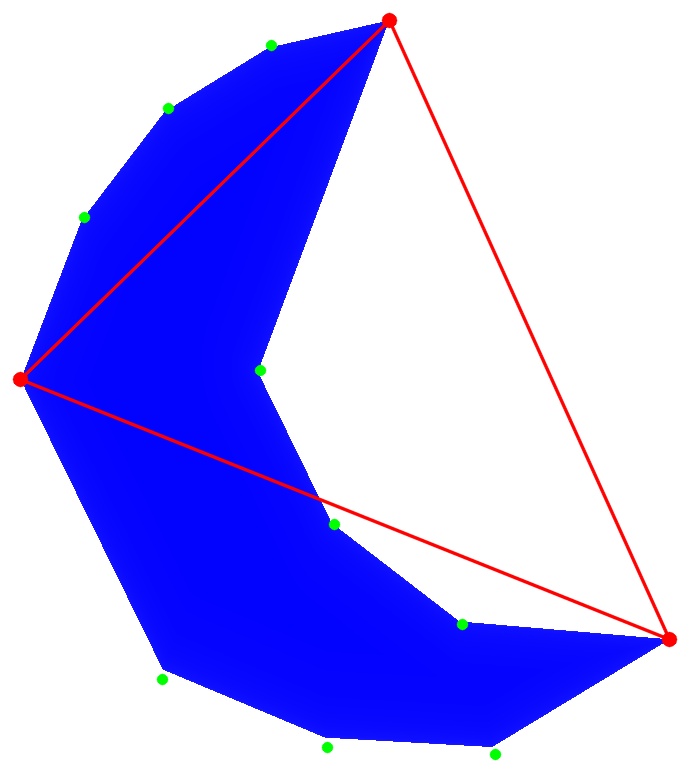} &
		\rimg{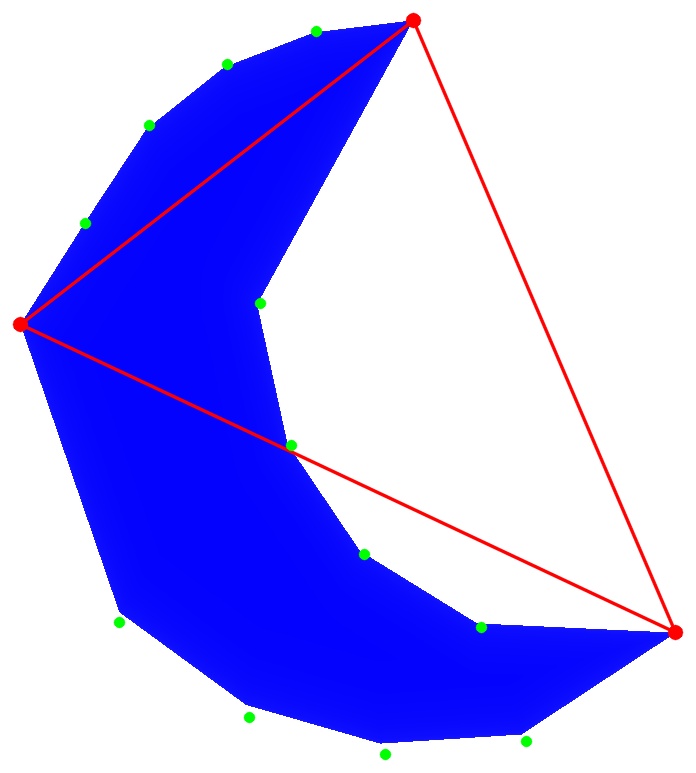} &
		\rimg{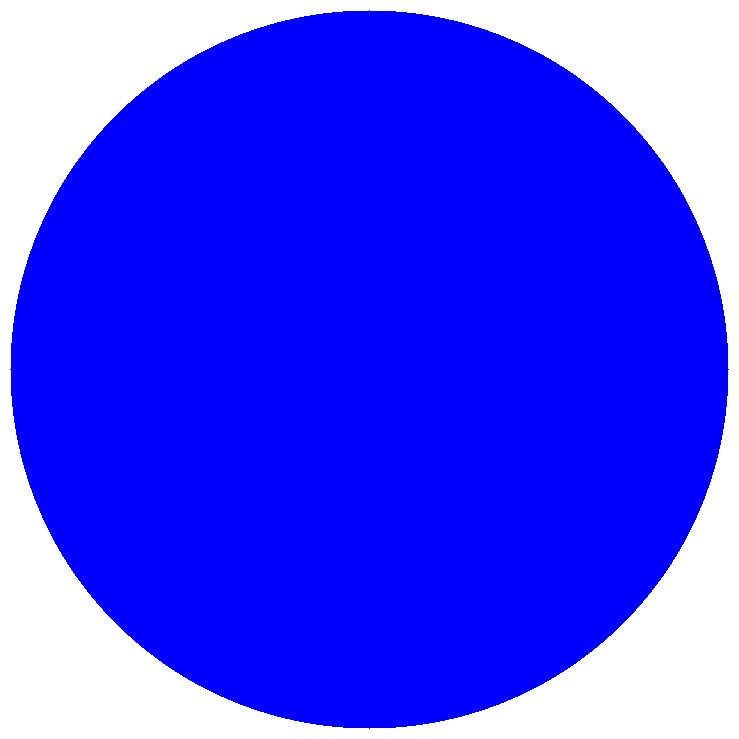} &
		\rimg{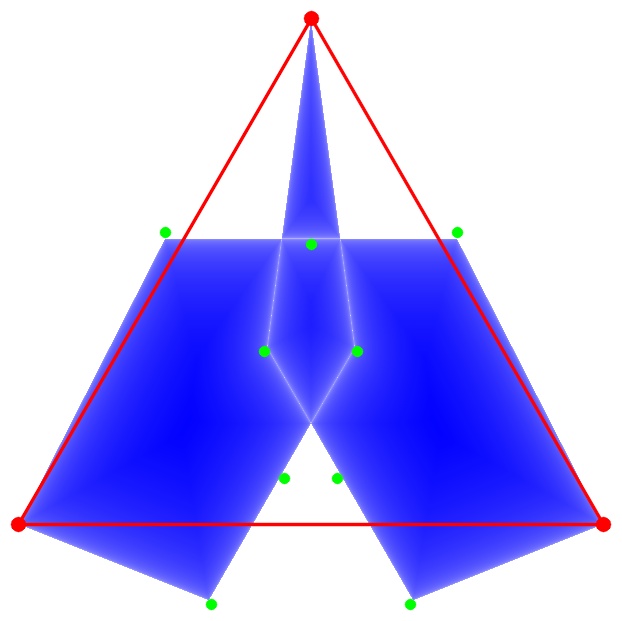} &
		\rimg{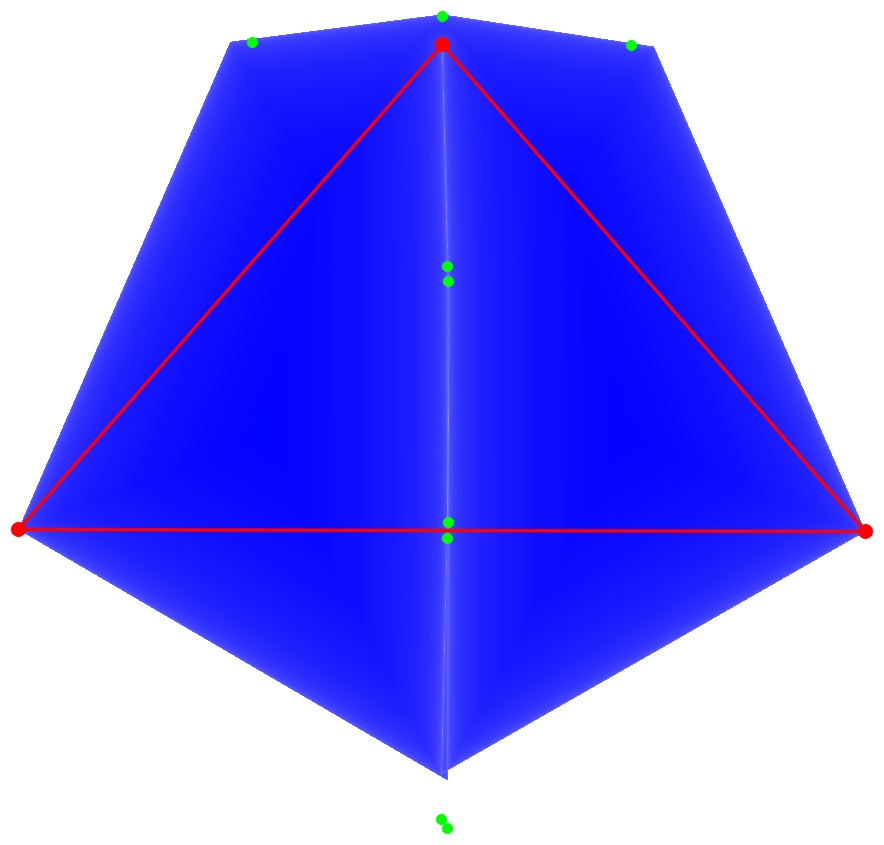} &
		\rimg{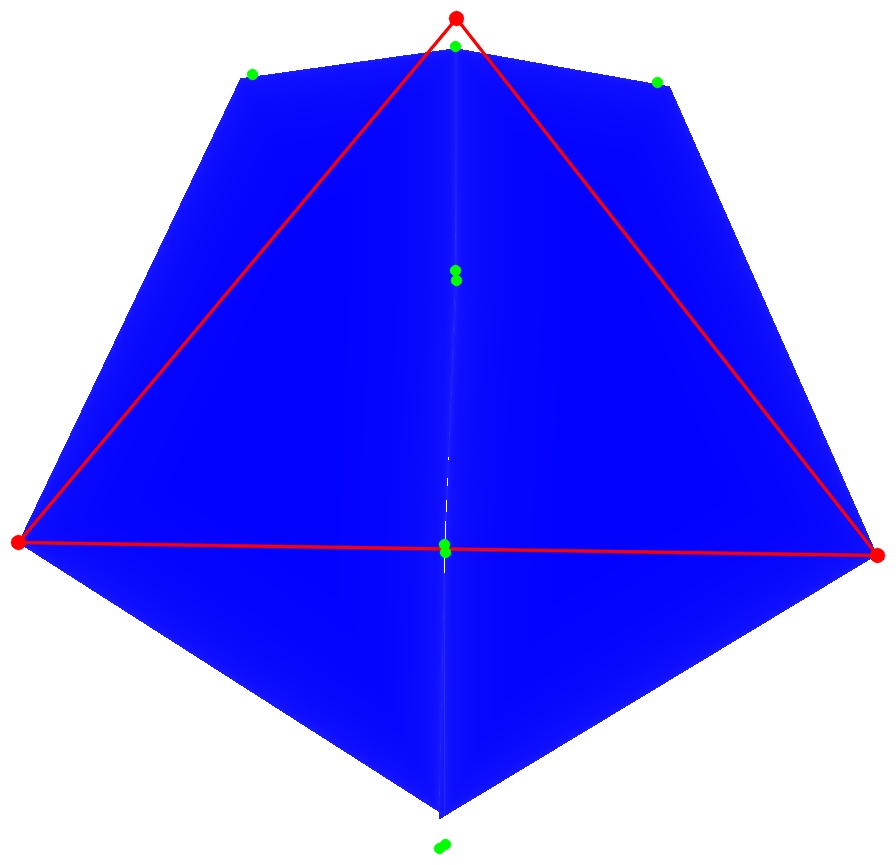} \\
		\multicolumn{4}{c}{\scriptsize Crescent fit vs.\ $K$} &
		\multicolumn{4}{c}{\scriptsize Self-intersection recovery} \\
	\end{tabular}
	\caption{\textbf{Robustness analyses.} \textbf{Left:}~A single deformable primitive fitting a crescent (which lacks edge-midpoint symmetry) at $K{=}2,3,4$; the fit improves monotonically (L1 $0.0185/0.0137/0.0112$) and the even case $K{=}4$ outperforms the odd $K{=}3$. \textbf{Right:}~Self-intersection stress test: a primitive initialized in a severely folded state ($d{=}{-}0.40$) recovers to a clean convex boundary while fitting a circle (loss $0.185{\to}0.035$ over 5k iterations).}
	\label{fig:supp_robust}
\end{figure}

\subsection{Per-Pixel Computational Cost}
\label{sec:flops}
Table~1 of the main paper reports the wall-clock cost as a function of $K$; here we break it down at the arithmetic level.
At $K{=}3$, classifying and shading a single interior pixel requires a winding-number test, $M{=}3(K{+}1){=}12$ point-to-segment distance evaluations, and the sequential smooth-minimum fold, totaling ${\sim}380$ floating-point operations (FLOP) per inside pixel, compared to ${\sim}76$ for the half-plane test of TS~\cite{held2025trianglesplatting}.
Because all remaining pipeline stages (projection, tile binning, sorting, and $\alpha$-blending) are shared, this ${\sim}5\times$ arithmetic increase translates to only a ${\sim}2\times$ slowdown in practice: at $K{=}3$ our method reaches $47$/$33$ FPS on Bonsai/Garden on the same RTX\,5090 used to benchmark TS~\cite{held2025trianglesplatting}.
The overhead is governed entirely by $K$ and, as shown in Table~1, quality already saturates at $K{=}2$--$3$, so the cost can be tuned without sacrificing fidelity.

\subsection{Stability Under Self-Intersection}
\label{sec:selfint}
Because a control point may displace a boundary vertex past one of its neighbors, a deformed polygon can become self-intersecting.
Our pipeline tolerates this gracefully through three mechanisms: the non-zero winding-number rule, which assigns a consistent inside/outside label even for overlapping regions; the $C^1$ polynomial smooth minimum, which keeps the distance field differentiable across folds; and the curvature regularizer $\mathcal{L}_\text{curv}$ (Eq.~(7) of the main paper), which discourages extreme displacements.
At convergence, $8\%$ and $13\%$ of the primitives in Bonsai and Garden are self-intersecting, yet training never diverges.
To stress-test this behavior, we initialize a single primitive in a severely folded configuration ($d{=}{-}0.40$) and optimize it to fit a circle (Fig.~\ref{fig:supp_robust}-right): the loss drops from $0.185$ to $0.035$ within 5k iterations and the polygon unfolds into a clean convex boundary.

\subsection{Asymmetric Shapes and Even/Odd $K$}
\label{sec:evenodd}
With control points placed at $t_s{=}(s{+}1)/(K{+}1)$ along each edge, odd $K$ positions a control point exactly at the edge midpoint while even $K$ does not, which might suggest a bias toward symmetric shapes.
In practice this is not a limitation: because the three triangle vertices are optimized jointly with the displacements, each primitive freely rotates and translates in 3D to align its control-point slots with the target geometry.
We verify this on a crescent target, which has no $t_s$-symmetry, at $K{\in}\{2,3,4\}$ (Fig.~\ref{fig:supp_robust}-left): the fit improves monotonically with $K$ (L1 $=0.0185$/$0.0137$/$0.0112$), and the even case $K{=}4$ outperforms the odd $K{=}3$.
The asymmetric Pac-Man target in Fig.~6 of the main paper likewise fits cleanly at every $K$.

\clearpage
\section{Per-Scene Results}
\label{sec:per_scene}

Table~\ref{tab:per_scene} reports per-scene metrics on Mip-NeRF 360~\cite{barron2022mipnerf360} and Tanks\,\&\,Temples~\cite{knapitsch2017tanks} datasets. Our method matches or outperforms Triangle Splatting~\cite{held2025trianglesplatting} on nine of the eleven scenes; the two exceptions are Stump and Treehill, where the additional deformation degrees of freedom provide limited benefit on large, roughly planar foliage regions.

\begin{table}[t!]
	\centering
	\caption{Per-scene results on Mip-NeRF 360~\cite{barron2022mipnerf360} (outdoor and indoor) and Tanks \& Temples~\cite{knapitsch2017tanks}. \colorbox{orange!25}{\textbf{Orange}} cells indicate where our method outperforms Triangle Splatting~\cite{held2025trianglesplatting}.}
	\label{tab:per_scene}
	\footnotesize
	\setlength{\tabcolsep}{4pt}
	\renewcommand{\arraystretch}{0.80}
	\resizebox{\textwidth}{!}{
		\begin{tabular}{l|l|ccccc|cccc|cc}
			\toprule
			       &                    & \multicolumn{5}{c|}{Outdoor Mip-NeRF 360} & \multicolumn{4}{c|}{Indoor Mip-NeRF 360} & \multicolumn{2}{c}{Tanks \& Temples}                                                                                                                       \\
			Metric & Method             & Bicycle                                   & Flowers                                  & Garden                               & Stump          & Treehill       & Room        & Counter     & Kitchen     & Bonsai      & Truck       & Train       \\
			\midrule
			\multirow{2}{*}{PSNR$\uparrow$}
			       & Triangle Splatting & 24.90                                     & 20.85                                    & 27.20                                & \textbf{26.29} & \textbf{21.94} & 31.05       & 28.90       & 31.32       & 31.95       & 24.94       & 21.33       \\
			       & Ours               & \win{25.10}                               & \win{20.93}                              & \win{27.38}                          & 26.28          & 21.69          & \win{31.34} & \win{29.26} & \win{31.48} & \win{32.44} & \win{25.21} & \win{21.41} \\
			\midrule
			\multirow{2}{*}{SSIM$\uparrow$}
			       & Triangle Splatting & 0.765                                     & 0.614                                    & 0.863                                & 0.759          & \textbf{0.611} & 0.926       & 0.911       & 0.929       & 0.947       & 0.889       & 0.823       \\
			       & Ours               & \win{0.774}                               & \win{0.617}                              & \win{0.869}                          & \win{0.764}    & 0.602          & \win{0.927} & \win{0.915} & \win{0.930} & \win{0.948} & \win{0.891} & \win{0.829} \\
			\midrule
			\multirow{2}{*}{LPIPS$\downarrow$}
			       & Triangle Splatting & 0.190                                     & 0.284                                    & 0.106                                & 0.214          & 0.289          & 0.186       & 0.171       & 0.115       & 0.169       & 0.108       & 0.179       \\
			       & Ours               & \win{0.178}                               & \win{0.266}                              & \win{0.097}                          & \win{0.204}    & \win{0.283}    & \win{0.183} & \win{0.163} & \win{0.113} & \win{0.162} & \win{0.100} & \win{0.166} \\
			\bottomrule
		\end{tabular}
	}
\end{table}

\begin{figure}[H]
	\centering
	\newcommand{\fimgd}[1]{\includegraphics[width=\linewidth]{#1}}
	\newcommand{\rlbld}[1]{\rotatebox[origin=c]{90}{\tiny #1}}
	\setlength{\tabcolsep}{0.5pt}
	\renewcommand{\arraystretch}{0}
	\begin{tabular}{@{} >{\centering\arraybackslash}m{5mm} @{\,}
		>{\centering\arraybackslash}m{0.235\linewidth} @{\,}
		>{\centering\arraybackslash}m{0.235\linewidth} @{\,}
		>{\centering\arraybackslash}m{0.235\linewidth} @{\,}
		>{\centering\arraybackslash}m{0.235\linewidth} @{}}
		                                                           & \scriptsize GT & \scriptsize Full ($K{=}3$) & \scriptsize Rigid ($d{=}0$) & \scriptsize $|\text{Full}{-}\text{Rigid}|$ \\[2pt]
		\rlbld{Bonsai}                                             &
		\fimgd{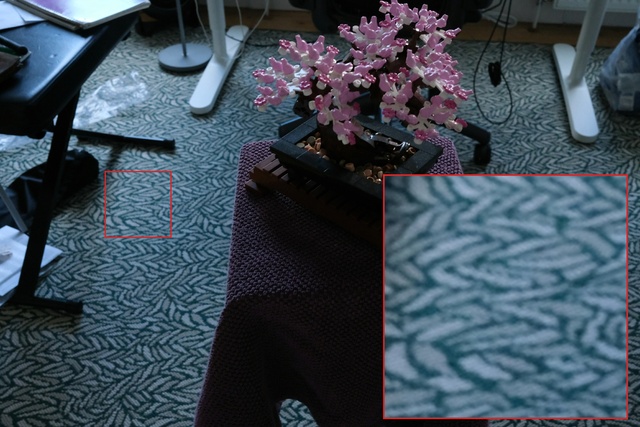}     &
		\fimgd{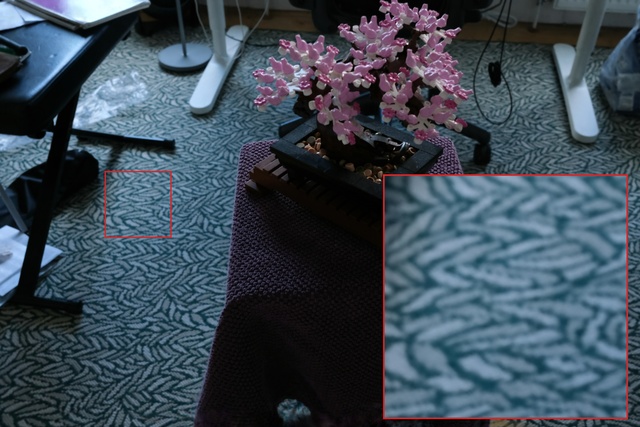}   &
		\fimgd{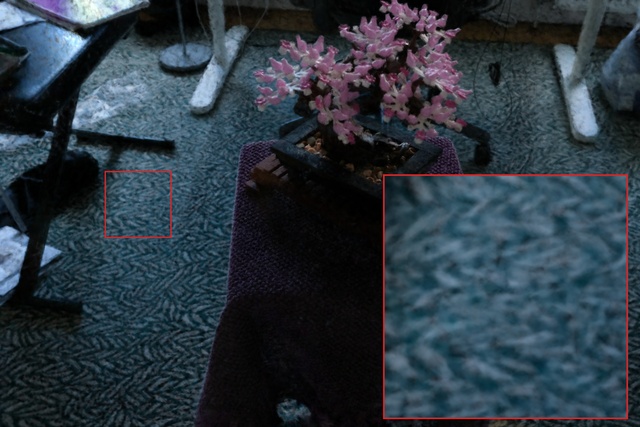}  &
		\fimgd{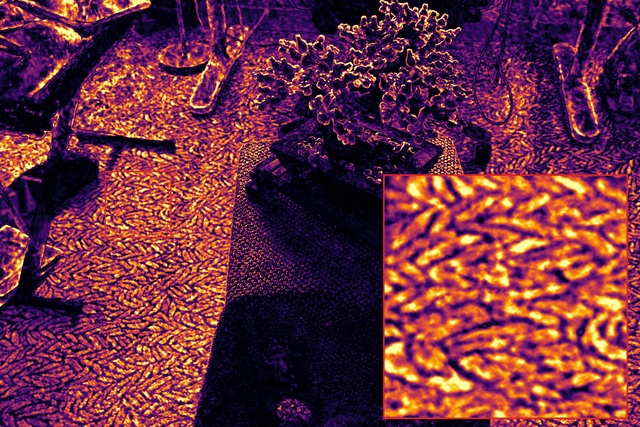}                                                                                                                         \\[2pt]
		\rlbld{Bicycle}                                            &
		\fimgd{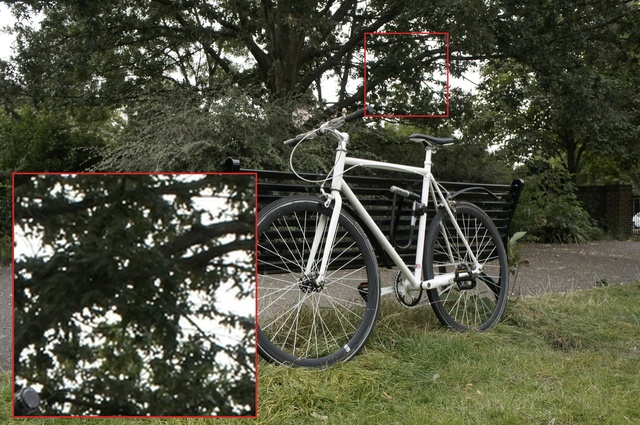}    &
		\fimgd{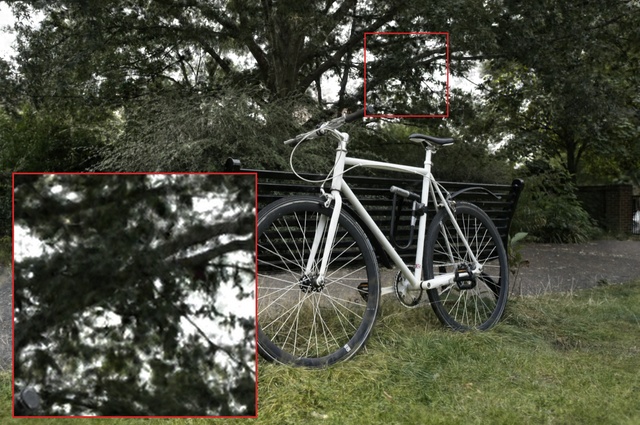}  &
		\fimgd{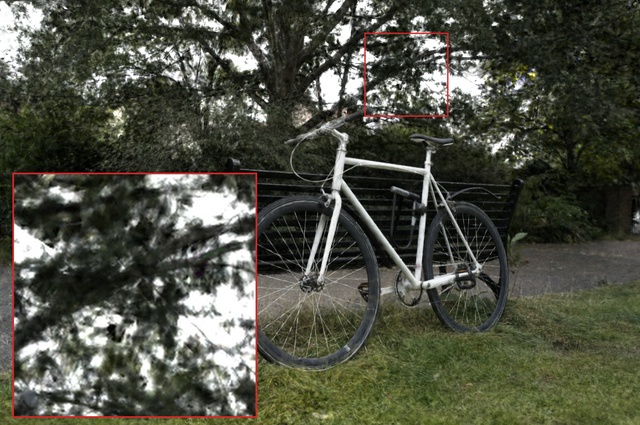} &
		\fimgd{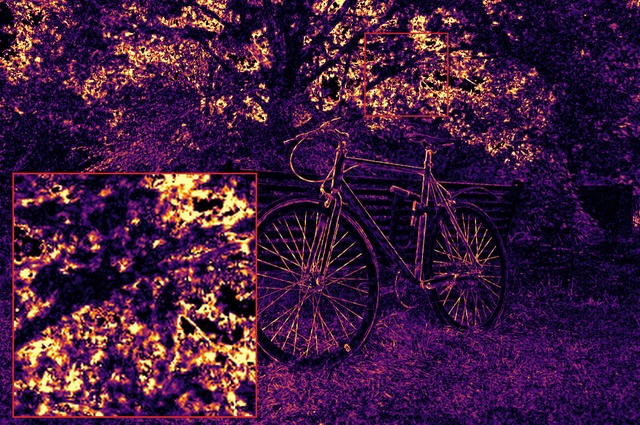}                                                                                                                        \\
	\end{tabular}
	\caption{\textbf{Deformation contribution maps (additional scenes).} For each scene we render the trained model with learned control-point displacements active (\emph{Full}) and with all displacements and corner smoothness $\delta$ zeroed (\emph{Rigid}, $d{=}0$), keeping all other parameters unchanged. The rightmost column shows the per-pixel absolute difference $|\text{Full}{-}\text{Rigid}|$ (inferno colormap, percentile-normalized), isolating the visual contribution of the learned boundary deformations. Red insets highlight the patch of highest deformation activity. Leaf edges and object silhouettes in Bonsai (top) and spoke tips and wheel rim in Bicycle (bottom)---the same regions that drive our LPIPS improvement over TS~\cite{held2025trianglesplatting}.}
	\label{fig:supp_deformation_contribution}
\end{figure}

\clearpage
\section{Extended Qualitative Evaluation}
\begin{figure}[H]
	\centering
	\newcommand{\fimgq}[1]{\includegraphics[width=\linewidth]{#1}}
	\newcommand{\rlblq}[1]{\rotatebox[origin=c]{90}{\tiny #1}}
	\setlength{\tabcolsep}{0.5pt}
	\renewcommand{\arraystretch}{0}
	\begin{tabular}{@{} >{\centering\arraybackslash}m{5mm} @{\,} >{\centering\arraybackslash}m{0.235\linewidth} @{\,} >{\centering\arraybackslash}m{0.235\linewidth} @{\,} >{\centering\arraybackslash}m{0.235\linewidth} @{\,} >{\centering\arraybackslash}m{0.235\linewidth} @{}}
		                                                      & \scriptsize GT & \scriptsize TS~\cite{held2025trianglesplatting} & \scriptsize 3DGS~\cite{kerbl20233dgs} & \scriptsize Ours \\[2pt]
		\rlblq{Bonsai}                                                     &
		\fimgq{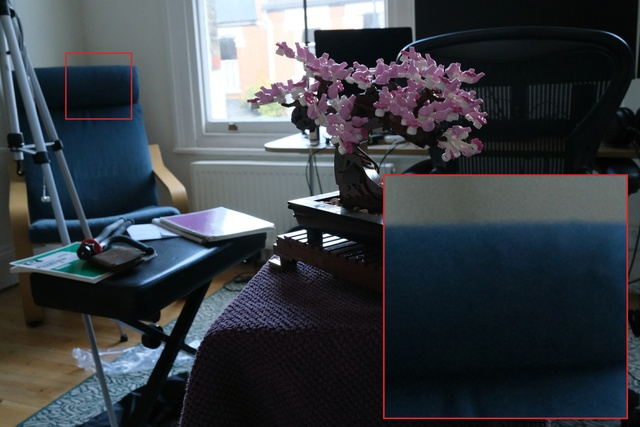}  &
		\fimgq{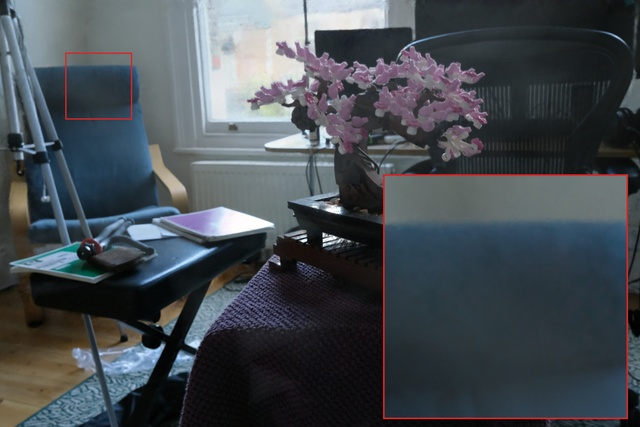}  &
		\fimgq{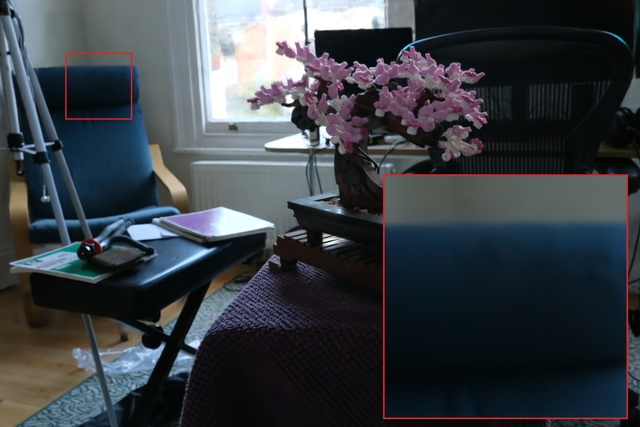}  &
		\fimgq{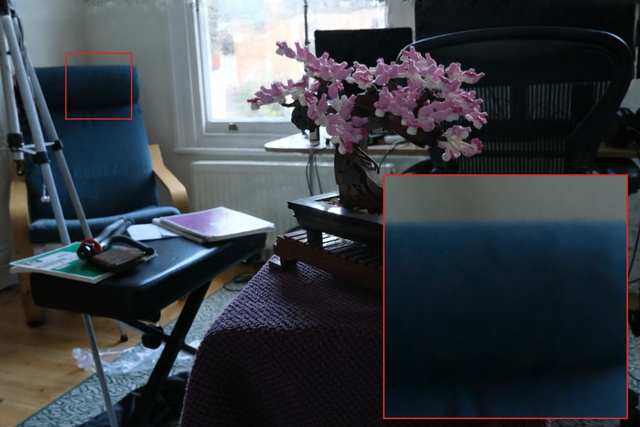}  \\
		\rlblq{Kitchen}                                                    &
		\fimgq{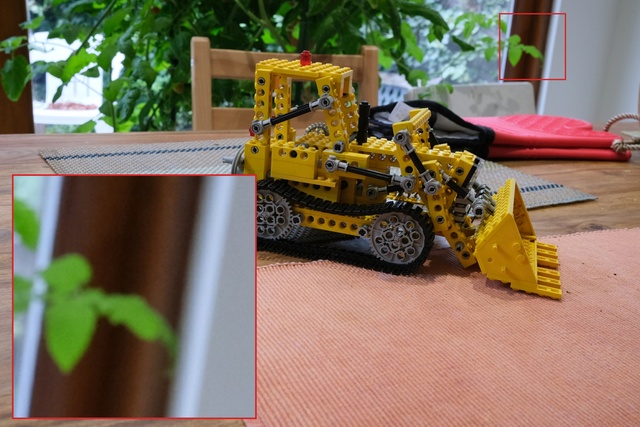} &
		\fimgq{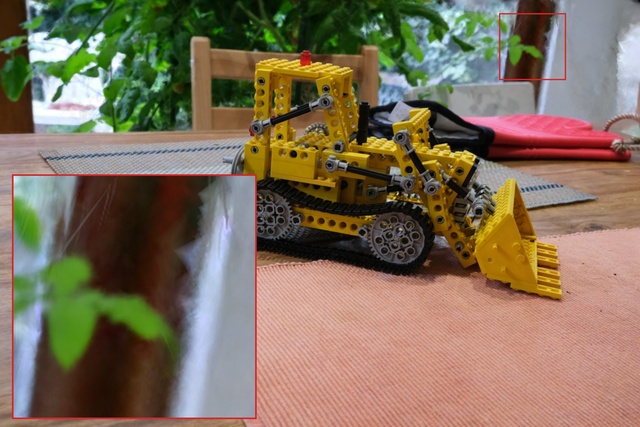} &
		\fimgq{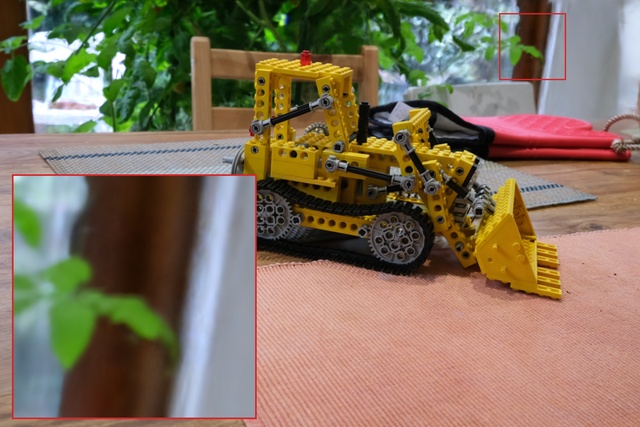} &
		\fimgq{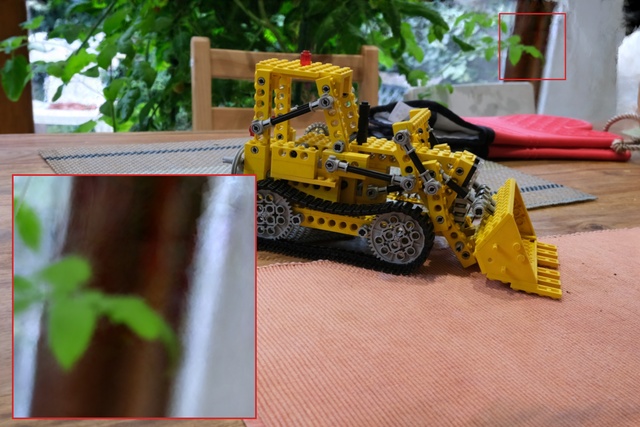} \\
		\rlblq{Room}                                                       &
		\fimgq{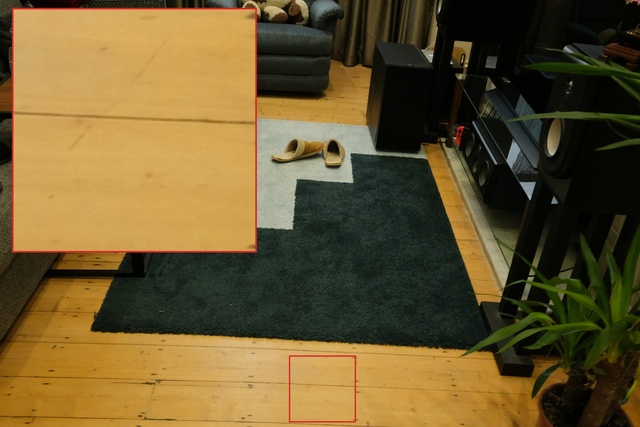}    &
		\fimgq{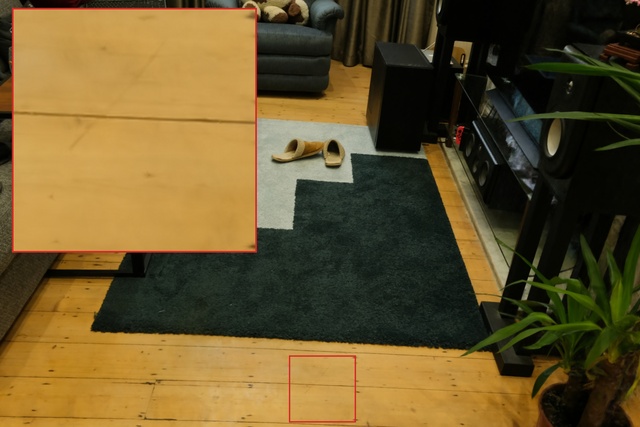}    &
		\fimgq{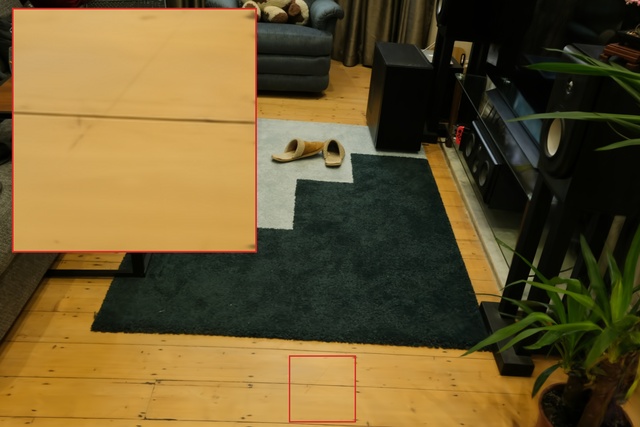}    &
		\fimgq{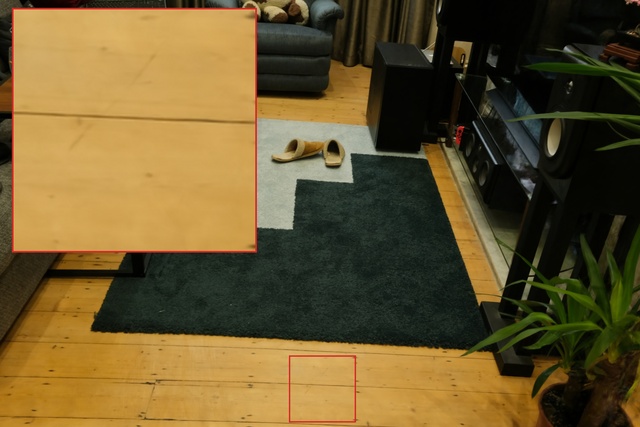}    \\
		\rlblq{Garden}                                                     &
		\fimgq{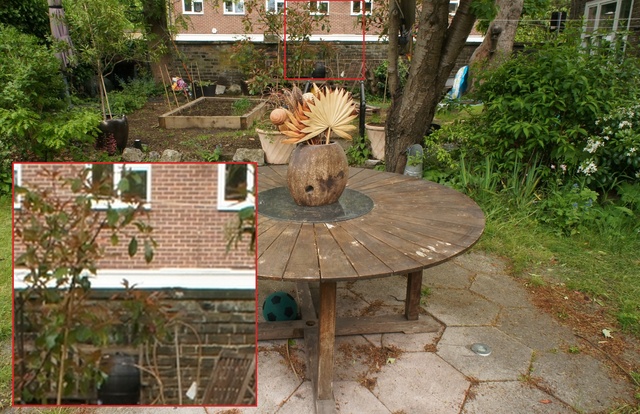}  &
		\fimgq{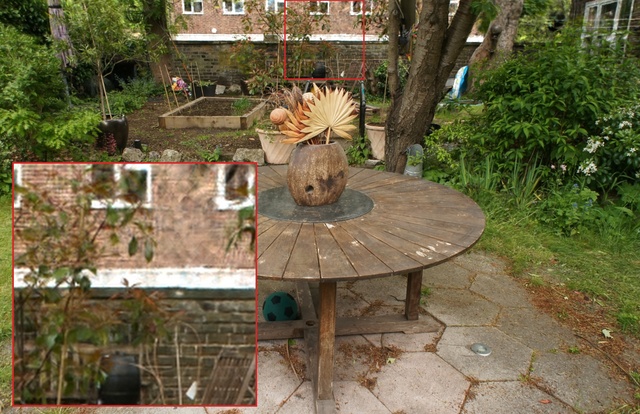}  &
		\fimgq{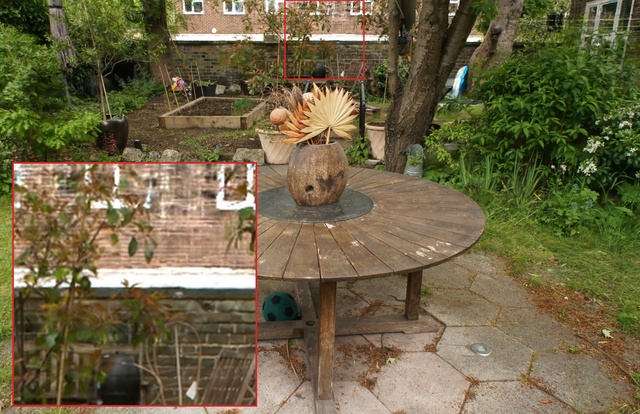}  &
		\fimgq{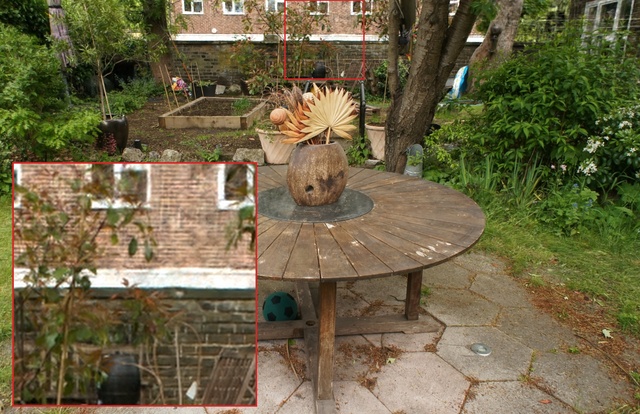}  \\
	\end{tabular}
	\caption{\textbf{Additional qualitative comparisons.} Held-out test views on four scenes not shown in the main paper, at full resolution. Each row shows the ground truth, TS~\cite{held2025trianglesplatting}, 3DGS~\cite{kerbl20233dgs}, and ours; red insets zoom into the patch where our method shows the largest improvement over TS.}
	\label{fig:supp_qualitative}
\end{figure}

Figures~\ref{fig:supp_deformation_contribution} and~\ref{fig:supp_qualitative} provide additional full-resolution renders on scenes not shown in the main paper, complementing the qualitative comparisons there to facilitate detailed visual inspection.

\clearpage
\section{Geometry Evaluation on DTU}
\label{sec:dtu_geometry}

Following TS~\cite{held2025trianglesplatting} and 2DGS~\cite{huang20242dgs}, we evaluate mesh reconstruction quality on the DTU dataset~\cite{jensen2014large} by applying Truncated Signed Distance Function (TSDF) fusion to predicted depth maps, followed by marching cubes and per-scene camera-mask culling.
We report the Chamfer distance (CD) averaged over the standard 15 scenes.

Following standard practice, we adopt dataset-specific hyperparameters for DTU, consistent with the evaluation protocol used by all compared methods.

Table~\ref{tab:dtu_cd} compares our method against five baselines.
Our method achieves CD\,$=$\,1.01, improving over TS~\cite{held2025trianglesplatting} (1.06) and substantially outperforming 3DGS~\cite{kerbl20233dgs} (1.96).
While the geometry-focused 2DGS~\cite{huang20242dgs}  (0.80) and BBSplat~\cite{svitov2024bbsplat}  (0.91) achieve lower CD through specialized depth/normal supervision, our method is competitive while using a simpler TSDF-based extraction pipeline identical to TS~\cite{held2025trianglesplatting}, obtaining a better joint estimate.

Figure~\ref{fig:dtu_mesh} shows extracted meshes for three representative scans.
The gray renders highlight geometric surface quality, while the vertex-colored renders demonstrate that the TSDF fusion preserves appearance information.
Architectural details (scan\,24), organic shapes (scan\,55), and fine sculptural features (scan\,114) are all faithfully reconstructed.

\begin{table}[H]
	\centering
	\caption{Chamfer distance (CD\,$\downarrow$) on the DTU dataset~\cite{jensen2014large}, averaged over 15 scenes.}
	\label{tab:dtu_cd}
	\small
	\setlength{\tabcolsep}{8pt}
	\renewcommand{\arraystretch}{0.9}
	\begin{tabular}{lc}
		\toprule
		Method                              & CD $\downarrow$ \\
		\midrule
		3DGS~\cite{kerbl20233dgs}           & 1.96            \\
		TS~\cite{held2025trianglesplatting} & 1.06            \\
		BBSplat~\cite{svitov2024bbsplat}    & 0.91            \\
		2DGS~\cite{huang20242dgs}           & \textbf{0.80}   \\
		Ours                                & 1.01            \\
		\bottomrule
	\end{tabular}
\end{table}

\vspace{-8pt}
\begin{figure}[H]
	\centering
	\newcommand{\fimgm}[1]{\includegraphics[height=0.092\textheight]{#1}}
	\setlength{\tabcolsep}{1pt}
	\renewcommand{\arraystretch}{0}
	\begin{tabular}{@{} c @{\;\;} c @{\;\;} c @{}}
		\scriptsize Scan 24                       & \scriptsize Scan 55 & \scriptsize Scan 114 \\[1pt]
		\fimgm{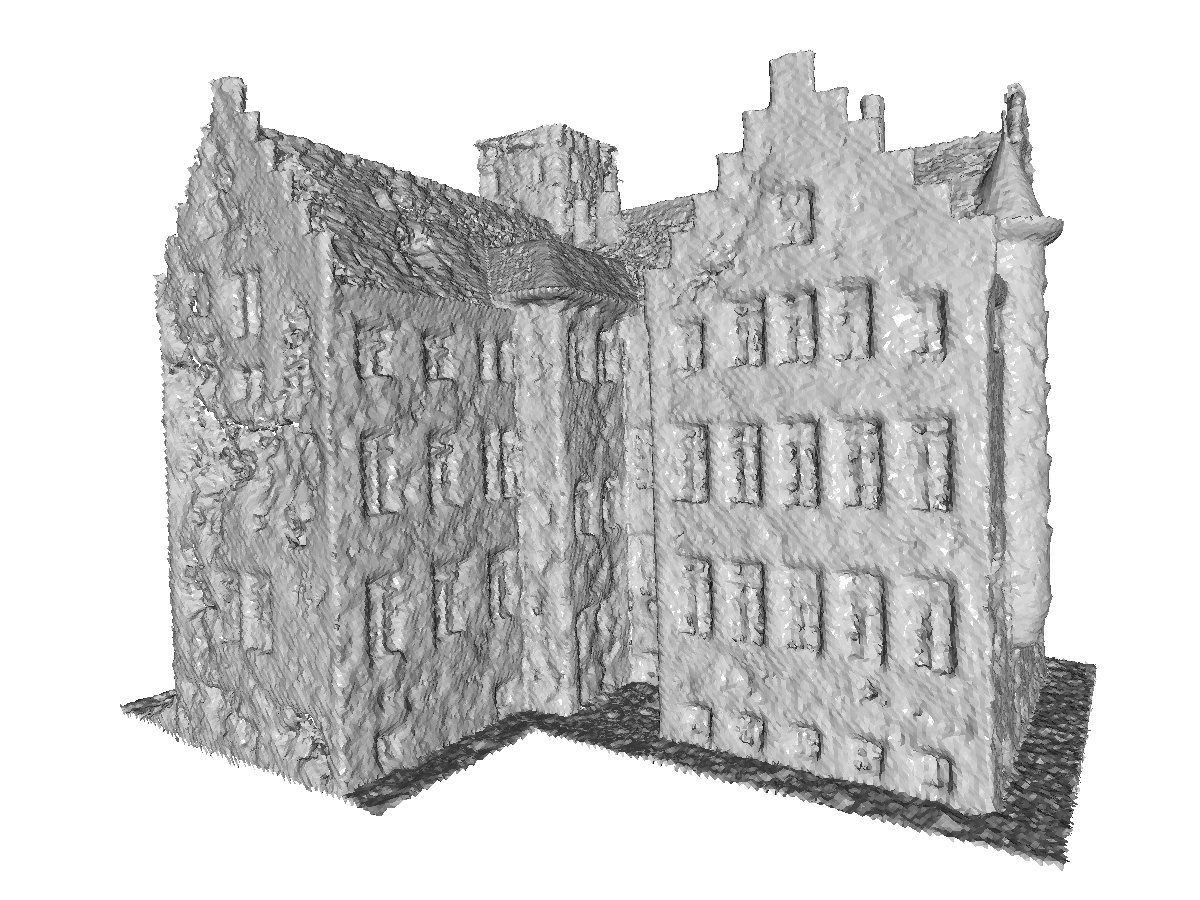}  &
		\fimgm{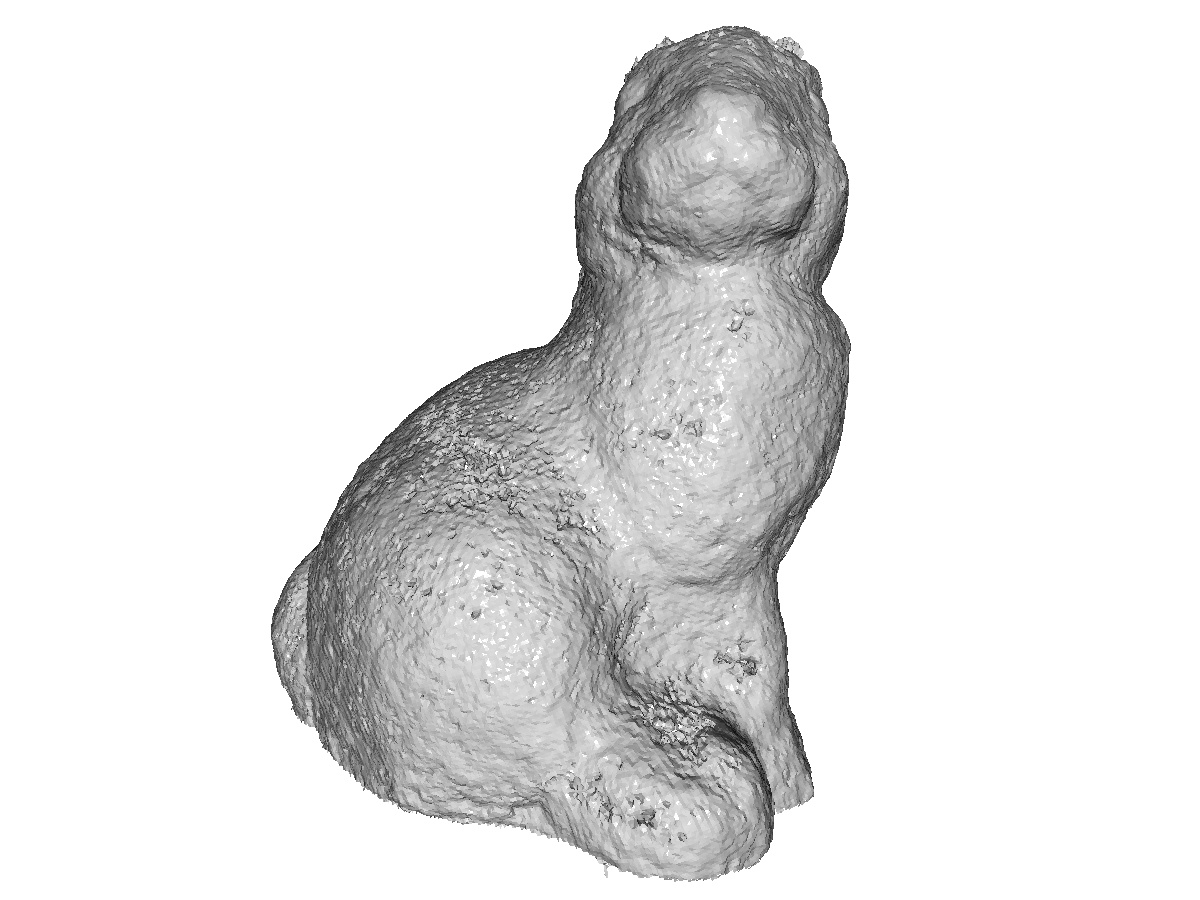}  &
		\fimgm{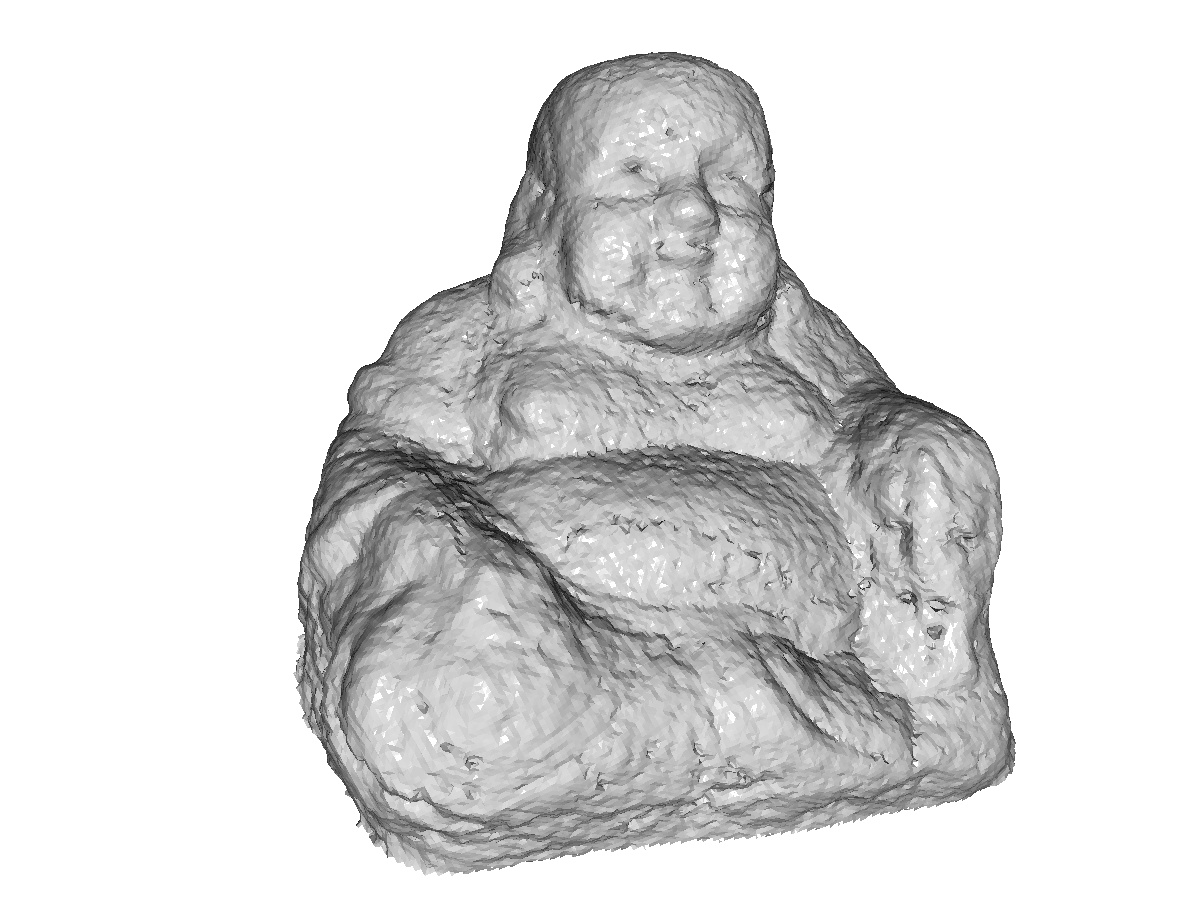}                                              \\[1pt]
		\fimgm{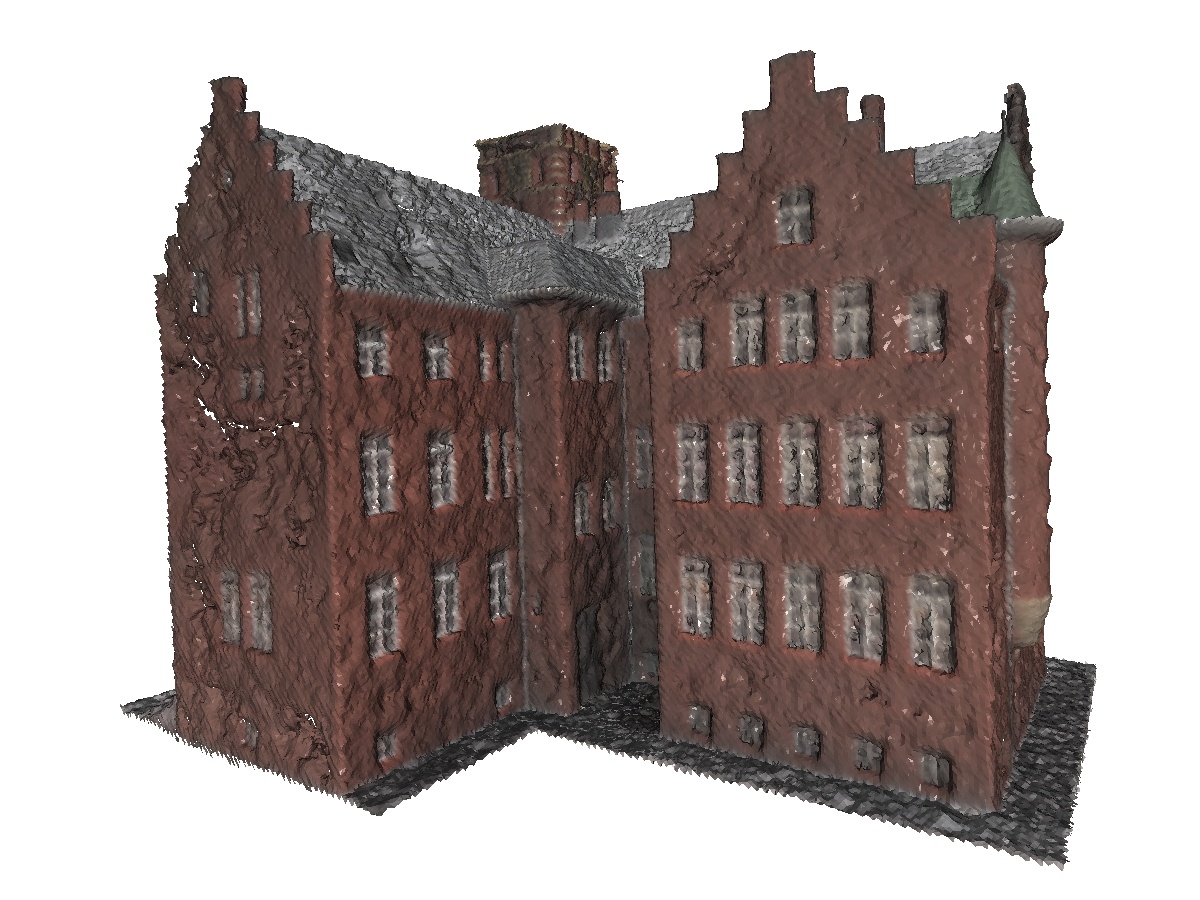} &
		\fimgm{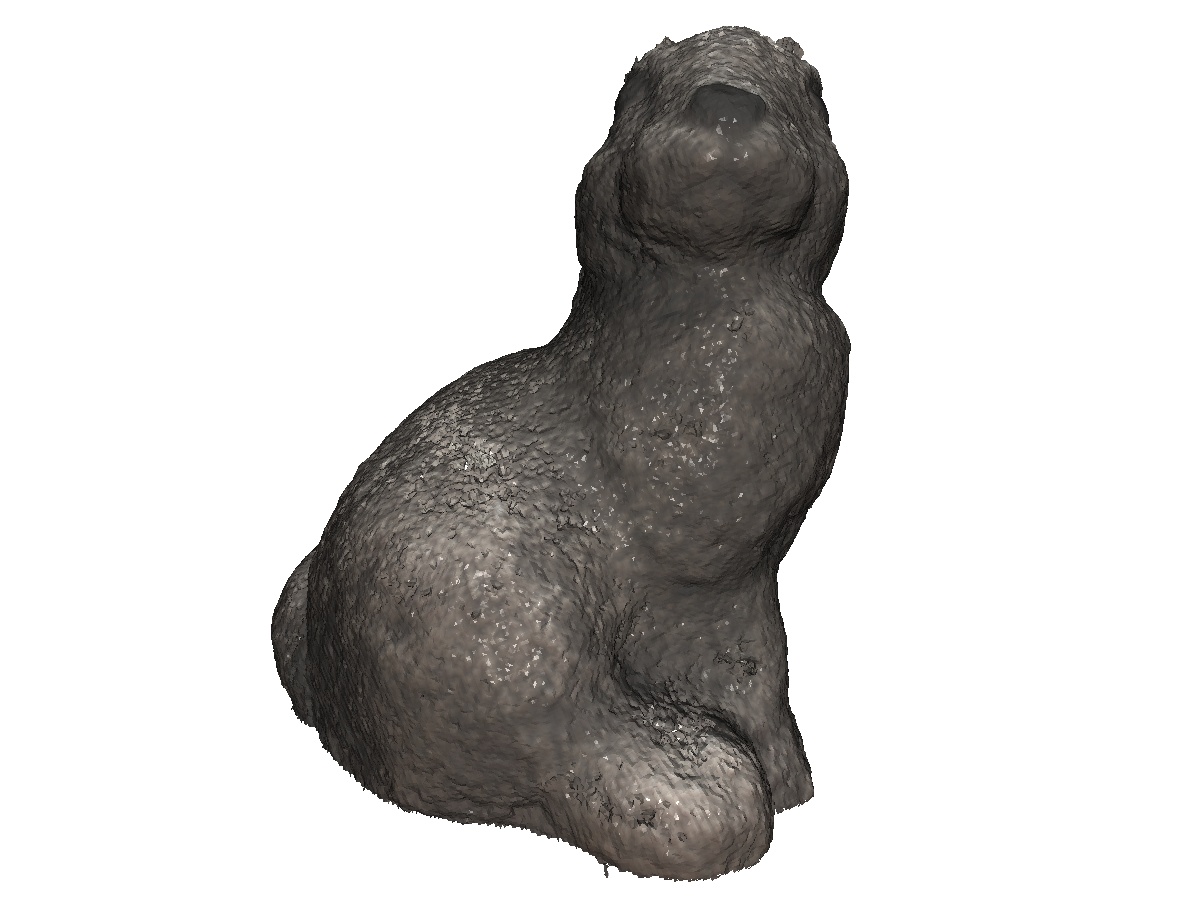} &
		\fimgm{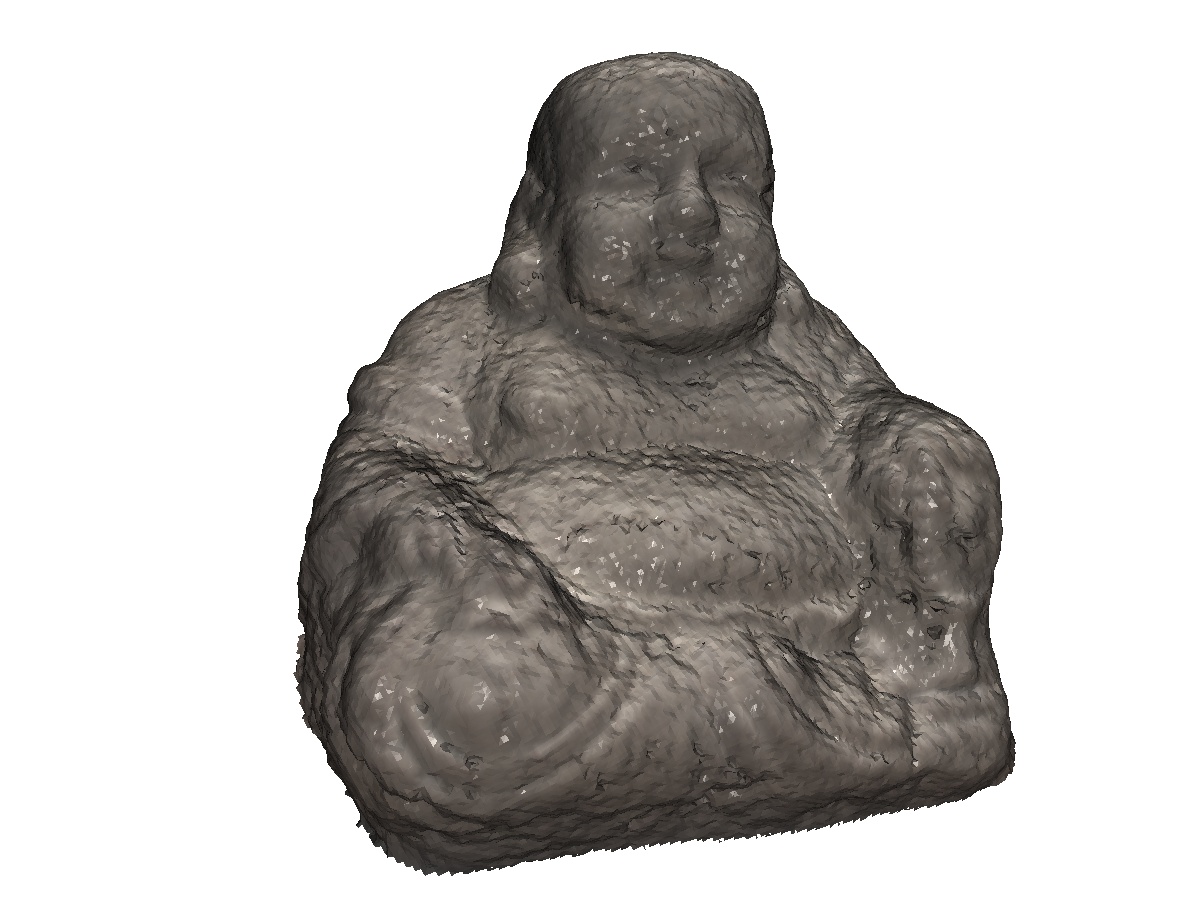}                                             \\
	\end{tabular}
	\caption{\textbf{Mesh extraction on DTU~\cite{jensen2014large}.} We extract meshes via TSDF fusion of predicted depth maps, following 2DGS~\cite{huang20242dgs}. \emph{Top:} gray surface renders highlighting geometric detail. \emph{Bottom:} vertex-colored renders. From left to right: scan\,24 (building), scan\,55 (figure), scan\,114 (bust).}
	\label{fig:dtu_mesh}
\end{figure}

\clearpage
\section{Normal Map Quality}
\label{sec:normal_maps}

A key benefit of using oriented triangles as rendering primitives is that each primitive carries a well-defined face normal, computed as the cross product of two triangle edges.
During alpha compositing, these per-triangle normals are blended to produce a dense normal map that reflects the local surface orientation at every pixel.

\begin{figure}[H]
	\centering
	\includegraphics[width=0.49\linewidth]{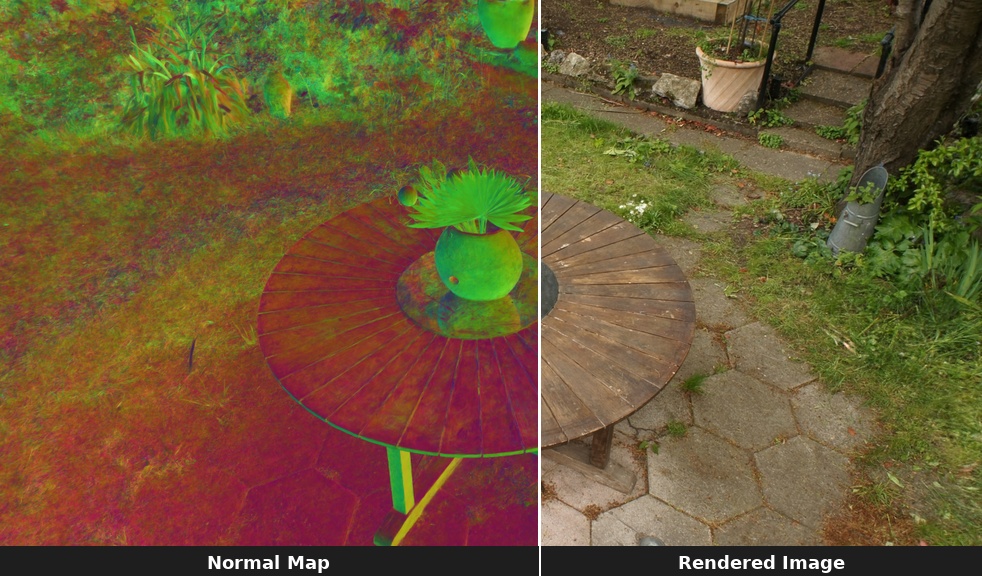}\hfill
	\includegraphics[width=0.49\linewidth]{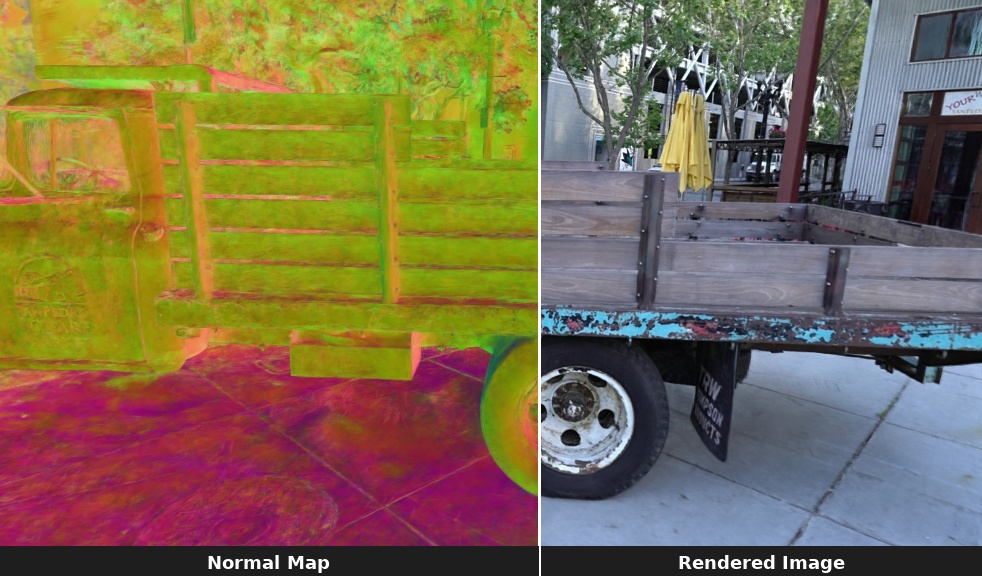}
	\caption{\textbf{Normal maps and rendered images.} Each panel shows the rendered normal map (\emph{left half}) and the corresponding RGB output (\emph{right half}) for the same test viewpoint. The normal map reveals smooth, geometrically coherent surfaces: triangle orientations are consistently aligned with the local geometry, capturing planar regions, curved surfaces, and fine structural details.
		\emph{Left:} Garden (Mip-NeRF\,360).
		\emph{Right:} Truck (Tanks \& Temples).}
	\label{fig:normal_maps}
\end{figure}

Figure~\ref{fig:normal_maps} visualizes the rendered normal maps alongside the corresponding RGB images for two representative outdoor scenes.
The normal orientations are geometrically coherent with the underlying scene structure: on the Garden table, the triangles share a consistent upward-facing orientation and lie flat on the planar surface, while the surrounding foliage exhibits smoothly varying normals that follow the organic leaf geometry.
Similarly, on the Truck scene, the normal map captures the flat panels of the vehicle body, the curvature of the wheel wells, and the ground plane, all with sharp, well-defined transitions at surface boundaries.

These normal maps are a direct by-product of the triangle-based representation and require no additional supervision beyond the standard normal consistency loss~(Sec.~3 of the main paper).
The geometric coherence they exhibit confirms that the optimized triangles are well aligned with the scene surfaces, which in turn enables high-quality depth maps for downstream mesh extraction~(Sec.~\ref{sec:dtu_geometry}).

\end{document}